\documentclass{article}
\pdfoutput=1

\PassOptionsToPackage{numbers, compress}{natbib}
\usepackage[preprint]{neurips_2024}

\usepackage{makecell}
\usepackage{multirow}
\usepackage[utf8]{inputenc} % allow utf-8 input
\usepackage[T1]{fontenc}    % use 8-bit T1 fonts
\usepackage{threeparttable}
\usepackage{hyperref}       % hyperlinks
\usepackage{url}            % simple URL typesetting
\usepackage{booktabs}       % professional-quality tables
\usepackage{amsfonts}       % blackboard math symbols
\usepackage{nicefrac}       % compact symbols for 1/2, etc.
\usepackage{microtype}      % microtypography
\usepackage[table]{xcolor}         % colors

\usepackage{soul}

\usepackage{graphicx}
\usepackage{bbm}
\usepackage[linesnumbered,ruled,vlined]{algorithm2e}

\usepackage{caption}
\usepackage{subcaption}
\usepackage{textcomp}
\usepackage{amsmath}
\usepackage{wrapfig}
\usepackage{lscape}
\usepackage[normalem]{ulem}
\newcommand{\model}{\textsc{RiGPS}}

\newcommand{\funnymodel}{\textit{\textbf{R}einforced \textbf{I}terative \textbf{G}ene \textbf{P}anel \textbf{S}election Framework}}
\DeclareMathOperator*{\argmax}{argmax} 

\title{Enhanced Gene Selection in Single-Cell Genomics: Pre-Filtering Synergy and Reinforced Optimization}
\author{%
    \textbf{Weiliang Zhang} \\
    CNIC, CAS\\
    University of CAS\\\And
     \textbf{Meng Zhen} \\
    CNIC, CAS\\\And
    \textbf{Dongjie Wang}\\
    University of Kansas\\
    \And \textbf{Min Wu}\\
        I$^2$R,A*STAR\\
    \And \textbf{Kunpeng Liu}\\
    Portland State Univerisity\\
    \And \textbf{Yuanchun Zhou}\\
    CNIC, CAS\\
   \And \textbf{Meng Xiao}\thanks{Corresponding Author}\\
    CNIC, CAS\\
    \texttt{shaow@cnic.cn}\\
}

\begin{document}

\maketitle

\begin{abstract}
Recent advancements in single-cell genomics necessitate precision in gene panel selection to interpret complex biological data effectively. 
Those methods aim to streamline the analysis of scRNA-seq data by focusing on the most informative genes that contribute significantly to the specific analysis task. 
Traditional selection methods, which often rely on expert domain knowledge, embedded machine learning models, or heuristic-based iterative optimization, are prone to biases and inefficiencies that may obscure critical genomic signals.
Recognizing the limitations of traditional methods, we aim to transcend these constraints with a refined strategy. 
In this study, we introduce an iterative gene panel selection strategy that is applicable to clustering tasks in single-cell genomics.
Our method uniquely integrates results from other gene selection algorithms, providing valuable preliminary boundaries or prior knowledge as initial guides in the search space to enhance the efficiency of our framework. 
Furthermore, we incorporate the stochastic nature of the exploration process in reinforcement learning (RL) and its capability for continuous optimization through reward-based feedback. 
This combination mitigates the biases inherent in the initial boundaries and harnesses RL's adaptability to refine and target gene panel selection dynamically.
To illustrate the effectiveness of our method, we conducted detailed comparative experiments, case studies, and visualization analysis.
% Our code and data are publicly accessible via ~\href{https://www.dropbox.com/scl/fo/7k9t43ymdmmwws2mxdrsv/AAgz7oY9AAOVwm1M336VLrs?rlkey=9qzlubycw1yiteoxft3kmzlwx&st=m32rxixx&dl=0}{Dropbox}.
% The efficacy of our approach is underscored by validation on real-world single-cell datasets, where it demonstrates substantial improvements over conventional methods. 
% Our framework's adaptability is particularly notable; it efficiently manages varying scales of genomic data and dynamically adjusts to evolving scientific insights, maintaining optimal performance.
% In conclusion, our iterative framework, which leverages pre-filtering synergy and reinforcement learning, offers a transformative strategy for gene panel selection in single-cell genomics. 
% It promises to refine our understanding of cellular mechanisms and accelerate personalized medicine applications, making it a valuable contribution to the field of computational biology.
\end{abstract}

\section{Introduction}
Single-cell RNA sequencing (scRNA-seq) represents a significant breakthrough in transcriptional data analysis~\cite{schwartzman2015single,gawad2016single,woodworth2017building}, offering a high-resolution, individualized view of each cell within tissues, organs, and organisms~\cite{lee2020single,baysoy2023technological}. 
The high-throughput data generated by this approach enables enhanced, differentiated individual tracking and analysis tasks, including spatial transcriptomic analysis of cell states~\cite{huynh2024topological}, exploration of the architecture of life at the tissue level~\cite{rao2021exploring_tissue_architecture,longo2021intercellulartissuedynamics}, identification of the cell subpopulations~\cite{sun2022identifyingphenotype-associatedsubpopulations}, or support the training of domain foundation model~\cite{cui2023scgpt,hao2023scfoundation,theodoris2023geneformer,yang2023genecompass}.
However, the data's inherent traits of high dimensionality, sparsity, and noise lead to an unavoidable curse of dimensionality in the analysis process~\cite{kiselev2019challenges(dimension_curse)}. 
These challenges motivate researchers to accomplish \emph{Gene Panel Selection} task,  which aims to strategically select a subset of genes that capture the most meaningful information with minimal redundancy. 

Prior literature efforts to partially address these issues have largely focused on three main approaches: 
\textbf{Dimensional Reduction Techniques} such as PCA~\cite{mackiewicz1993principal}, t-SNE~\cite{kobak2019art}, and UMAP~\cite{becht2019dimensionality} are essential for managing the complexity of scRNA-seq data, especially for visualization. However, these methods also have several drawbacks:
These methods can result in the \emph{loss of subtle yet biologically significant information}, \emph{distort the true structure of the data}, and are \emph{highly dependent on the choice of parameters}, such as the number of principal components or the perplexity value in t-SNE. 
\textbf{Statistical Methods},  including the use of p-values, fold changes~\cite{dalman2012fold}, or analysis of highly variable genes (HVGs)~\cite{yip2019evaluation,luecken2022benchmarking}, are fundamental step in identifying significant features in scRNA-seq data analysis~\cite{hetzel2022predicting}, or domain foundation model research~\cite{cui2023scgpt,gong2024xtrimogene}. 
However, these methods often assume data normality and independence—\emph{assumptions that may not hold true in scRNA-seq contexts} and \emph{sensitive to the inherent noise and sparsity of the data}, potentially leading to inaccuracies by either masking biological signals or amplifying artifacts. 
\textbf{Gene Selection Approaches} derived from the concept of feature selection~\cite{yu2003feature,xiao2023discrete}, tailored specifically for genomics research, including scRNA-seq studies. 
Those approaches, whether highly \emph{depend on well-trained embedded machine learning models}~\cite{CellBRF} to identify the importance of each gene, or they \emph{utilize heuristic metrics} to determine key genes~\cite{geneBasis, HRG}, are always \textit{unstable and not optimization-directed}. 

\textbf{Challenges Summary: } In summary, although existing methods have alleviated the challenges posed by high dimensionality, sparsity, and noise to a certain extent, they still exhibit several limitations. 
To address these limitations, we introduce reinforcement learning (RL) techniques into the gene selection process. 
By formulating the gene selection task as a discrete decision problem, we can design reward functions that align with specific objectives, leveraging the iterative nature of RL frameworks for optimization. 
However, applying RL to high-dimensional gene data presents its own set of challenges: 
One major challenge is \textbf{the vast action space when dealing with a large number of genes}. 
The high dimensionality of gene data can lead to exponential growth in possible actions, making it computationally intractable for reinforcement learning algorithms to explore and learn effectively. 
This curse of dimensionality can hinder the convergence and efficiency of the learning process.
Another challenge lies in \textbf{the suboptimal search performance caused by randomly chosen starting points}. 
The initial starting points of RL-based methods are significant; a poorly chosen starting point can significantly impact the effectiveness of these methods, leading to prolong search process and suboptimal gene combinations. Therefore, careful consideration of the initialization strategy of search starting points is essential for enhancing the overall performance of RL-based approaches in effective gene selection.

\textbf{Our Contribution: A reinforced iterative framework with coarser boundary and refined start points.} 
In this study, we introduce a novel framework, namely \funnymodel\ (\textbf{\model}), to apply across a diverse range of gene analysis tasks. 
Our approach is distinguished by its ability to integrate prior knowledge from existing gene panel selection algorithms. 
This prior knowledge serves as \emph{valuable preliminary boundaries} or \emph{essential prior experiences} that bootstrapped the initial phase of gene panel selection. 
Specifically, our framework leverages these preliminary boundaries as starting points, using them as initial guides in exploring the gene selection search space. 
This integration significantly boosts the efficiency of our model on high-throughput datasets, allowing for a more directed and informed initial search. 
Thus, computational overhead is reduced, and efforts are focused on the most promising gene candidates.
Moreover, we incorporate the principles of \emph{stochastic exploration} in RL~\cite{xiao2022traceable,xiao2024traceable} and its continuous optimization capabilities~\cite{wang2024reinforcement} through a reward-based feedback mechanism. 
This innovative combination allows our model to adjust and refine the gene panel selection process, mitigating the biases and limitations inherent in the initial boundaries set by previous algorithms.
The results from our experiments demonstrate substantial improvements in both the accuracy and operational efficiency of gene panel selection, paving the way for more precise biological insights and advancements in genomic research methodologies.

% \section{Preliminary}
% \subsection{Important Definition}

% \textbf{Gene Panel.} 
% \vspace{-0.3cm}
\section{Background and Preliminary}
\vspace{-0.3cm}
\textbf{Gene Panel Selection:} is a critical process in the analysis of genomic data, particularly in the context of high-throughput technologies such as single-cell RNA sequencing (scRNA-seq). 
The primary goal of gene selection is to identify a subset of genes that are most informative for specific downstream analytical tasks. 
This involves determining which genes are crucial for understanding complex biological phenomena and can vary significantly depending on the specific objectives of the study.

\textbf{Clustering Task in Single-Cell Data:} In the context of scRNA-seq, clustering is a common downstream task where gene selection plays a pivotal role. 
Clustering involves grouping cells based on their gene expression profiles to discover cell types, states, or patterns in an unsupervised manner. 
Effective gene selection is crucial here because: 
(1) \textit{Signal Enhancement}: It helps in enhancing the signal-to-noise ratio by focusing on genes that are most variable or informative across different cells.
(2) \textit{Biological Relevance}: Selected genes can highlight biological pathways and processes that define cell identity or state.
(3) \textit{Computational Efficiency}: By reducing the number of genes analyzed, computational resources are better utilized, and analyses become more manageable. 

\textbf{Gene Selection Problem for Clustering:} 
Formally, the given scRNA-seq dataset can be denoted as $\mathcal{D}=\{G, y\}$, where $G_i=\{g_i^j\}_{j=1}^n$ represents cell-$i$'s genes expression within the genomic dataset and $y_i$ is its correlated cell type. 
A gene set $\mathcal{G}$ denoted all sequencing genes within the data set. 
We aim to develop a generalized yet robust gene selection method that can identify the optimal key gene panel $\mathcal{G}^*$ from a scRNA-seq dataset $\mathcal{D}$ for downstream clustering tasks, given as:
\begin{equation}
    \label{objective}
    \mathcal{G}^{*} = \argmax_{\mathcal{G}'\subseteq \mathcal{G}} \mathcal{E}(\mathcal{C}(G[\mathcal{G'}])),
\end{equation}
where $\mathcal{G}'$ is a subset of overall gene set  $\mathcal{G}$. $\mathcal{E}$ and $\mathcal{C}$ denoted the evaluation metric and clustering method, respectively. 
We use $G[\cdot]$ to represent the selection of a specific gene subset from the scRNA-seq data. 
It is worth noting that the most different setting between feature selection and gene panel selection is that the latter will be conducted without engaging any real label.
% The clustering task $\mathcal{C}$ and the single-cell genomics dataset $\mathcal{D}$ can be replaced with any other downstream analysis task or other high-throughput omics datasets.

\vspace{-0.3cm}
\section{\model\ Framework}
\begin{figure}[!t]
    \centering
    % \vspace{-0.1cm}
    \includegraphics[width=\linewidth]{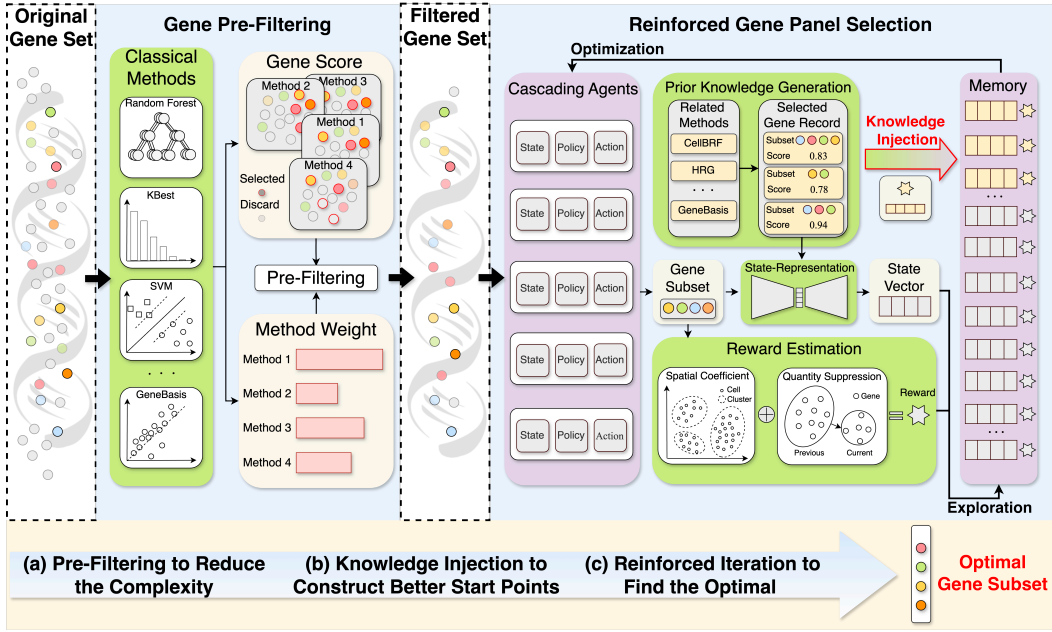}
    \caption{The overview of \model\ framework. \model\ consists of three main stages: (a) Gene Pre-Filtering to reduce the gene complexity; (b) Knowledge Injection for better start points; (c) Reinforced Key Gene Select Iteration to find the optimal selection.}
    \vspace{-0.3cm}
    \label{fig:framework}
\end{figure}
\vspace{-0.2cm}
\subsection{Framework Overview}
\vspace{-0.1cm}
\model\ is a highly efficient iterative high-throughout gene panel selection method (as illustrated in Figure~\ref{fig:framework}). 
It encapsulates a three-stage process to refine the selected gene set and extract the most informative genes for subsequent analysis.

\noindent\textbf{{Gene Pre-Filtering to Reduce Complexity:}} 
The gene expression matrix associated with the original gene set is characterized by high sparsity and noise, presenting a challenging landscape for data analysis. 
The goal of this stage is to utilize the prior knowledge inherent in basic filter-based feature and gene selection methods to discard genes with low informational content, thus providing a coarse boundary for the subsequent reinforcement learning-based selection process. 
Each method will estimate the importance scores of individual genes and then meta-vote them with their performance in downstream tasks, such as clustering, to rank and establish an initial boundary for gene selection. 
This pre-filtering stage significantly reduces the problem's complexity, allowing for a more focused and computationally manageable subsequent optimization. 

\noindent\textbf{{Knowledge Injection to Construct Better Start Points:}} 
Although pre-filtering delineates a broad boundary to control the complexity of the search space, the filtered gene set still contains a substantial number of candidate genes. 
To expedite the convergence efficiency of the entire framework, this stage leverages gene selection methods to make an informed selection within the filtered gene set. 
This selection is then injected as initial experience into the experience replay mechanism of the subsequent reinforcement learning framework. 
This knowledge injection is not merely a reduction of the search space; it is a strategic positioning that aligns the reinforcement learning with biologically relevant and statistically significant gene features, which are more likely to be conducive to discerning the underlying biological patterns and structures in the data.
By doing so, this stage provides an improved starting point for the reinforcement learning iterative optimization process, significantly enhancing the convergence efficiency of the reinforcement learning iterations. 

% \noindent\textbf{{Gene Panel State Representation}}

\noindent\textbf{{Reinforced Iteration to Find the Optimal Gene Subset:}}
The core of \model\ is its reinforced iteration mechanism, which intelligently navigates through the gene selection space. 
Within the filtered gene set, each candidate gene is evaluated by an agent that estimates the value of selecting or not selecting that gene, thereby making the optimal choice under the current state. 
These agents cooperate through a reward allocation mechanism to select the most optimal gene subset. 
The process is guided by a state-representation vector that captures the current status of the gene selection process. 
Unlike greedy search processes that rely on heuristic metrics for evaluation, the iterative reinforcement learning optimization process can learn an optimization-oriented gene panel selection strategy.
The iterative nature of the reinforced learning approach allows for continuous improvement and refinement of the gene panel, with each cycle building upon the knowledge gained from the previous one. 
This results in a robust and adaptive selection mechanism that is better suited for the complex and often nonlinear relationships inherent in gene expression data.

In the following section, we delve deeper into the strategies and methodologies under the \model\ framework, focusing on the cascading agents for gene selection. 
This includes a detailed description of the gene pre-filtering strategy (Section~\ref{met1}), cascading gene agents (Section~\ref{met2}), reward estimation (Section~\ref{met3}), the knowledge injection and iterative optimization processes (Section~\ref{met4}).

% \noindent\textbf{{Reward Estimation}}

% \noindent\textbf{{Reinforced Optimization}}
\vspace{-0.2cm}
\subsection{Gene Pre-Filtering Strategy}\label{met1}
\vspace{-0.1cm}
As illustrated in the left panel of Figure~\ref{fig:framework}, the original gene set $\mathcal{G}$ will feed into the gene pre-filtering component and form a coarse-refined filtered gene set.
We give a formal description of the gene pre-filtering strategy:

\textbf{(1) Gene Score Estimation:} 
The gene pre-filtering aims to combine basic methods to identify a coarse boundary and reduce the complexity. 
Formally, the basic selection method pipeline can be divided into estimating the importance of the gene, ranking and selecting the top-$k$ genes, denoted as: 
    $\mathcal{G} \xrightarrow[]{f^{pre}(\mathcal{D})} S,$
where $f^{pre}(\cdot)$ is the basic selection method and $S=\{s^i\}_{i=1}^n$ is the estimated score of each gene within $\mathcal{G}$. 
The method will then select the gene subset according to the score. 

\textbf{(2) Reliable Weight Evaluation:} Suppose that we have a $m$ basic gene selection methods, denoted as ${F}^{p}=\{f^{pre}_i\}_{i=1}^m$. 
Each gene in the original gene set can have its significance score calculated using the methods in ${F}^{p}$, represented as $\mathbf{S} = \{S_i\}_{i=1}^m$. 
The evaluation of each method's chosen top-$k$ gene set can be performed by assessing its downstream clustering efficacy, which is considered its reliable weight, represented as $\mathbf{P} = \{p_i\}_{i=1}^m$.

\textbf{(3) Meta-Vote for Pre-Filtering:} 
The estimated gene score and the reliable weight of each basic method then identify the gene set's coarse boundary. 
Specifically, we first calculate the normalized weights for each model. 
For model $i$, its normalized weight $w_i$ can be obtained by:
\begin{equation}
    w_i = \frac{p_i}{\sum_{p_j \in \mathbf{P}} p_j}.
\end{equation}
Then, the weight of each method can be denoted by $\mathbf{W} = \{w_i\}_{i=1}^m$. 
For gene $g_i$, its meta-vote score $\hat{s}^i$ can be obtained by weighted aggregation from the reliable weight of each method: 
\begin{equation}
    \hat{s}^i = \sum_{j\in\{1,\dots, m\}} w_j \cdot s^i_j,
\end{equation}
To identify genes whose meta-vote scores significantly deviate from the average, we first calculate the mean $\mu$ and standard deviation $\sigma$ of the scores across all genes:

\begin{equation}
\mu = \frac{1}{n} \sum_{i=1}^n \hat{s}^i,\text{   } \sigma = \sqrt{\frac{1}{n} \sum_{i=1}^n (\hat{s}^i - \mu)^2}.
\end{equation}

The filtered gene set is then selected based on whether their scores fall outside the range defined by two standard deviations from the mean (2-sigma): $\mathcal{G}_{pre} = \{g^i: \hat{s}^i > \mu + 2\sigma\}$, 
where filtered gene set $\mathcal{G}_{pre} \subseteq \mathcal{G}$ is the set of genes $g^i$ whose meta-vote scores $\hat{s}^i$ are significantly higher than the mean by at least two standard deviations. 
With the filtered gene set, the objective of the gene selection problem in Equation~\ref{objective} can be reformulated as:
\begin{equation}
    \label{final_objective}
    \mathcal{G}^{*} = \argmax_{\mathcal{G}'\subseteq \mathcal{G}_{pre}} \mathcal{E}(\mathcal{C}(G[\mathcal{G'}])),
\end{equation}
\vspace{-0.2cm}
\subsection{Cascading Agents for Gene Selection}\label{met2}
\vspace{-0.1cm}
The right panel of Figure~\ref{fig:framework} illustrates cascading agents iteratively collaborating to select the most informative genes. Specifically, we construct agents with the same number as the candidate gene from the filtered gene set. 
The learning system of gene agents consists of the following: 

\textbf{(1) Action}: the action $a_t^i$ of gene $i$'s agent at $t$-th iteration is to select or discard its corresponding gene, denoted as $a_t^i\in\{select, discard\}$. 

\textbf{(2) State}: the state at $t$-th iteration is a vectorized representation derived from the selected gene subset $\mathcal{G}_t$. 
First, we extract each gene's descriptive statistics from the selected subset to preserve the biological signal (e.g., the standard deviation, minimum, maximum, and the first, second, and third quartile, etc.). 
Then, we flatten and concatenate all descriptive statistics vectors and feed them into an autoencoder. 
This autoencoder has a fixed $k$-length latent vector and variable input and output dimensions according to the selected gene subset. 
Its goal is to minimize the reconstruction loss between the input and output, thus compressing the information from descriptive statistics vectors into a fixed size. 
After the autoencoder converges, the hidden vector $\mathcal{S}_t$ with dimension of $k$ will be used as the state representation at the $t$-th iteration.

\textbf{(3) Policy}: Every gene agent will share the state in each iteration. 
Their policy network $\pi(\cdot)$ is a feed-forward neural network with a binary classification head. 
Formally, for gene $i$, its action in $t$-th iteration is then derived by: $a_t^i=\pi^i(\mathcal{S}_t)$.
\vspace{-0.2cm}
\subsection{Reward Estimation:}\label{met3}
\vspace{-0.1cm}
Each gene's correlated agent will decide to select or discard its corresponding gene in each iteration by policy network. 
By combining those decisions, we can obtain the selection in the current iteration, given as $\mathcal{A}_t = \{a^i_t\}_{i=1}^n$.
Meanwhile, the selected gene panel can be refined by $\mathcal{G}_{pre}\xrightarrow{\mathcal{A}_t}\mathcal{G}_t$, where $\mathcal{G}_t$ is the selected subset derived from pre-filtered gene set $\mathcal{G}_{pre}$ in $t$-th iteration. 
As illustrated in Figure~\ref{fig:framework}, we designed the reward function from two perspectives: to facilitate cell spatial separability and ensure a compact number of genes.

\textbf{(1) Spatial Separability:} The first aspect of the reward function evaluates spatial separability through the normalized mutual information.
In each step, the model first clusters the cells with the current selected gene's expression and assigns each cell a pseudo-label $\hat{y}$.
Then, the reward estimator will obtain the spatial separability reward by $\hat{y}$. 
Specifically, the spatial separability reward is then calculated as follows: 
\begin{equation}
    r_t^s = \frac{2 \times I(G[\mathcal{G}_t]; \hat{y})}{H(G[\mathcal{G}_t]) + H(\hat{y})},
\end{equation}
where $I(G[\mathcal{G}_t]X; \hat{y})$ denotes the mutual information between the selected gene expression of each cell $G[\mathcal{G}_t]X$ and the pseudo labels $\hat{y}$, and $H(G[\mathcal{G}_t])$ and $H(\hat{y})$ are the entropies of $G[\mathcal{G}_t]$ and $\hat{y}$, respectively. 
This metric rewards gene agents for an effective unsupervised spatial separation understanding between and within each cluster. 

\textbf{(2) Compact Size:} The second perspective focuses on ensuring a compact number of genes through: 
\begin{equation}\label{reward_compact}
    r_t^c = \frac{|\mathcal{G}_{pre}| - |\mathcal{G}_t|}{|\mathcal{G}_{pre}| + \lambda \cdot |\mathcal{G}_t|},
\end{equation}
where $\lambda$ is a hyperparameter and $|\cdot|$ denoted the size of given set. 
This formula balances the reduction of the gene set size with the penalty for overly aggressive reduction. 
As $\lambda$ increases, the penalty for keeping too many genes (large $|\mathcal{G}_t|$) becomes more severe, thus encouraging more substantial gene reduction. 
Conversely, a lower value of $\lambda$ relaxes the penalty against the size of $|\mathcal{G}_t|$, suitable when minimal reduction is sufficient. 
This metric ensures that the selection process strategically reduces the number of genes. 

\textbf{(3) Reward Assignment:} Then we combine two perspectives and obtain the reward in step-$t$: 
\begin{equation}\label{reward_overall_func}
    r_t = \alpha \cdot r^s_t + (1-\alpha) \cdot r^c_t,
\end{equation}
where $r_t$ is the total reward in this step. $\alpha$ is a hyperparameter to adjust the weight of two perspectives. 
After obtaining the reward, the framework will assign the reward equally to each agent. 
\vspace{-0.2cm}
\subsection{Iteration and Optmization}\label{met4}
\vspace{-0.1cm}
We divide the model training into three phases. 
In the first phase, we adopt basic gene selection methods to construct better start points and then inject them into the memory queue, denoted as $\mathcal{M} = \{M_i\}_{i=1}^n$. $M_i$ represents the memory queue of gene $i$'s agent.
In the second phase, we explore and refine the selection based on the pre-filtered gene set, collect experiences, and inject them into the memory queue.
In the third phase, we train each gene agent on the experiences within the memory queue.
As illustrated in Figure~\ref{fig:framework}, the model's memory is initialized by the first phase. 
Then, the model repeats the second exploration phase to collect experiences.
When the memory queue exceeds a sufficient number of experiences, the model will explore and optimize each gene agent alternately. 

\textbf{(1) Knowledge Injection Phase:} 
Knowledge injection plays a key role in the start-up of {\model}. Given a set of basic gene selection methods, denoted as $F^k = \{f_i^{k}\}_{i=1}^m$. 
Those methods decide whether to select or discard genes from pre-filtered gene set $\mathcal{G}_{pre}$. 
Accordingly, with any given basic selection method, for gene agent $i$, an experience of the following form is injected into its memory queue: $m_i = \{\mathcal{S}^{0}, a_i^0, r_i^0, \mathcal{S}^{1}\}$. 
Here, $a_i^0$ represents select or discard the gene $i$. 
$\mathcal{S}^{0}$ and $\mathcal{S}^{1}$ are the state representation extracted from $\mathcal{G}_{pre}$ and selected gene subset.
$r_i^0$ is the reward based on the pre-filtered gene subset calculated following the reward estimation.

\textbf{(2) Exploration Phase:}
Each gene agent executes actions guided by their policy networks during the exploration phase. 
These agents process the current state as the input and choose to pick or discard its correlated gene. 
Those actions will then affect the size and composition of the gene subset, consequently refining a newly selected gene subspace. 
Concurrently, the actions carried out by the feature agents accumulate an overall reward, which is subsequently assigned to all the participating agents in the optimization phase. Specifically, for gene $i$, in step-$t$, the collected experience can be denoted as: $m_i^t =\{\mathcal{S}^{t}, a_i^t, r_i^t, \mathcal{S}^{t+1}\}$.

\textbf{(3) Optimization Phase:}
In the optimization phase, each gene agent will train their policy independently via the memory mini-batch derived by prioritized experience replay~\cite{schaul2015prioritized}. 
We optimized the policy based on the Actor-Critic approach~\cite{haarnoja2018soft}, where the policy network $\pi(\cdot)$ is the actor and $V(\cdot)$ is its correlated critic.
We define the optimization objective for agent $i$ using the expected cumulative reward, formulated as:
\begin{equation}
    \max_{\pi} \mathbb{E}_{m_i^t \sim \mathcal{B}} \left[ \sum_{t=0}^{T} \gamma^t r_i^t \right]
\end{equation}
where $\mathcal{B}$ denotes the distribution of experiences within the prioritized replay buffer, $\gamma$ is the discount factor, and $T$ represents the temporal horizon of an episode.
Moreover, we introduce the Q-function, denoted as $Q(\mathcal{S}, a)$, which represents the expected return of taking action $a$ in state $\mathcal{S}$ and following policy $\pi$ thereafter:
\begin{equation}
    Q(\mathcal{S}, a) = \mathbb{E} \left[ r + \gamma \max_{a'} Q(\mathcal{S}', a') \mid \mathcal{S}, a \right]
\end{equation}
The training updates for the actor and critic networks are computed as follows:
\begin{align}
    \text{Critic Update:} & \quad L(V) = \mathbb{E}_{m_i^t \sim \mathcal{B}} \left[ \left( V(\mathcal{S}^t) - \left( r_i^t + \gamma V(\mathcal{S}^{t+1}) \right) \right)^2 \right], \\
    \text{Actor Update:} & \quad \nabla_{\theta} J(\pi) = \mathbb{E}_{m_i^t \sim \mathcal{B}} \left[ \nabla_{\theta} \log \pi(a_i^t | \mathcal{S}^t) A(\mathcal{S}^t, a_i^t) \right].
\end{align}
where $A(\mathcal{S}, a) = Q(\mathcal{S}, a) - V(\mathcal{S})$ represents the advantage function, facilitating the gradient estimation for policy improvement.

\vspace{-0.3cm}
\section{Experiments}
\vspace{-0.3cm}
This section reports the details of the quantitative experiments performed to assess  \model\ with other baselines and ablation variations. 
And two qualitative analyses: gene expression 2-D and heatmap visualization. 
To thoroughly analyze the multiple characteristics of \model, we also analyzed the hyperparameter in reward function, the time/space scalability, gene pre-filter setting, knowledge injection setting, and reinforced optimization iteration.
For the details of those experiments, please refer to Appendix~\ref{appendix_exp}. 
The experiment settings, including the description of the data set, the evaluation metrics, the compared methods, the hyperparameter settings, and the settin of the platform, are provided in the Appendix~\ref{exp_settings}.

\vspace{-0.3cm}
\subsection{Overall Comparison}\label{overall_com}
\vspace{-0.2cm}
\begin{figure}[!t]
    \centering
    \begin{subfigure}[b]{0.32\textwidth} % [b]表示对齐底部，0.4\textwidth表示子图占页面宽度的40%
        \includegraphics[width=\textwidth]{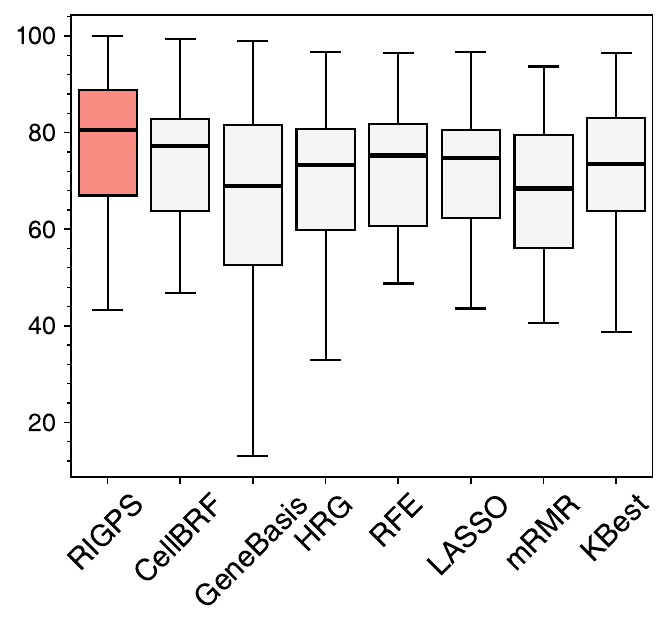}
        \caption{NMI}
        \label{fig:sub1}
    \end{subfigure}
    \begin{subfigure}[b]{0.32\textwidth} % 同样地，创建第二个子图
        \includegraphics[width=\textwidth]{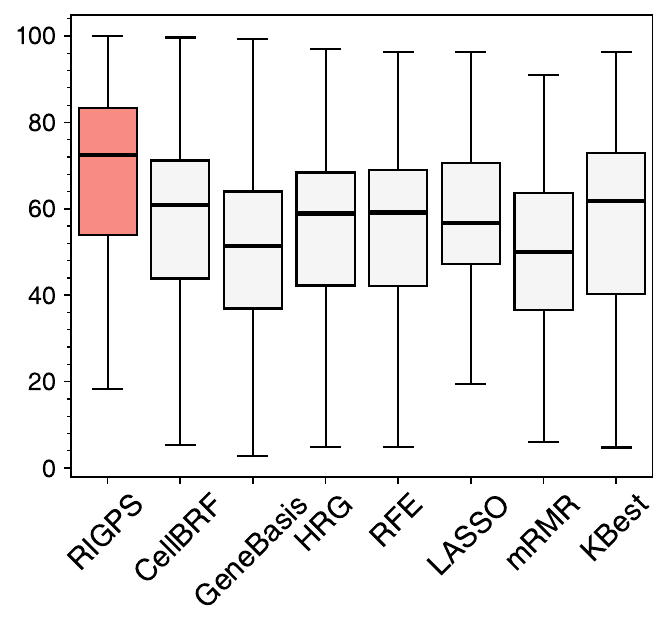}
        \caption{ARI}
        \label{fig:sub2}
    \end{subfigure}
    \begin{subfigure}[b]{0.32\textwidth} % 同样地，创建第二个子图
\includegraphics[width=\textwidth]{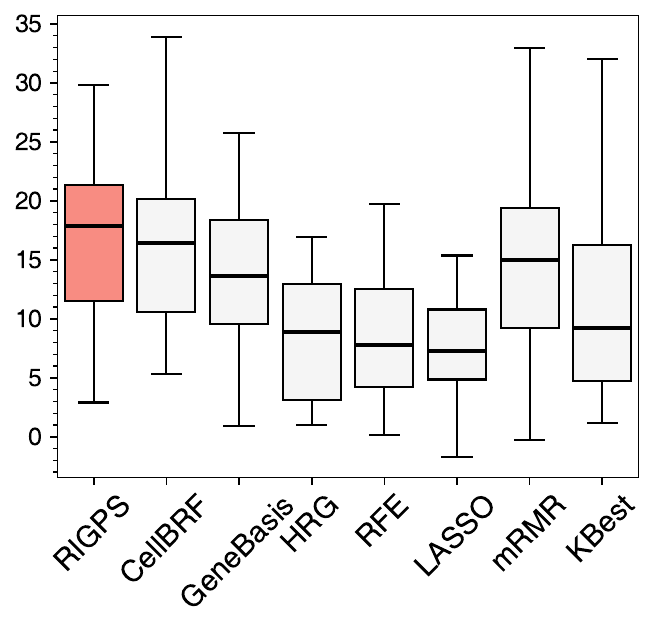}
        \caption{SI}
        \label{fig:sub2}
    \end{subfigure}
   
    \begin{subfigure}[b]{0.56\textwidth} % [b]表示对齐底部，0.4\textwidth表示子图占页面宽度的40%
    \includegraphics[width=\textwidth]
    {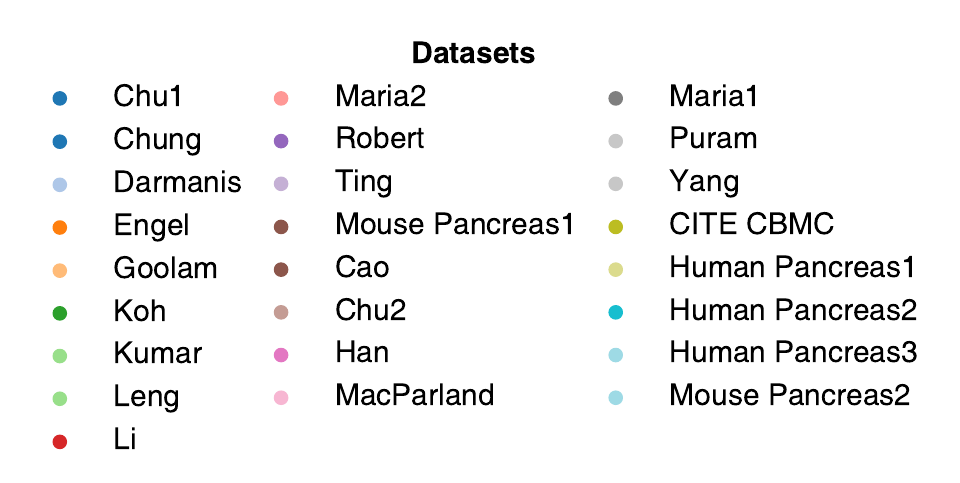}
    \end{subfigure}
    % \hspace{0.2cm}
    \begin{subfigure}[b]{0.4\textwidth} % [b]表示对齐底部，0.4\textwidth表示子图占页面宽度的40%
        \includegraphics[width=\textwidth]{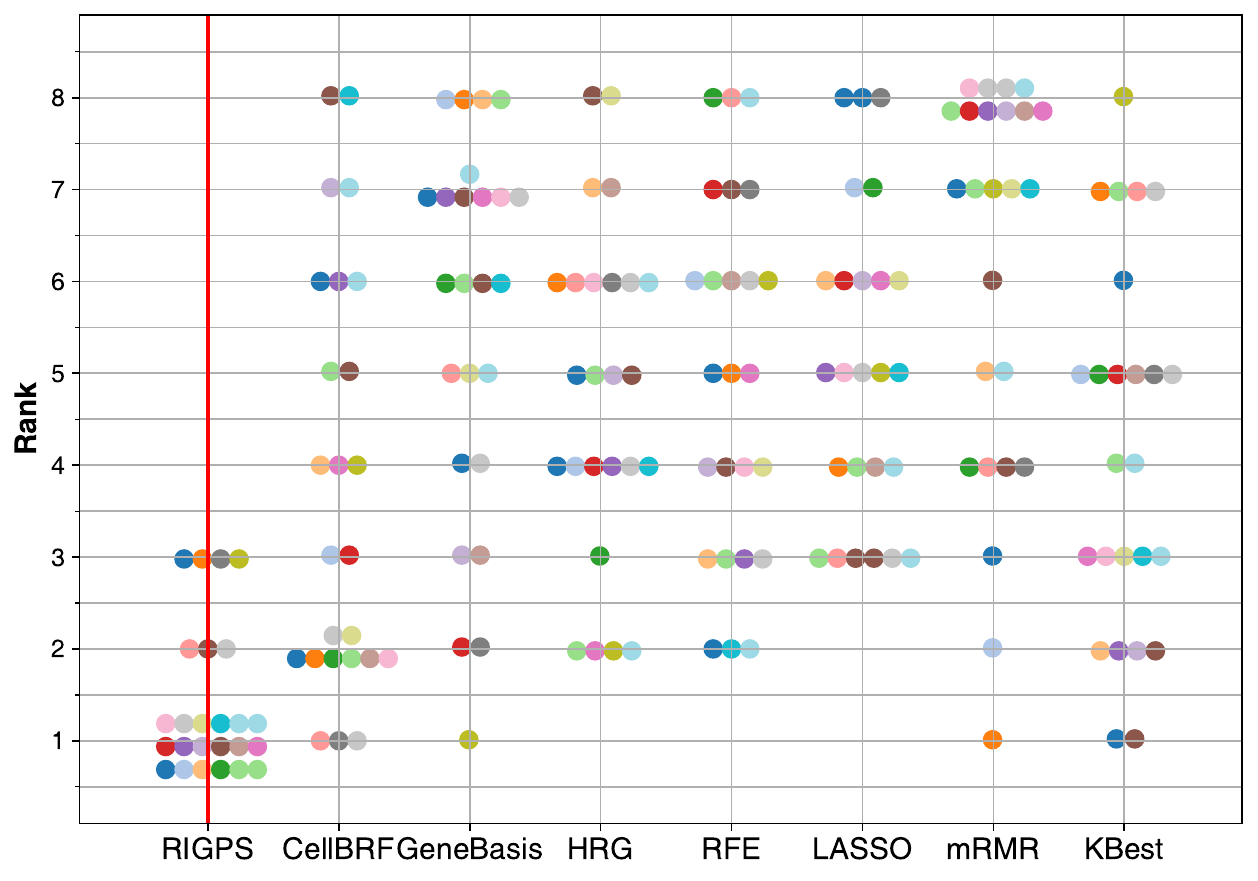}
        \caption{Performance Rank}
        \label{fig:sub4}
    \end{subfigure}
    % \begin{subfigure}[b]{0.45\textwidth} % 同样地，创建第二个子图
    %     \includegraphics[width=\textwidth]{figure/exp01/rank_ARI.pdf}
    %     \caption{Rank of ARI}
    %     \label{fig:sub5}
    % \end{subfigure}
    % \begin{subfigure}[b]{0.45\textwidth} % 同样地，创建第二个子图
    %     \includegraphics[width=\textwidth]{figure/exp01/rank_SI.pdf}
    %     \caption{Rank of SI}
    %     \label{fig:sub6}
    % \end{subfigure}
    \vspace{-0.1cm}
    \caption{Overall performance comparison: (a-c) Comparison of \model\ with seven state-of-the-art gene panel selection methods for single-cell clustering in ARI, NMI, and SI. 
    (d) Performance Rank of the gene panel selection methods in NMI.}
    \label{fig:main_figure}
    \vspace{-0.3cm}
\end{figure}

This experiment aims to answer: \textit{Is \model\ capable of effectively identifying a critical gene panel across diverse and complex datasets?}
Figure~\ref{fig:main_figure} (a-c) compares \model\ with seven gene panel selection methods for single-cell clustering in 25 datasets regarding NMI, ARI, and SI. 
Figure~\ref{fig:main_figure} (d) then shows their rank distribution (these methods are ranked from 1 to 8 by NMI for each dataset).
We observed that the average performance of \model\ on every dataset outperforms all the baseline methods. 
Additionally, \model\ achieves the highest rank on 21 out of 25 datasets and ranks within the top 3 for all datasets in terms of NMI.
The underlying driver for this observation is that \model\ eliminates redundant genes through gene pre-filtering and then effectively selects the most vital gene panel by reinforcement-optimized strategy. 
% Another interesting observation is that \model\ demonstrates leading performance in NMI, but its performance is less significant when evaluated by ARI and SI. 
% This discrepancy is due to the inherent imbalance issue present in scRNA-seq data. This imbalance arises from the nature of single-cell RNA sequencing datasets, where certain cell types are overrepresented while others are underrepresented. 
% Another interesting observation is that some methods, such as mRMR and CellBRF, partially perform well for datasets while underperforming the rest regarding specific metrics. 
% This finding suggests that the methods based on classical machine learning models can cope with the challenges of gene selection in a limited range of data environments and metrics but do not perform well when applied to a wider range of the two. 
% In \model, classical methods are integrated into prior knowledge to provide a more effective starting point for the reinforcement iteration, addressing the problem of the low robustness of classical methods.
Overall, this experiment demonstrates that \model\ is effective and robust across diverse datasets, encompassing various species, tissues, and topic-related complexities, underscoring its broad applicability for single-cell genomic data analysis tasks.
The numerical comparison results on each dataset regarding NMI, ARI, and SI are provided in Appendix~\ref{main_table}.
\vspace{-0.3cm}
\subsection{Study of the Impact of Each Technical Component}\label{ablation}
\begin{figure}[!htbp]
    \centering
    \begin{subfigure}[b]{0.48\textwidth} % [b]表示对齐底部，0.4\textwidth表示子图占页面宽度的40%
        \includegraphics[width=\textwidth]{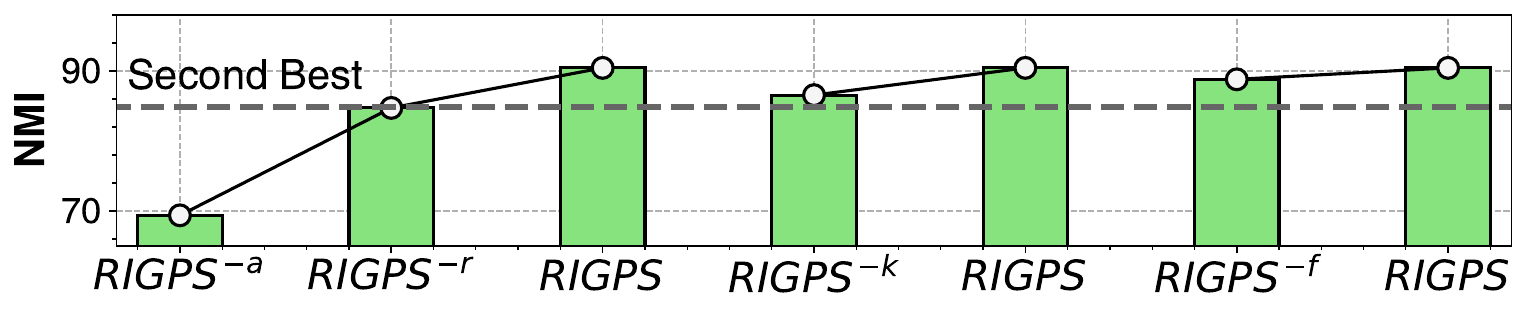}
        \caption{Chu1}
    \end{subfigure}
    \begin{subfigure}[b]{0.48\textwidth} 
        \includegraphics[width=\textwidth]{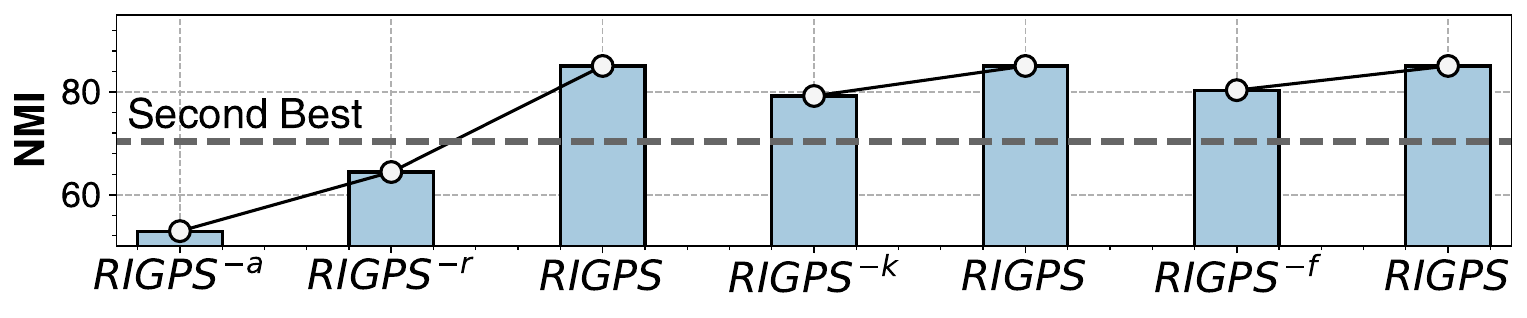}
        \caption{Leng}
    \end{subfigure}
    \begin{subfigure}[b]{0.48\textwidth} 
        \includegraphics[width=\textwidth]{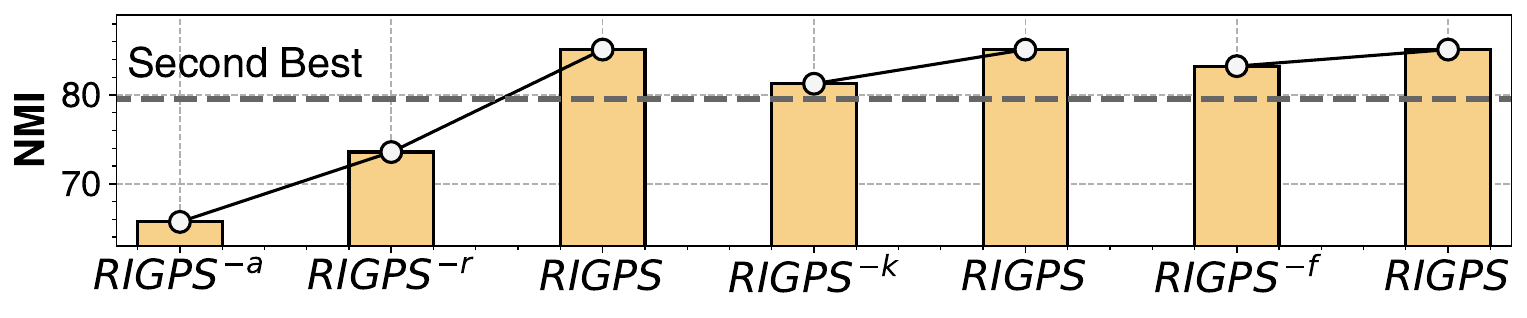}
        \caption{Puram}
        \label{fig:sub2}
    \end{subfigure}
    \begin{subfigure}[b]{0.48\textwidth} 
        \includegraphics[width=\textwidth]{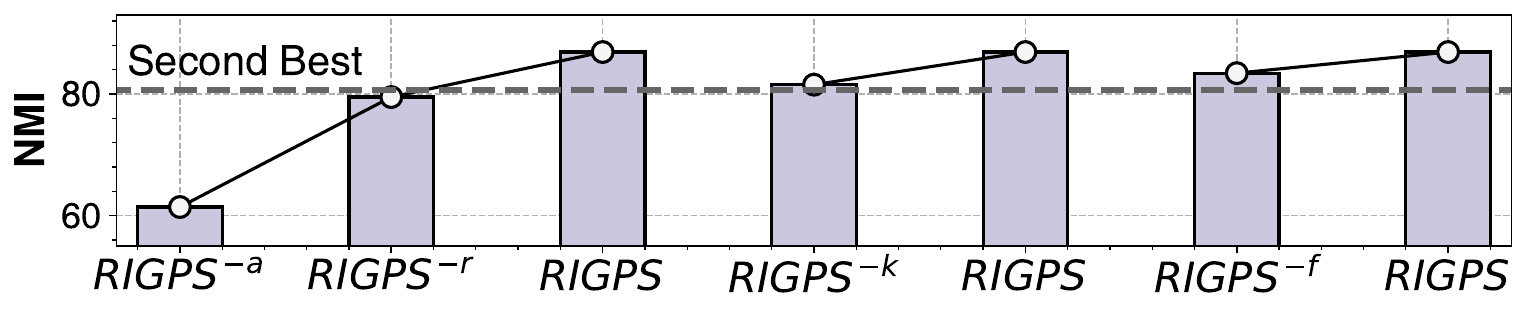}
        \caption{Mouse Pancreas1}
    \end{subfigure}
    \caption{
    Ablation studies of \model\ in terms of NMI.
    }\vspace{-0.3cm}
    \label{ablation_fig}
\end{figure}

\vspace{-0.2cm}
This experiment aims to answer: 
\textit{How does each technical component of \model\ affect its performance?}
We developed four variants of \model\ to validate the impact of each technical component.
(i) \textbf{$\model^{-r}$} uses the gene subset obtained by pre-filtering as the final gene panel without the reinforced optimization.
(ii) \textbf{$\model^{-k}$} reinforced optimize the whole pipeline without the knowledge injection.
(iii) \textbf{$\model^{-f}$} reinforced optimize the whole pipeline without the pre-filtering component.
(iv) \textbf{$\model^{-a}$} ablated all components, i.e., the performance on the original dataset.
Figure~\ref{ablation_fig} illustrates the results on Chu1, Leng, Puram, and Mouse Pancreas1 datasets.
We observed that \model\ significantly outperforms $\model^{-r}$ and $\model^{-a}$ in terms of performance.
The underlying driver is that reinforcement iteration has a powerful learning ability to screen the key gene panel from the pre-filter gene subset through iterative feedback with the reward estimation.
We also observed that \model\ is superior to $\model^{-k}$ 
in all cases.
The underlying driver is that prior knowledge injection provides a better starting point for reinforcement optimization. 
Then, RL's stochastic nature will explore and enhance them to a higher-performance gene subset.
Moreover, We found that \model\ surpasses $\model^{-f}$.
The underlying driver is that gene pre-filtering integrates multiple gene importance evaluation methods to ensure it removes the most redundant genes. 
It obtains a modest set of genes, reducing the complexity of the gene panel selection problem and helping the reinforcement iteration to find a gene panel with even better performance. 
In summary, this experiment validates that the individual components of \model\ can greatly enhance performance.

\begin{figure}[htbp]
\centering
% 第一列，竖直排列子图 a 和 b
\hspace{-0.3cm}
% \begin{minipage}[b]{0.215\linewidth}
\begin{minipage}[b]{0.27\linewidth}
    \subfloat[]{
    \includegraphics[width=\linewidth]{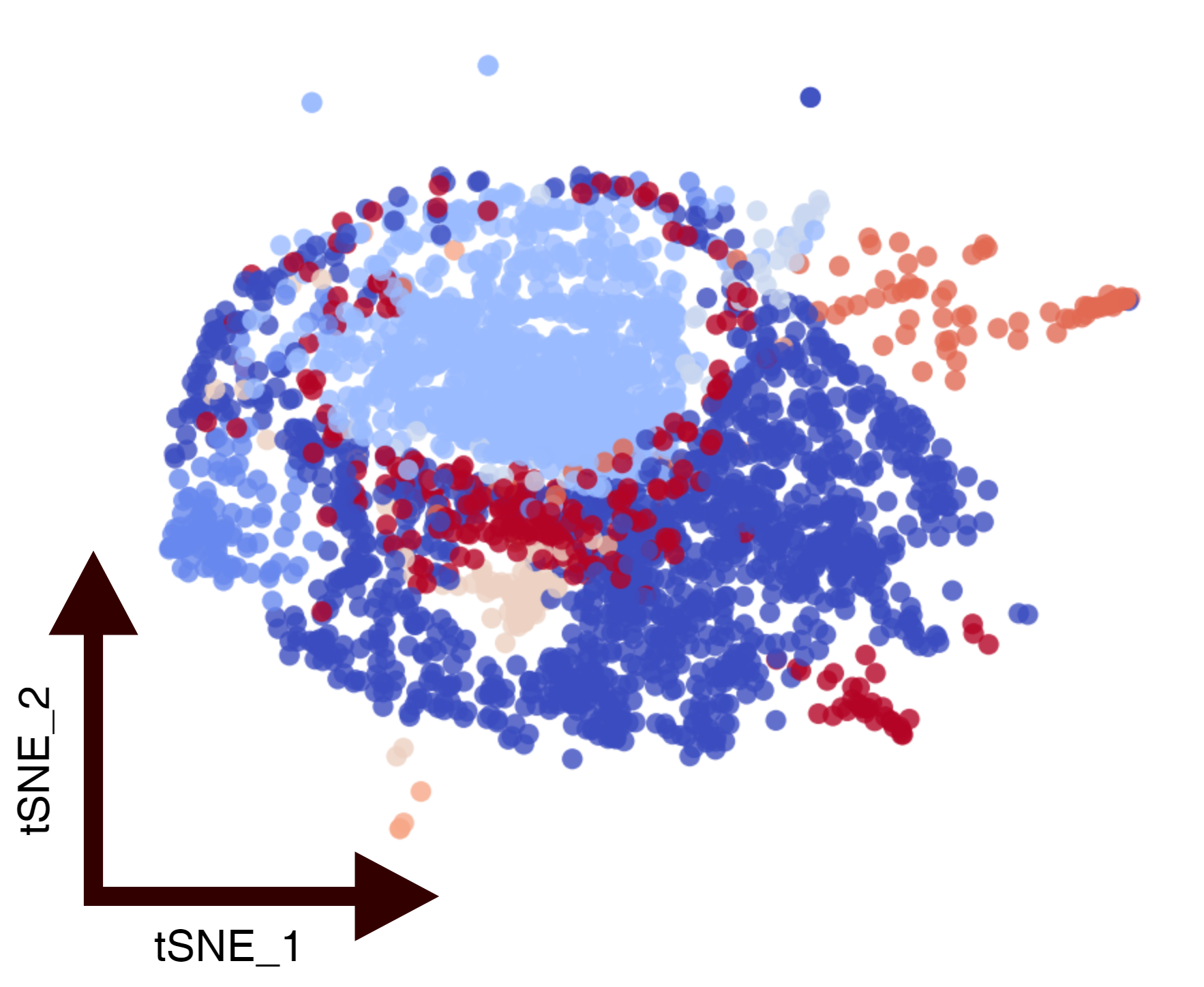}
    }\\ % 这里的 "\\" 表示换行，使得两个图像竖直排列
    \subfloat[]{
    \includegraphics[width=\linewidth]{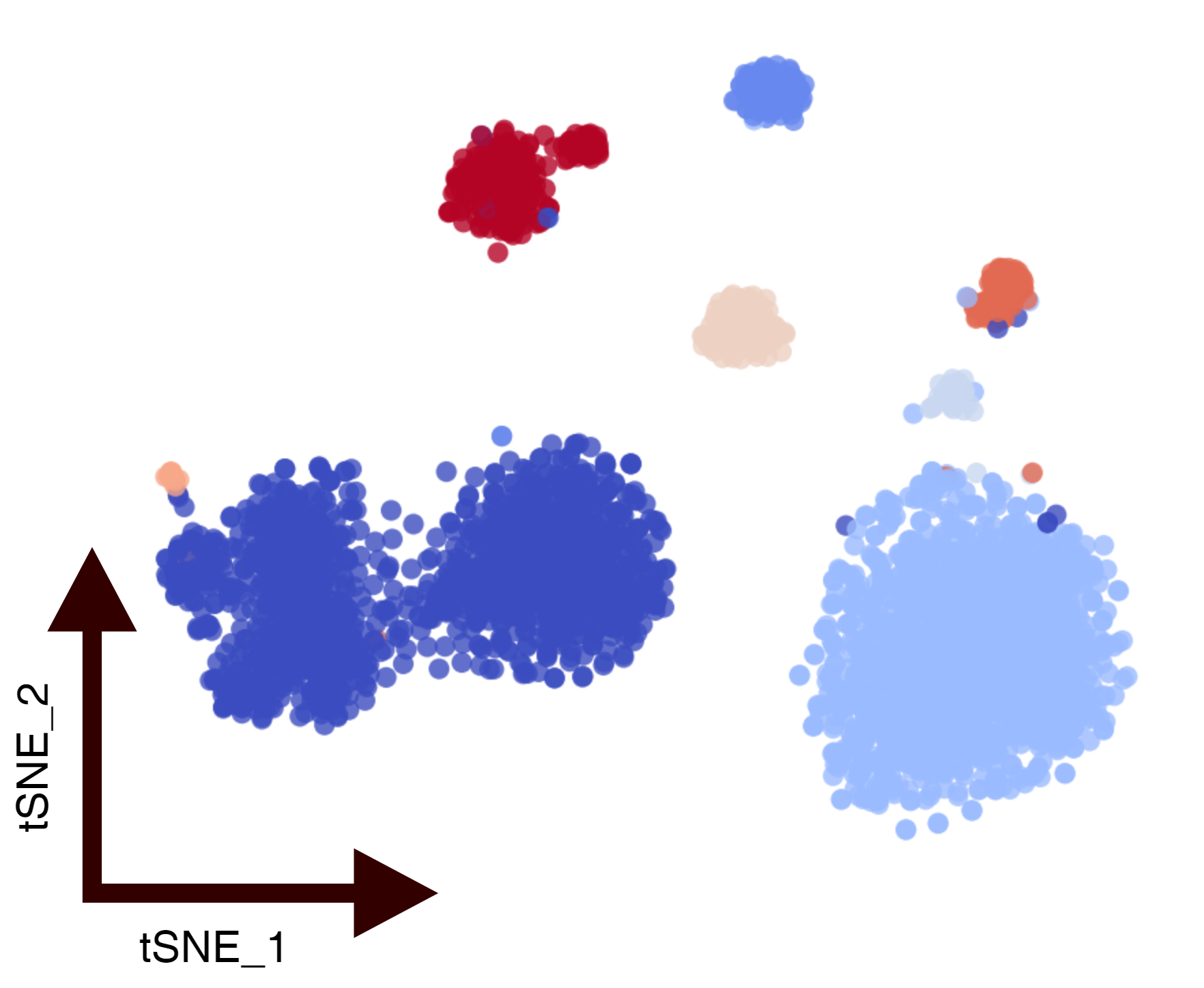}
    }
\end{minipage}
\hspace{0.1cm}
% 第二列，子图 c
\begin{minipage}[b]{0.32\linewidth}
    \subfloat[]{\includegraphics[width=\linewidth]{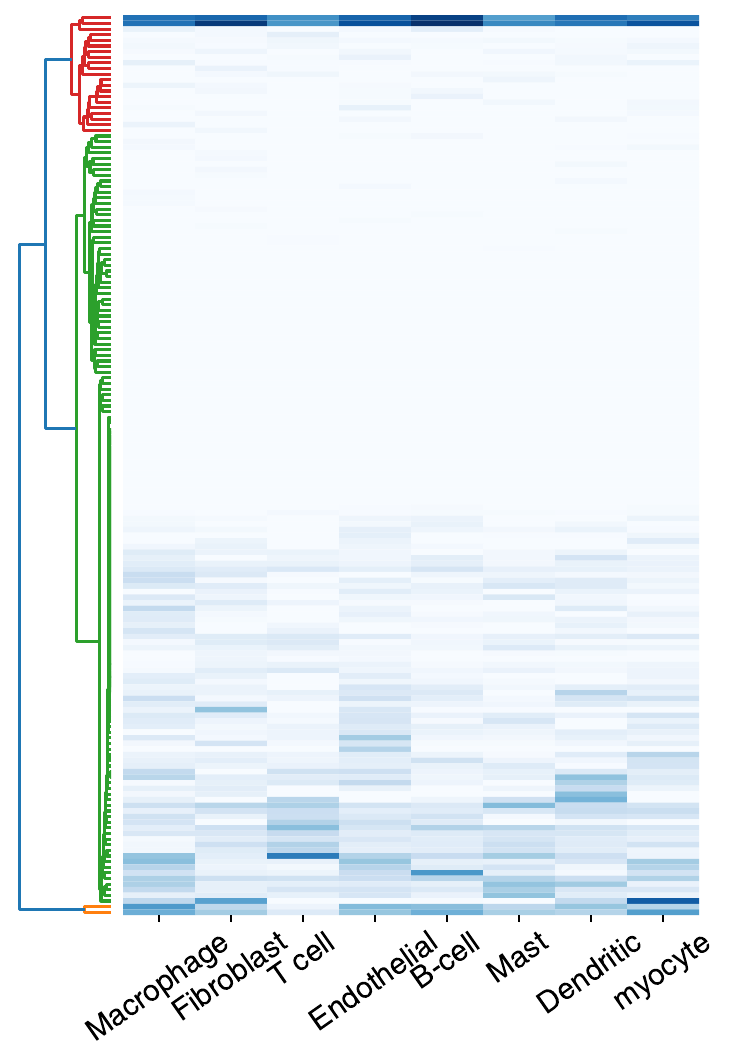}
    }
\end{minipage}
% 第三列，子图 d
\begin{minipage}[b]{0.32\linewidth}
    \subfloat[]{
    \includegraphics[width=\linewidth]{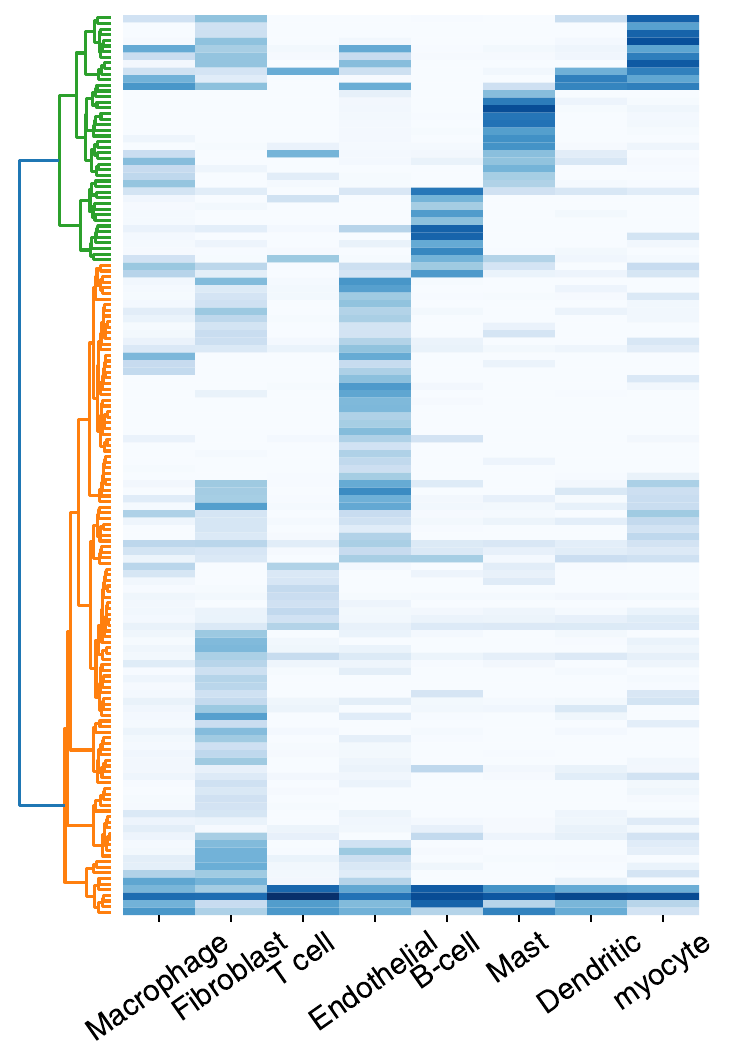}
    }
\end{minipage}
\vspace{-0.2cm}
\caption{Visualization analysis of the Puram dataset. (a) t-SNE visualization of the original dataset; (b) t-SNE visualization of \model\ optimized dataset; (c) expression heatmap of genes on the original dataset; (d) expression heatmap of genes selected by \model.}
\vspace{-0.2cm}
\label{visualization}
\end{figure}

\vspace{-0.3cm}
\subsection{Visualization Analysis of the Selected Gene Expressions}\label{vis}
\vspace{-0.2cm}
This experiment aims to answer: 
\textit{Can \model\ effectively identify key genes?}
Figure~\ref{visualization} (a-b) applies t-SNE to visualize the Puram dataset with the original genes and the gene panel selected by \model.
We found that cells with the gene subset selected by \model\ self-grouped into distinct groups according to their type, whereas cells using the original genes were tightly jumbled, and it was impossible to distinguish their cell types. 
This finding corresponds with the analysis shown in Figure~\ref{visualization} (c-d), which shows the expression heatmap for both the original genes and the gene subset chosen by \model, with the horizontal and vertical axes indicating various cells and genes, respectively. 
The intensity of the gene color increases with the level of gene expression. 
We found that the genes selected by \model\ expressed significantly different patterns between each cell type.
In contrast, the gene expression patterns from the original dataset are extremely similar and difficult to distinguish.
Those observations indicate that by following a spatial separability-based reward function, \model\ can spontaneously find the key genes that most determine cell type, resulting in a visible improvement in these visualizations. 
The t-SNE visualization and expression heatmap for the rest of the 24 datasets are shown in Appendix~\ref{rest_vis} and with the same observation.
% This observation also shows that the reinforcement iteration module of \model screens out crucial genes expressed by different cell types through the spatial coefficients in the rewards obtained in each iteration.
% Overall, this experiment validates that \model\ can select the key gene panel based on the unsupervised reward signal. 

\vspace{-0.3cm}
\subsection{Study of the Selected Gene Panel Size}
\vspace{-0.2cm}
\begin{figure}[htbp]
    \centering
\includegraphics[width=0.75\textwidth]{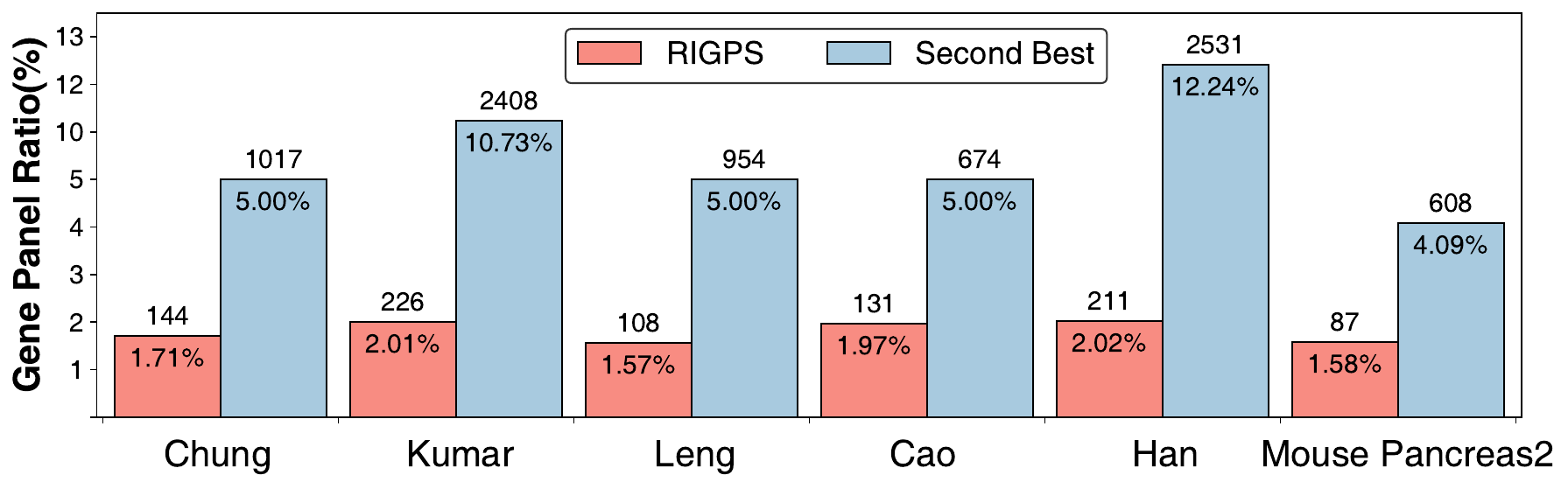}
 \vspace{-0.1cm}
    \caption{
     Comparison between \model\ and the runner-up regarding the selected gene panel size.
    }
    \vspace{-0.3cm}
    \label{compress_fig}
\end{figure}

This experiment aims to answer this question:
\textit{Is our proposed model capable of selecting a small yet effective gene subset? }
We illustrate the selected gene panel ratio between \model\ and the second-best baseline model on six datasets in Figure~\ref{compress_fig}.
We found that the gene panel obtained by \model\ is significantly more compact than the second best while still outperforming it.
We speculate the underlying driver for this observation is that gene pre-filtering will remove a mass number of redundant genes. 
Then, our reinforcement iteration carried out further screening to obtain a compact but effective gene subset. 
Furthermore, this experiment demonstrates that the gene panel selected by \model\ can effectively decrease computational expenses with better performance.

\vspace{-0.2cm}
\section{Related Work}
\vspace{-0.2cm}
Gene panel selection can be broadly categorized by selection strategies based on the statistical measure of the individual gene, the correlation among genes, or the relevance of genes and cell type.
Initial studies~\cite{yip2019evaluation,luecken2022benchmarking} often employ simple statistical metrics such as variance and mean to select genes. 
However, such methods can be suboptimal as genes with random expression across cell types may also display high variance, rendering them only marginally better than random selection~\cite{M3Drop}. 
More recent efforts have shifted towards exploring the correlation among genes. 
geneBasis~\cite{geneBasis} utilizes a k-nearest neighbor (k-NN) graph to select genes that maximize discrepancies within the graph iteratively. 
Despite their utility, these approaches often overlook the noise in gene expression-based correlation, resulting in a suboptimal performance.
Concurrently, there has been an increasing focus on the relevance of genes to specific cell types. 
These methods~\cite{FEATS, FEAST, NS-forest, gpsFISH} are generally more effective for tasks directly related to cell type. 
However, their performance may falter in applications less tied to cell typology.
Specifically, CellBRF~\cite{CellBRF} employs RandomForest to model cell clustering tasks, thereby selecting genes based on their discriminative power in tree partitioning. 
Different from these studies~\cite{}, \model, raising a new gene panel selection perspective, integrates results from other gene panel selection algorithms as prior knowledge and then employs the reinforced iteration to determine the optimal gene panel efficiently.

\vspace{-0.2cm}
\section{Conclusion Remarks}\label{limitation}
\vspace{-0.2cm}
This paper addresses the challenges inherent in single-cell genomic data analysis, such as clustering and cell type annotation, which are compounded by issues like high dimensionality, sparsity, and noise. 
% Traditional gene panel selection methods often fall short, losing critical yet subtle biological information, distorting gene expression structures, and exhibiting sensitivity to data quality, instability, or lack of optimization direction. 
We reformulate the gene selection problem through an iterative reinforced optimization approach. 
Initially, we simplify the problem by integrating basic methods to establish a preliminary boundary. 
% Within this framework, gene selection techniques are then applied to generate essential experiences. 
By leveraging the inherent stochasticity of reinforcement learning, \model\ can refine gene selection in a targeted optimization manner. 
% We conducted comprehensive quantitative and qualitative evaluations of \model, demonstrating its robust and superior performance across various scRNA-seq datasets from different species and tissues. 
We conducted comprehensive experiments to demonstrate the significance of each component.
The most noteworthy research finding reveals that \model, through the deployment of a cascading gene agent, autonomously develops a more effective gene selection strategy than traditional heuristic-based methods. 
This discovery underscores the efficacy of adopting a learning-based paradigm as the core mechanism for developing a transferable gene selection strategy applicable to diverse multi-omic and multi-species datasets. 
Our methodology does introduce certain limitations that merit discussion.
Since \model\ will construct an agent for each included gene, the overall space complexity is heavily dependent on the ability of the gene filtering process to effectively reduce the search space. 
We plan to address these limitations by exploring a hierarchical design of reinforcement learning agents that could reduce the number of gene agents. 

% \section{Limitations}
\newpage
\bibliography{ref}
\bibliographystyle{unsrt}

\newpage
\appendix
\section{Appendix}\label{Appendix}
% \begin{center}
% \textbf{\fontfamily{ppl} \fontsize{13}{0}\selectfont
% Appendix
% }%
% \bigskip
% \end{center}
% This section reports the details of the experimental settings.

\subsection{Supplementary Experiment}\label{appendix_exp}
To thoroughly analyze the multiple characteristics of \model, we provide the detailed comparison result (Appendix~\ref{main_table}), the study of the hyperparameter in reward function (Appendix~\ref{hyper_study}), the study of time/space scalability (Appendix~\ref{scalable_check}), the study of gene pre-filter setting (Appendix~\ref{gpf}), the study of knowledge injection setting (Appendix~\ref{ki}), the study of reinforced optimization iteration (Appendix~\ref{reinforce}), and the overall visualization study (Appendix~\ref{rest_vis}).

\subsubsection{Main Comparison Results}\label{main_table}
The details of the model performance comparison on each dataset regarding NMI, ARI, and SI are provided in Table~\ref{tab:main_table}.

\subsubsection{Study of Trade-off in Reward Function}\label{hyper_study}
\begin{figure}[htbp]
    \centering
        \begin{subfigure}[b]{0.47\textwidth} 
        \includegraphics[width=\textwidth]{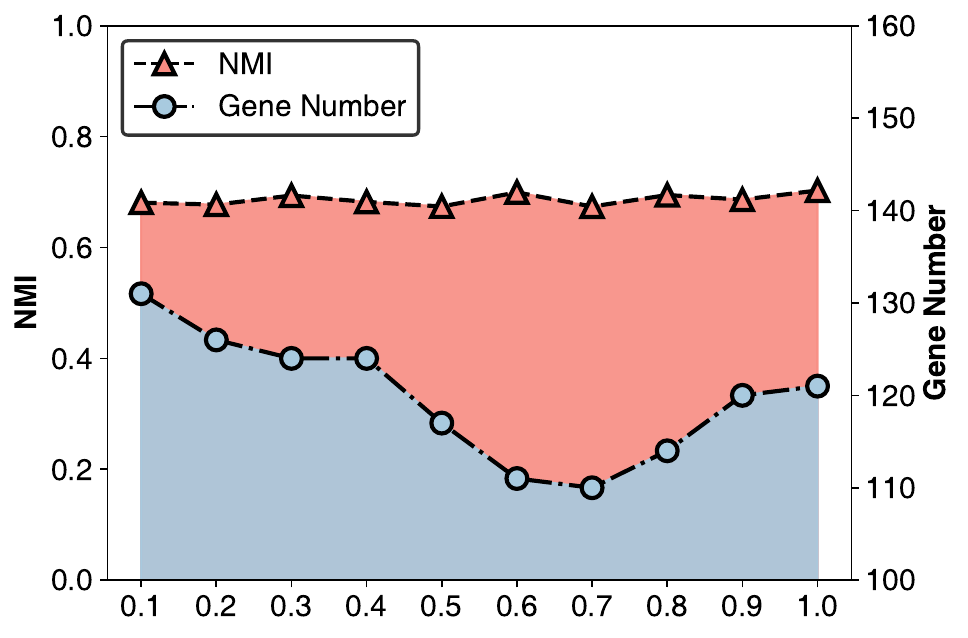}
        \caption{$\lambda$ in Equation~\ref{reward_compact}}
    \end{subfigure}
    \begin{subfigure}[b]{0.48\textwidth} % [b]表示对齐底部，0.4\textwidth表示子图占页面宽度的40%
        \includegraphics[width=\textwidth]{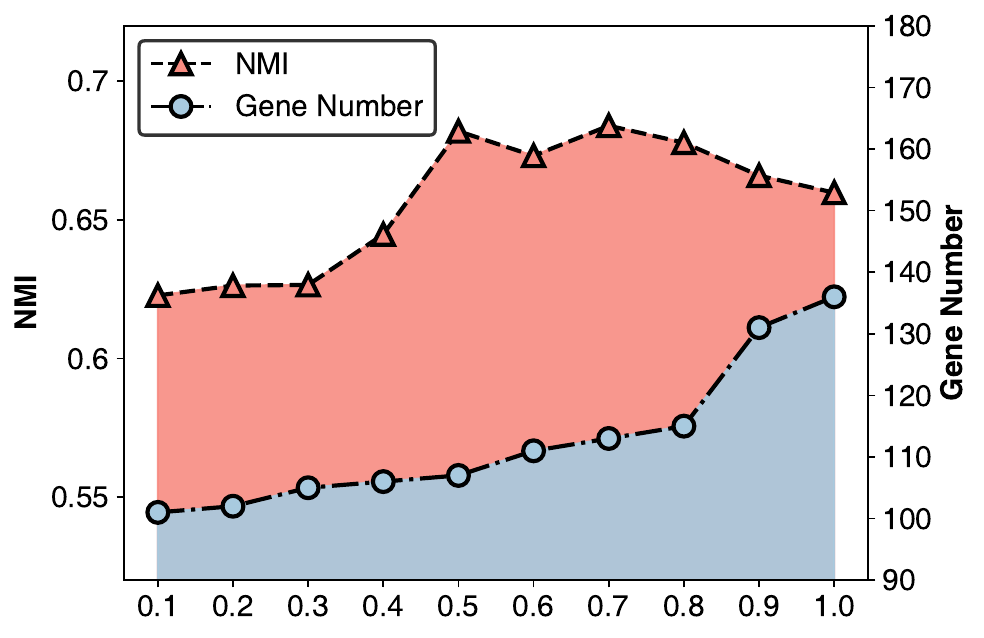}
        \caption{$\alpha$ in Equation~\ref{reward_overall_func}}
    \end{subfigure}
    \caption{
     The result of the hyperparameter sensitivity test on Cao.
    }
    \label{hyper_fig}
\end{figure}

This experiment aims to answer:
\textit{How do the reward function's hyperparameters affect the model's performance and the number of genes selected?}
In Equation~\ref{reward_overall_func}, a higher value of $\alpha$ causes the model to prioritize performance over compactness in gene selection. Conversely, a higher $\lambda$ in Equation~\ref{reward_compact} compels the model to favor a smaller gene panel.
We adjusted both $\alpha$ and $\lambda$ within the range of 0.1 to 1.0 and trained the model using the Cao dataset. 
The results are depicted in Figure~\ref{hyper_fig}.
Regarding the impact of $\lambda$, changes in this parameter do not significantly alter the model's performance, but they do affect the number of genes selected, which initially decreases and then increases. 
This pattern indicates that a higher $\lambda$ effectively suppresses the number of genes selected initially. 
However, as $\lambda$ increases, the range of variation for $r^{c}_t$ narrows when $k$ (the number of selected genes) increases, potentially leading to a reduced impact on gene number suppression. 
Consequently, the number of selected genes decreases initially but then increases as $\lambda$ rises.
Another interesting observation is that as $\alpha$ increases, the model's performance initially improves but subsequently deteriorates while the number of genes selected consistently increases.
This trend suggests that an increase in $\alpha$ reduces the influence of gene quantity suppression in the reward function, leading to an increase in the number of selected genes.
Simultaneously, the rise in spatial coefficients (due to an increased $\alpha$) initially boosts model performance.
However, the performance eventually declines due to the selection of an excessive number of genes, which introduces redundancy.
These observations confirm that the hyperparameters $\alpha$ and $\lambda$ significantly influence both the number of genes selected and the model's performance.
Optimal results are achieved with intermediate values of these parameters.
Based on our findings, we set $\alpha$ to 0.5 and $\lambda$ to 0.7 for balanced performance and gene selection compactness. 
This adjustment ensures an effective trade-off between model accuracy and the complexity of the gene panel.
The findings of this experiment effectively demonstrate that the hyperparameters $\alpha$ and $\lambda$ impact the reward function in a manner that aligns well with our objectives.

\subsubsection{Study of the Time/Space Efficiency}\label{scalable_check}
\begin{figure}[htbp]
    \centering
    \begin{subfigure}[b]{0.475\textwidth} % [b]表示对齐底部，0.4\textwidth表示子图占页面宽度的40%
        \includegraphics[width=\textwidth]{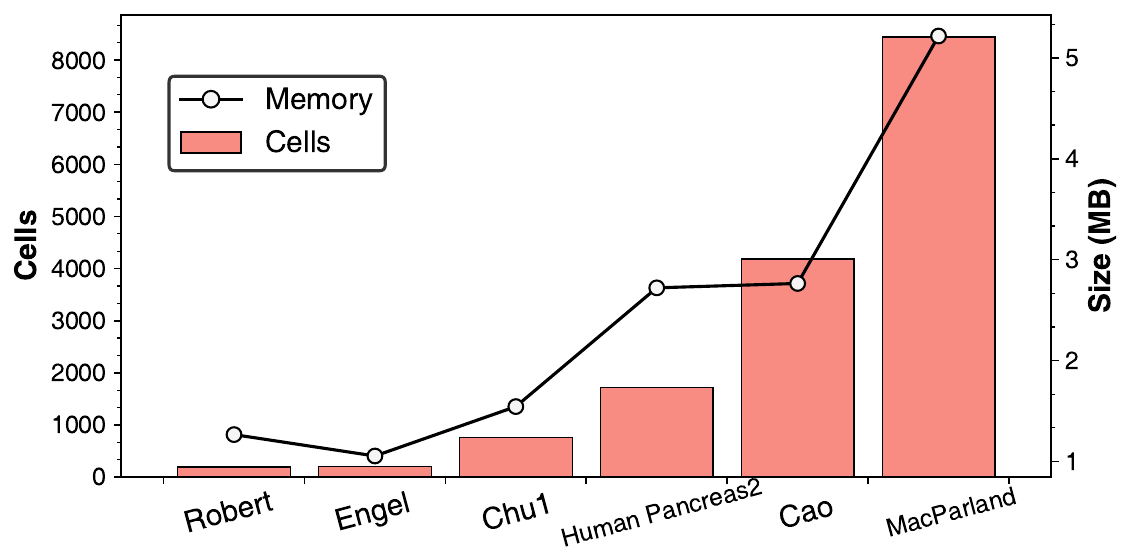}
        \caption{Parameter Size}
    \end{subfigure}
    \begin{subfigure}[b]{0.49\textwidth} % [b]表示对齐底部，0.4\textwidth表示子图占页面宽度的40%
        \includegraphics[width=\textwidth]{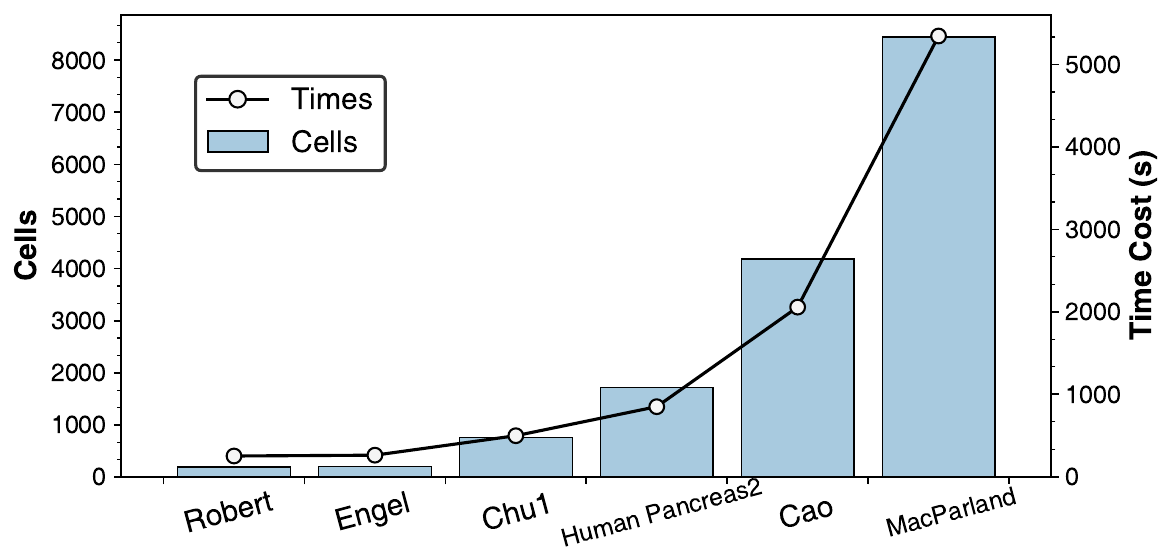}
        \caption{Train Time}
    \end{subfigure}
    \caption{
      Scalability check of \model\ regarding parameter size and training time.
    }
    \label{time_space_efficiency_fig}
\end{figure}

This experiment aims to answer the following question:
\textit{is  \model\ excels in both temporal (time efficiency) and spatial (memory usage)?}
To this end, we selected six scRNA-seq datasets varying in cell count—Robert, Engel, Chu1, Human Pancreas2, Cao, and MacParland—ranging from small to large to provide a comprehensive evaluation. 
Figure \ref{time_space_efficiency_fig} illustrates the comparison results in terms of model parameter size and training time across these datasets. Our analysis revealed the following key insights: \textbf{(1) Parameter Size Efficiency:} We observed that the parameter size of \model\ increases proportionally with the number of cells. This indicates that the state representation component of the reinforcement iteration, specifically the autoencoder, efficiently compresses the gene panel into a k-length latent vector. This transformation significantly reduces the parameter size compared to models that might not leverage such efficient encoding mechanisms, thus demonstrating spatial efficiency. \textbf{(2) Training Time Efficiency:} The training time exhibited a linear relationship with the number of cells. This linear scalability suggests that \model\ maintains consistent training durations relative to dataset size, which is indicative of robust learning capabilities. The reinforcement iteration mechanism of \model\ effectively identifies the most efficient gene panel within a limited number of iterations, showcasing its temporal efficiency. 
The datasets chosen for this experiment are particularly challenging due to their high-throughput nature—typically having a much larger number of genes than cells. Despite these challenges, the fact that both parameter sizes and training times of \model\ are proportional to the number of cells underscores the model's adaptability and efficiency.
In conclusion, \model\ demonstrates significant advantages in terms of both temporal and spatial complexities when applied to scRNA-seq datasets. Its ability to scale linearly with the number of cells, combined with the efficient data representation using autoencoders, clearly highlights its superiority in handling large-scale genomic data. This makes \model\ an attractive solution for applications requiring efficient data processing in both time and space dimensions.

\subsubsection{Study of Gene Pre-Filtering Module Settings}\label{gpf}

\begin{figure}[htbp]
    \centering
    \begin{subfigure}[b]{0.245\textwidth} % [b]表示对齐底部，0.4\textwidth表示子图占页面宽度的40%
        \includegraphics[width=\textwidth]{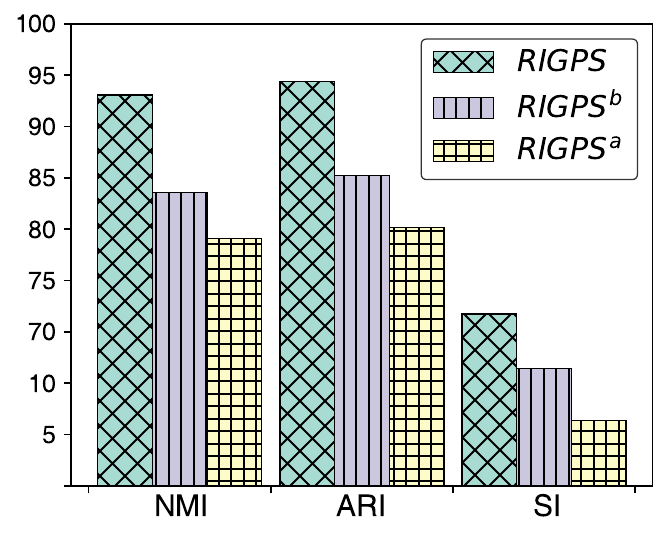}
        \caption{Leng}
    \end{subfigure}
    \begin{subfigure}[b]{0.24\textwidth} 
        \includegraphics[width=\textwidth]{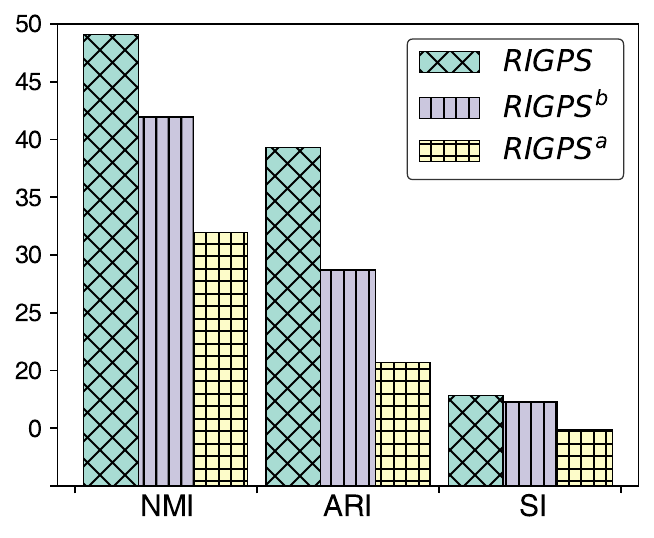}
        \caption{Maria2}
    \end{subfigure}
    \begin{subfigure}[b]{0.24\textwidth} 
        \includegraphics[width=\textwidth]{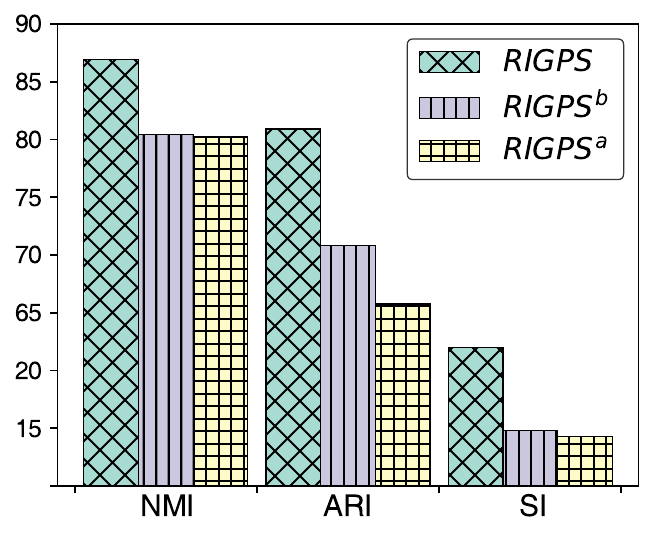}
        \caption{Mouse Pancreas1}
        \label{fig:sub2}
    \end{subfigure}
    \begin{subfigure}[b]{0.24\textwidth} 
        \includegraphics[width=\textwidth]{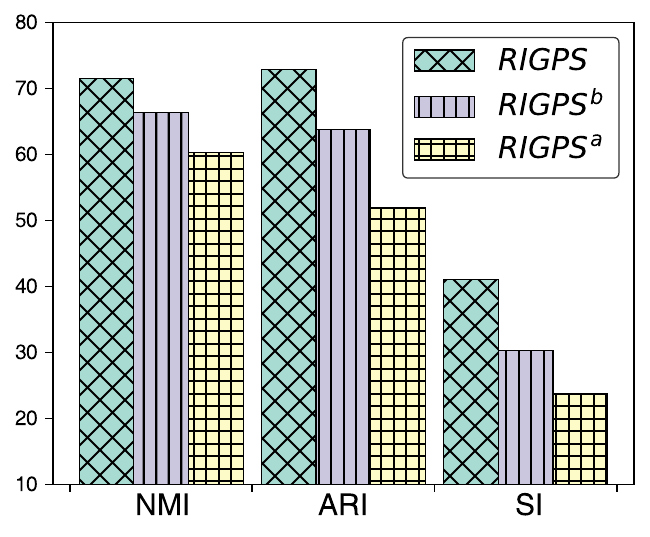}
        \caption{Robert}
    \end{subfigure}
    \caption{
     $\model^{a}$ adopts RandomForest as the gene evaluating method in the gene pre-filtering module and $\model^{b}$ adopts RandomForest, SVM, and RFE as gene evaluating methods in the gene pre-filtering module.
    }
    \label{gene_pre_filter_fig}
\end{figure}

This experiment aims to answer: 
\textit{How do different basic gene selection method combinations in pre-filtering affect the performance of \model?}
To examine the impact of different gene pre-filtering module settings, we developed two model variants of \model:
(i) $\model^{a}$: adopting \textit{Random Forest} as the gene evaluating method in the gene pre-filtering module.
(ii) $\model^{b}$: adopting \textit{Random Forest}, \textit{SVM}, and \textit{RFE} as gene evaluating methods in the gene pre-filtering module. 
(iii) $\model$: as introduced in Appendix~\ref{exp_settings}, the basic methods in our method consist of \textit{Random Forest}, \textit{SVM}, \textit{RFE}, \textit{geneBasis}, and \textit{KBest}
The comparative analysis of these variants was conducted using datasets from Leng, Maria2, Mouse Pancreas1, and Robert, with the results depicted in Figure \ref{gene_pre_filter_fig}. The findings from this study are as follows:
We found that the performance of downstream clustering tasks correlates with the number of gene-evaluating methods; the more gene-evaluating methods there are, the better the clustering effect. 
This illustrates that gene pre-filtering is scalable.
It uses many gene-evaluating methods from multiple perspectives to identify a more comprehensive and complete set of important genes, making the reinforcement iteration more likely to converge in a gene subset with superior performance.
Thus, our gene pre-filtering module options a larger and more comprehensive subset of vital genes and avoids the problem of missing key information, which is highly scalable and correlated with the performance of \model.

\begin{figure}[htbp]
    \centering
        \begin{subfigure}[b]{0.24\textwidth} % [b]表示对齐底部，0.4\textwidth表示子图占页面宽度的40%
        \includegraphics[width=\textwidth]{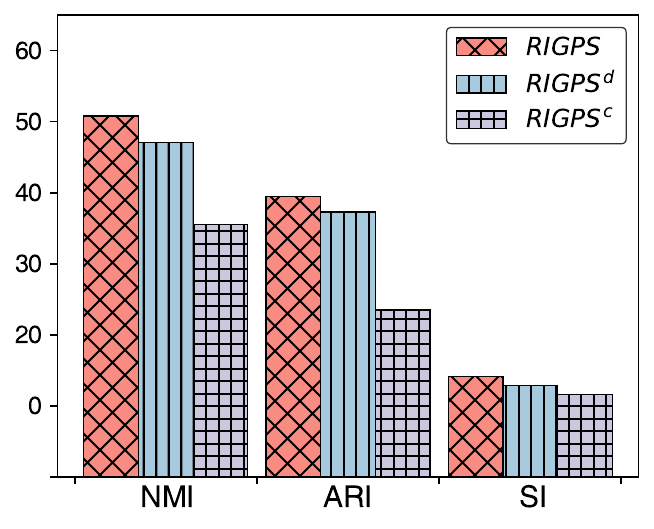}
        \caption{Maria2}
    \end{subfigure}
    \begin{subfigure}[b]{0.24\textwidth} 
        \includegraphics[width=\textwidth]{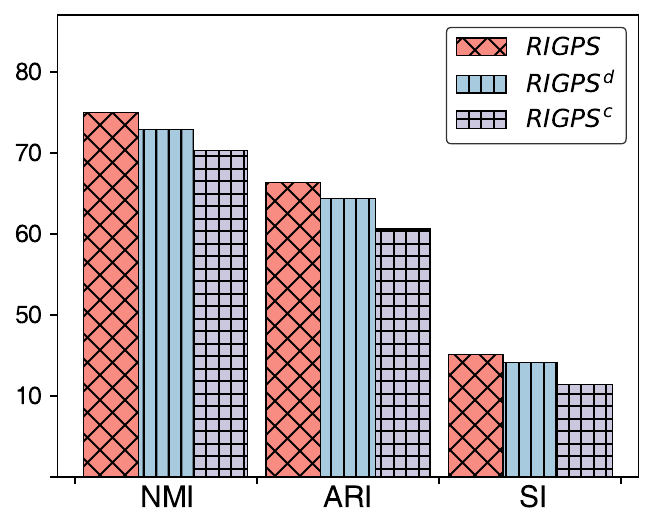}
        \caption{Cao}
    \end{subfigure}
    \begin{subfigure}[b]{0.24\textwidth} 
        \includegraphics[width=\textwidth]{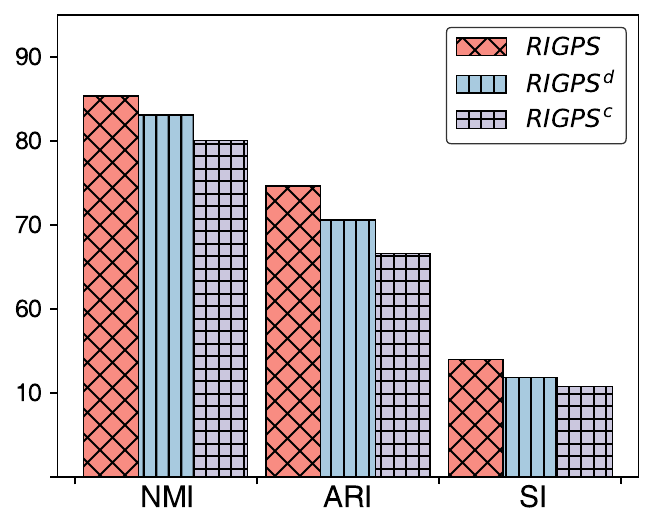}
        \caption{Puram}
        \label{fig:sub2}
    \end{subfigure}
    \begin{subfigure}[b]{0.245\textwidth} 
        \includegraphics[width=\textwidth]{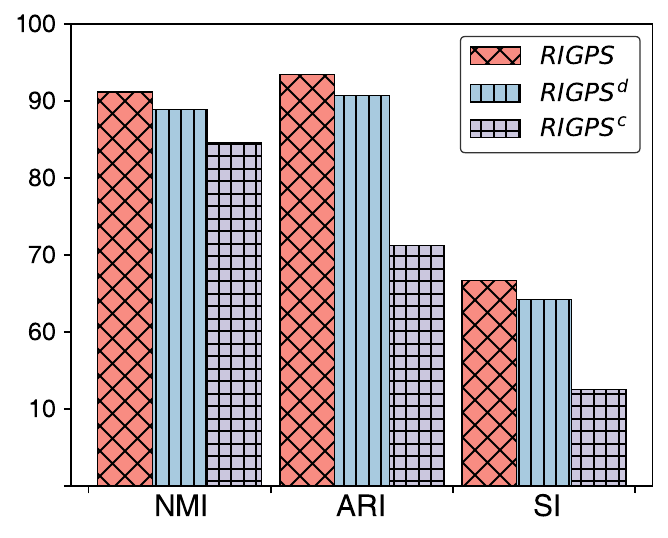}
        \caption{Human Pancreas2}
    \end{subfigure}
    \begin{subfigure}[b]{\textwidth} % [b]表示对齐底部，0.4\textwidth表示子图占页面宽度的40%
        \includegraphics[width=\textwidth]{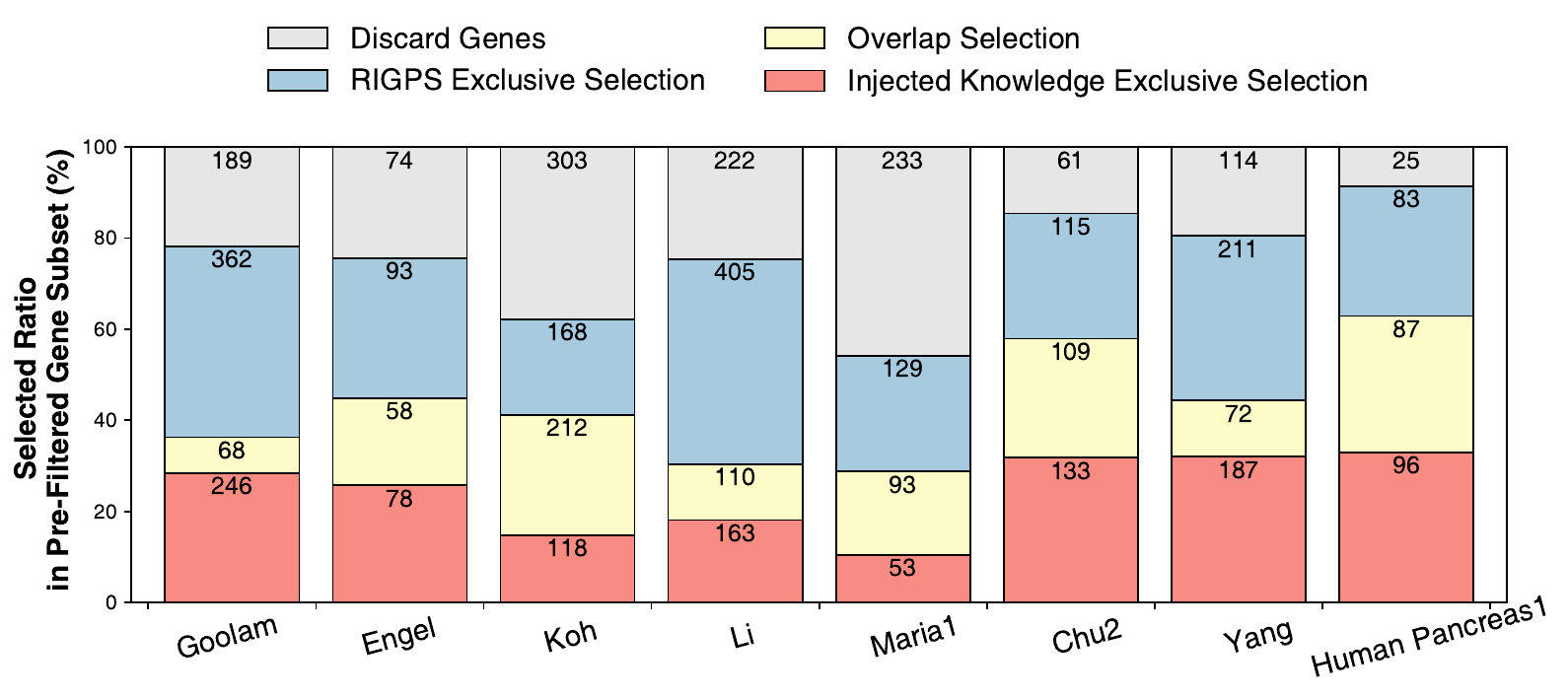}
        % \vspace{-0.2cm}
        \caption{Overlap comparison}
    \end{subfigure}
    \caption{
    (a-d) The performance of $\model$, $\model^{c}$, and $\model^{d}$ on Maria2, Cao, Puram, Human Pancreas2 datasets. 
    (e) The comparison of the selected result in the pre-filtered gene subset by \model\ exclusive selection, overlap selection, and injected knowledge exclusive selection.
    }
    \label{prior_knowledge_fig}
\end{figure}

\subsubsection{Study of Prior Knowledge Injection Settings}\label{ki}
This first part of this experiment aims to answer:
\textit{How do different basic gene selection method combinations in knowledge injection affect the performance?} 
To validate the effectiveness and extensibility of knowledge injection, we developed two model variants to establish the control group: 
(i) $\model^{c}$, we injected the gene panel selected by \textit{CellBRF} as the prior knowledge. 
(ii) $\model^{d}$, we injected the gene panels selected by \textit{CellBRF}, \textit{geneBasis}, and \textit{HRG} as the prior knowledge.
(iii) $\model$, as introduced in Appendix~\ref{hyper}, we injected the gene panels selected by \textit{CellBRF}, \textit{geneBasis}, \textit{HRG}, \textit{mRMR}, and \textit{KBest} as the prior knowledge. 
Figure \ref{prior_knowledge_fig} (a-d) shows the comparison results on Maria2, Cao, Puram, and Human Pancreas2. 
We found that as prior knowledge increases, the gene panel obtained by reinforcement iterations becomes increasingly effective. 
This illustrates that increasing prior knowledge injection allows the reinforcement iteration module to attain more high-quality starting points, leading to a better-performing gene panel.
While models such as CellBRF, which uses a single classical machine learning method, and geneBasis, which iterates using artificial statistical metrics, both have limitations in the gene panel obtained, \model\ can integrate the gene subsets from these methods through the prior knowledge injection to find a superior gene panel in performance.
Thus, prior knowledge injection does help \model\ to find a unique and enhanced genes panel while distinguished by its robust scalability.

The second part of this experiment aims to answer: \textit{Will \model\ adopt the stochastic nature to refine the start point?} 
Figure \ref{prior_knowledge_fig} (e) shows the comparison of the selected ratio in the pre-filtered gene subset by \model\ exclusive selection, overlap selection, and injected knowledge exclusive selection on 8 randomly selected datasets.
From the figure, we can first observe the overlap (colored in yellow) between the injected gene set and the RL-refined gene set in a relatively small proportion. 
We also found that the gene panel selected by \model\ is substantially varied from prior knowledge.
This illustrates that reinforcement iteration with prior knowledge does not simply repeat the injected selection pattern. 
In contrast, prior knowledge will help reinforcement iteration to get a better starting point while allowing the framework to refine the selection and search for a more excellent gene panel.

In summary, the experiments validate that integrating diverse gene selection methods as prior knowledge and the stochastic nature of reinforcement learning contribute significantly to the superior performance and robustness of \model.

\subsubsection{Study of Comparison between Reinforced Optimization and Heuristic Optimization}\label{reinforce}
% 与其他迭代方法对比收敛情况

\begin{figure}[htbp]
    \centering
    \begin{subfigure}[b]{0.245\textwidth} % [b]表示对齐底部，0.4\textwidth表示子图占页面宽度的40%
        \includegraphics[width=\textwidth]{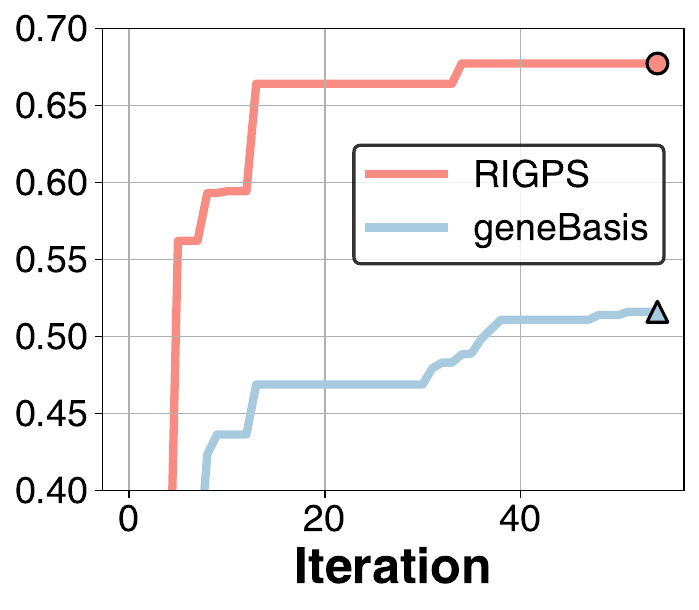}
        \caption{Cao}
    \end{subfigure}
    \begin{subfigure}[b]{0.24\textwidth} 
        \includegraphics[width=\textwidth]{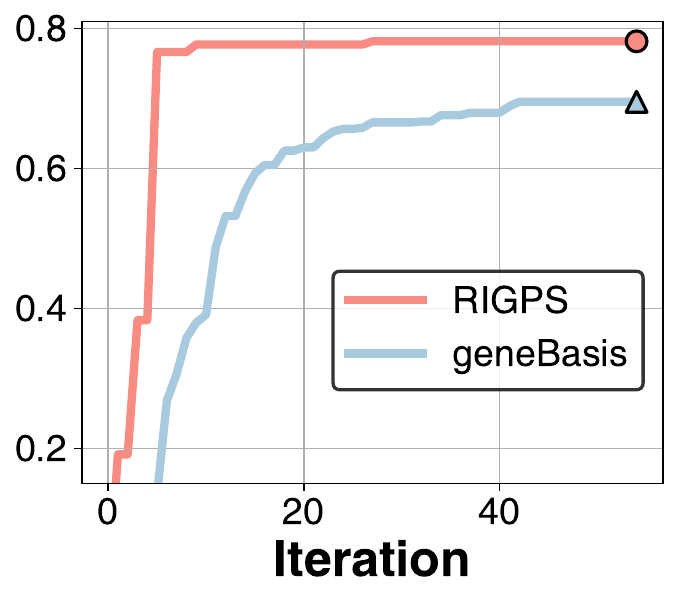}
        \caption{Han}
    \end{subfigure}
    \begin{subfigure}[b]{0.24\textwidth} 
        \includegraphics[width=\textwidth]{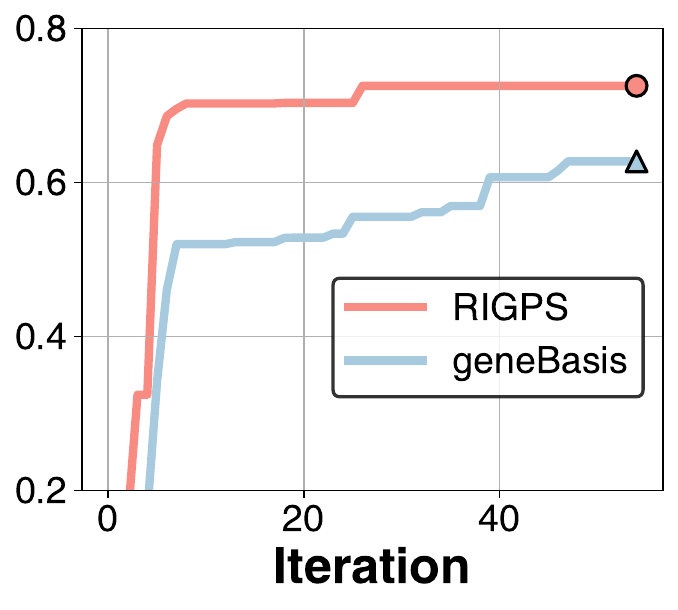}
        \caption{Yang}
        \label{fig:sub2}
    \end{subfigure}
    \begin{subfigure}[b]{0.24\textwidth} 
        \includegraphics[width=\textwidth]{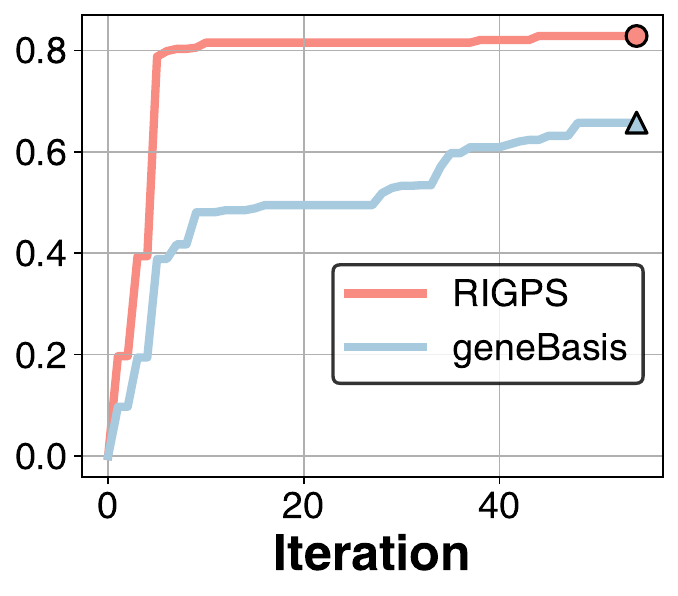}
        \caption{Puram}
    \end{subfigure}
    \caption{
        Iterative convergence speed of \model\ and geneBasis
    }
    \label{iteration_speed_fig}
\end{figure}

This experiment aims to answer: \textit{Will the rules learned by \model\ outperform heuristic iteration?} 
Figure~\ref{iteration_speed_fig} shows the performance (NMI) of genes selected in the first 50 iterations of \model\ and geneBasis (a commonly used iteration-based gene selection method by optimizing and selecting the gene that can minimize Minkowski distances in each step) on Cao, Han, Yang, and Puram datasets.
We found that the speed of convergence and the performance of \model\ at convergence are far better than geneBasis. 
This observation indicates that our reinforcement iteration can quickly and accurately find the best-performing gene subset by interacting with the environment through the rewards of each iteration, compared to geneBasis which simply considers maximizing statistical metrics at each iteration. 
This demonstrates that the reinforcement iteration possesses strong learning capabilities and robustness. 
Therefore, this experiment proves that \model\ is superior to existing methods both in terms of the speed of iterative convergence and the performance of the gene subset obtained after convergence.

\subsubsection{Study of Visualization on Each Dataset}\label{rest_vis}
\begin{figure}[htbp]
\centering
\renewcommand{\thesubfigure}{\arabic{subfigure}}
\subfloat[Cao]{
		\includegraphics[scale=0.10]{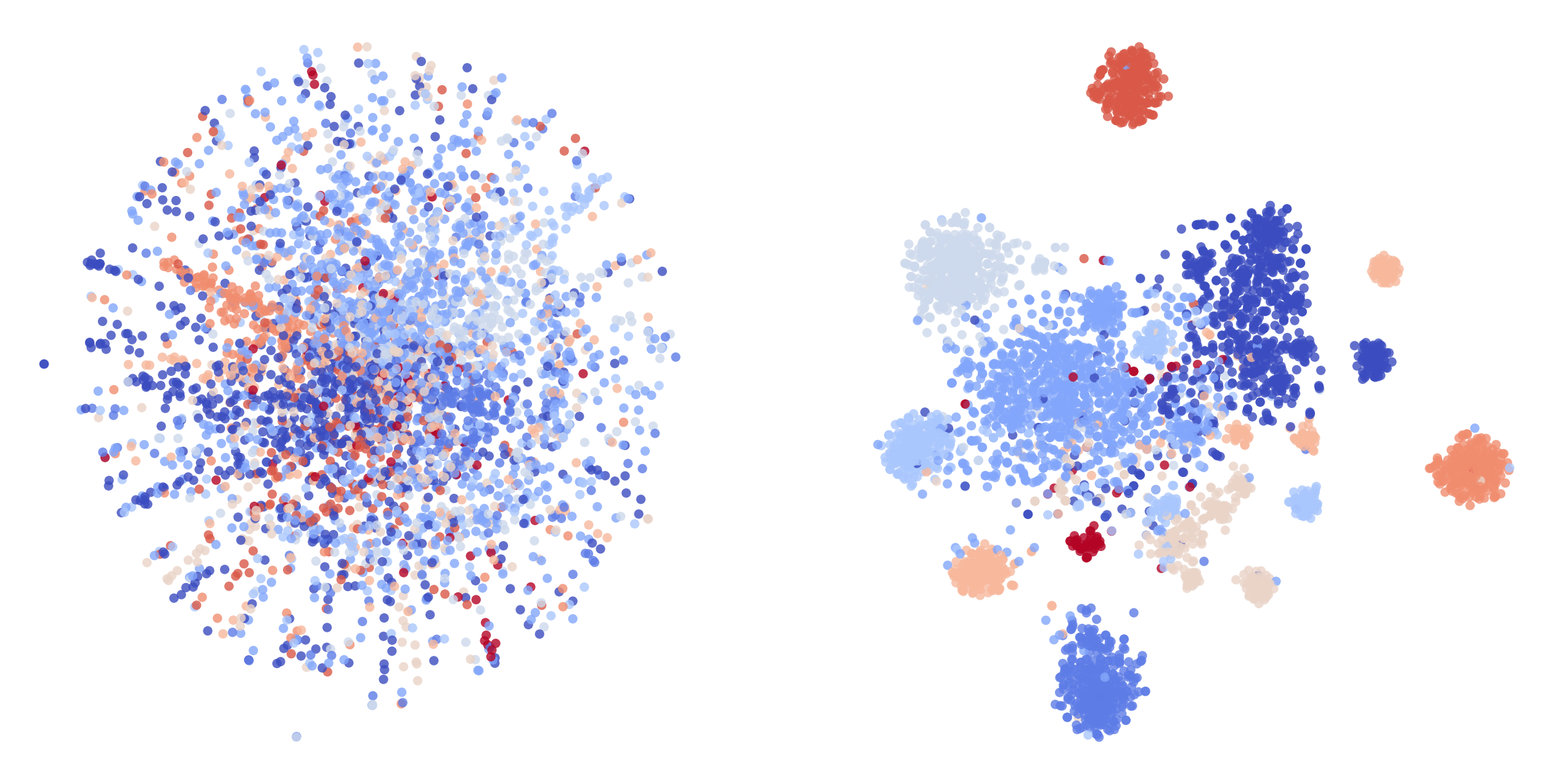}}
\subfloat[Chu1]{
		\includegraphics[scale=0.10]{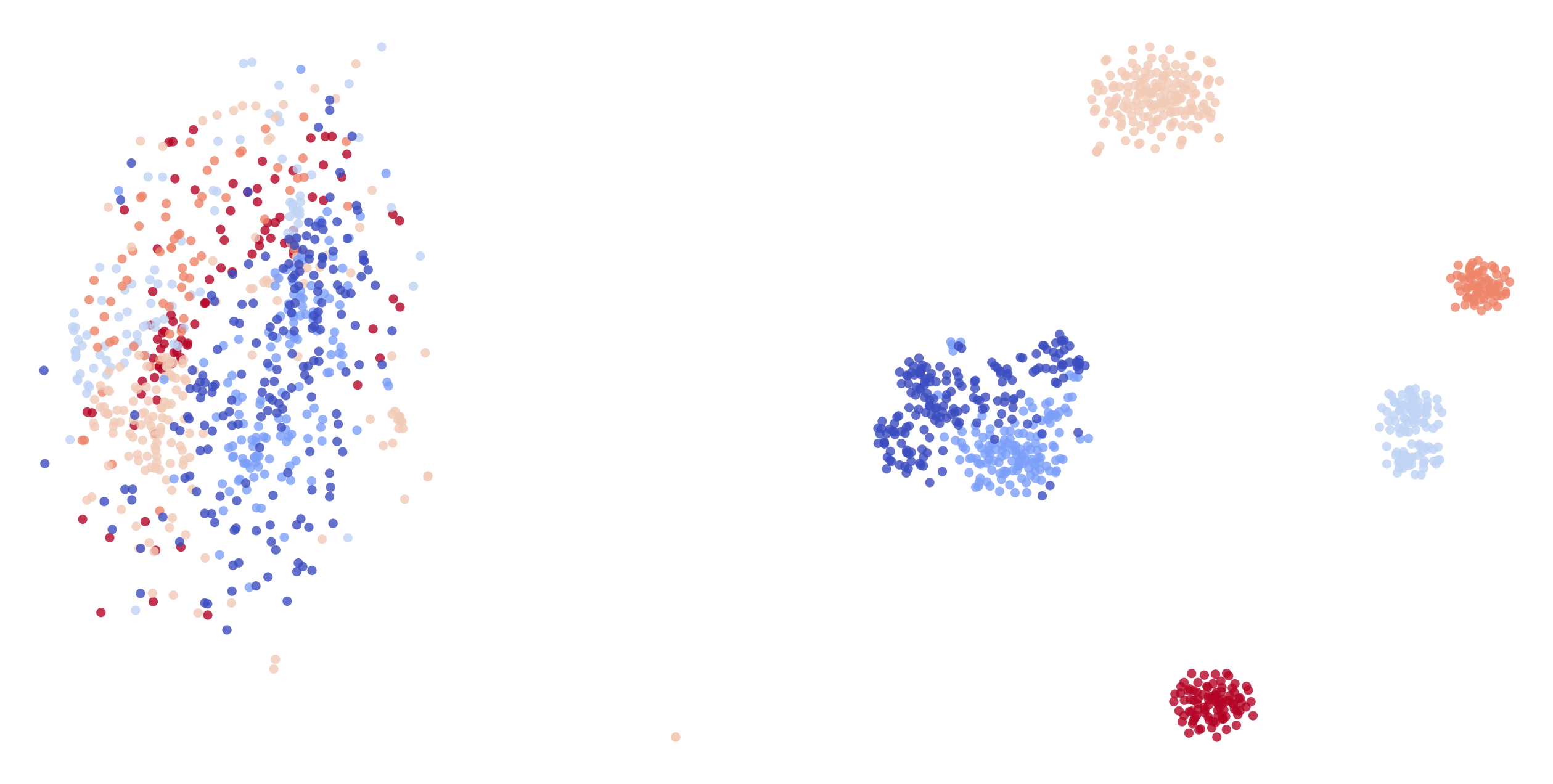}}
\subfloat[Chu2]{
		\includegraphics[scale=0.10]{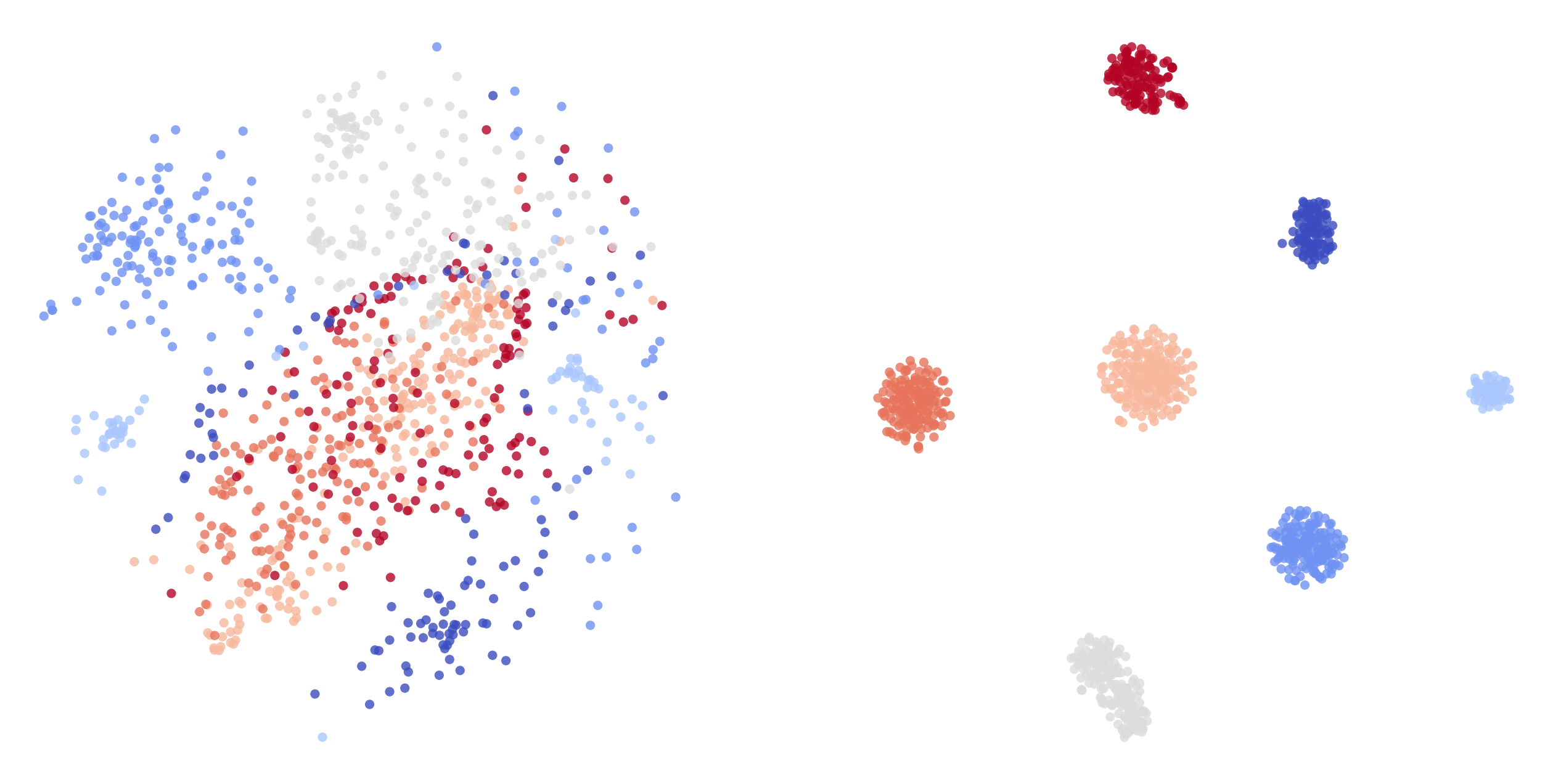}}
\subfloat[Han]{
		\includegraphics[scale=0.10]{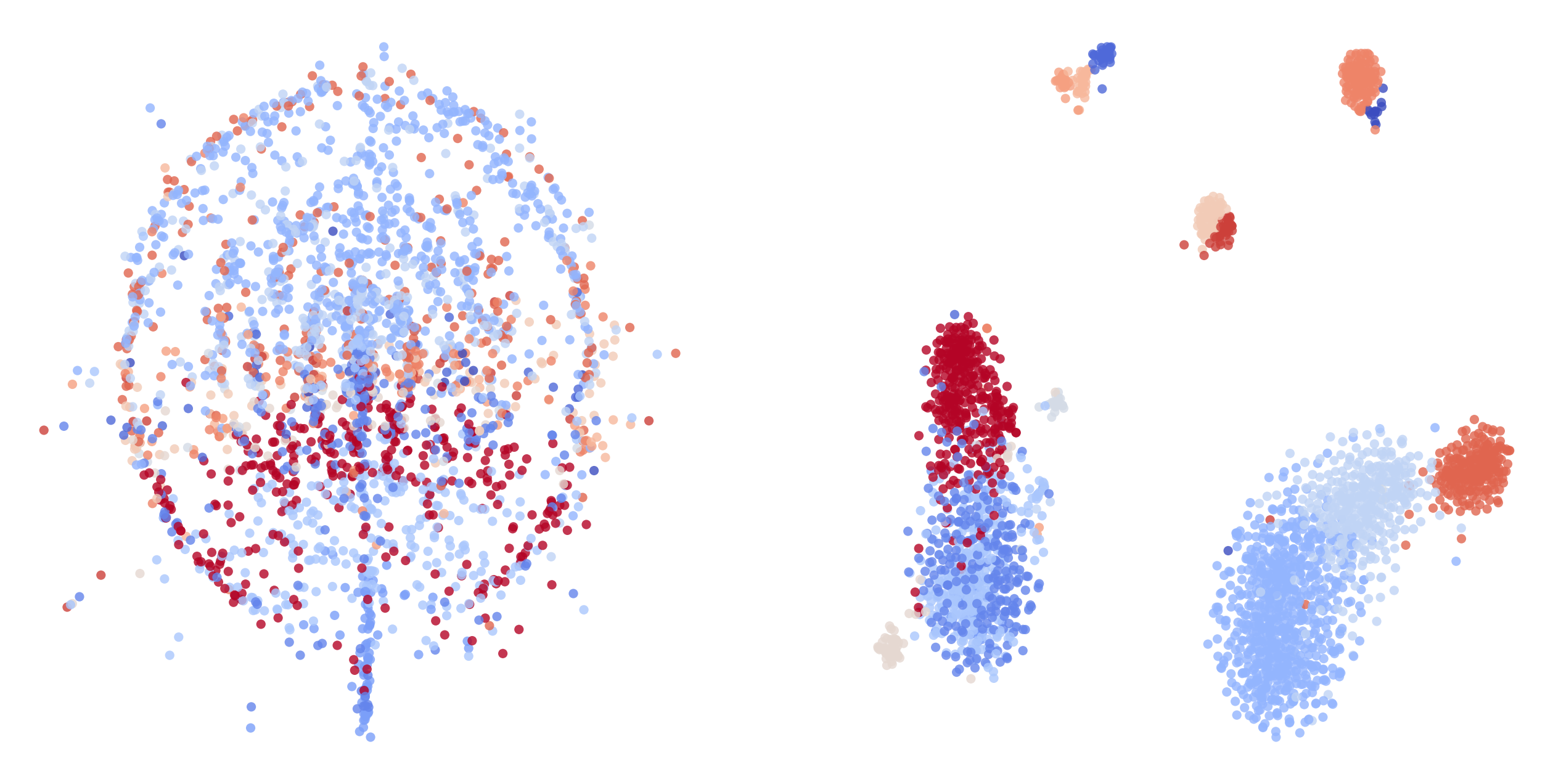}}
\\
\subfloat[CITE CBMC]{
		\includegraphics[scale=0.10]{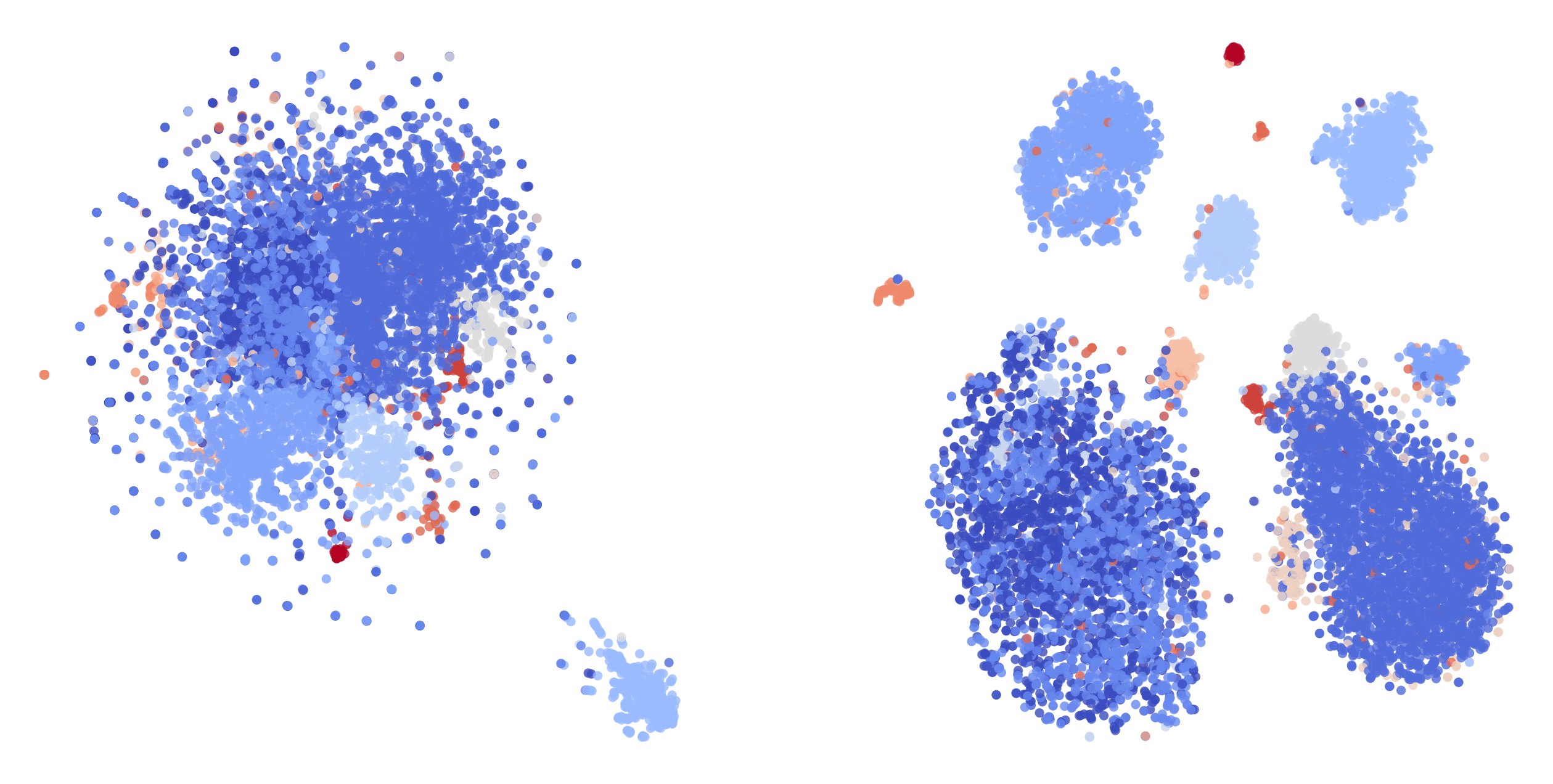}}
\subfloat[Human Pancreas1]{
		\includegraphics[scale=0.10]{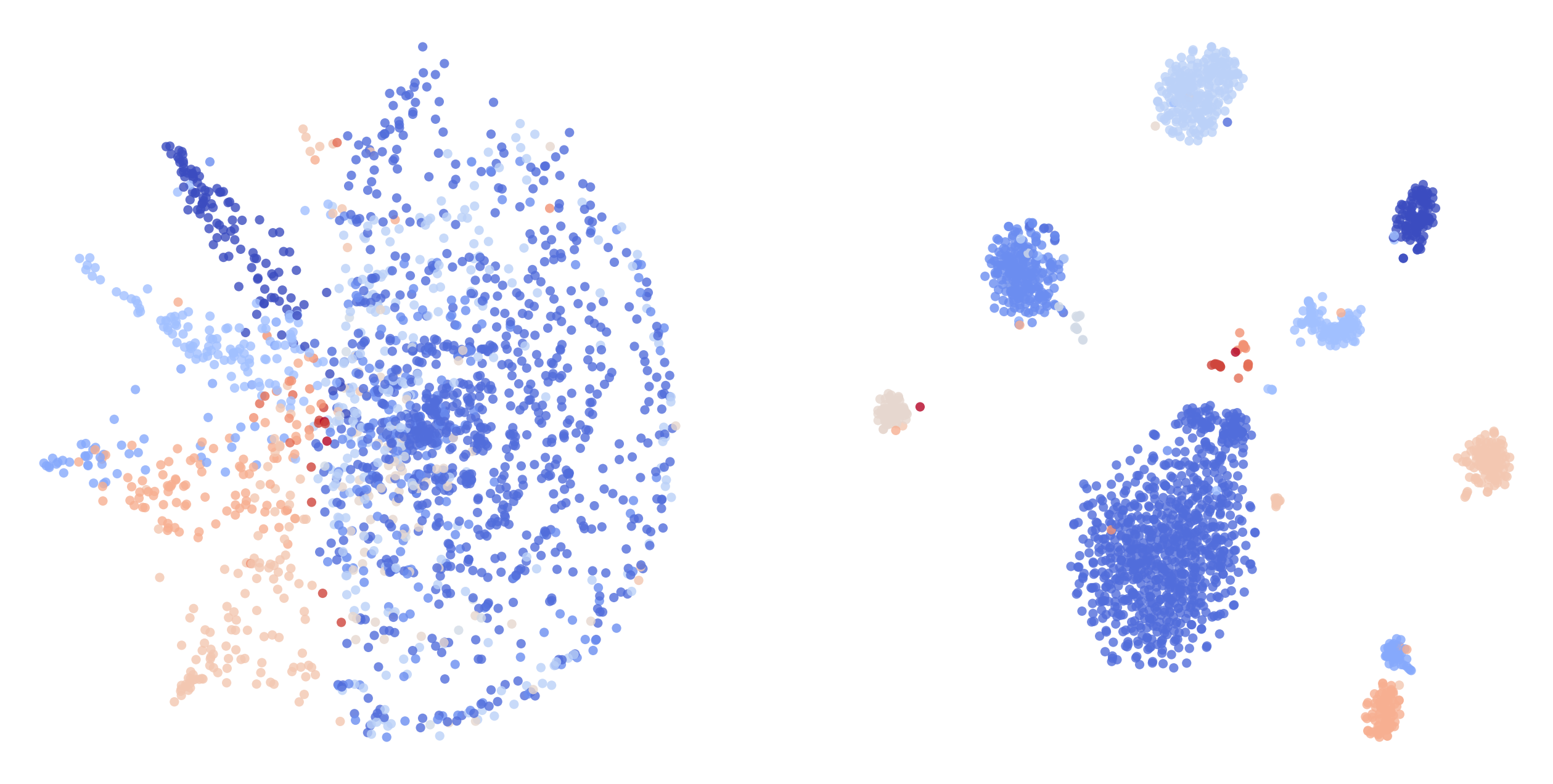}}
\subfloat[Human Pancreas2]{
		\includegraphics[scale=0.10]{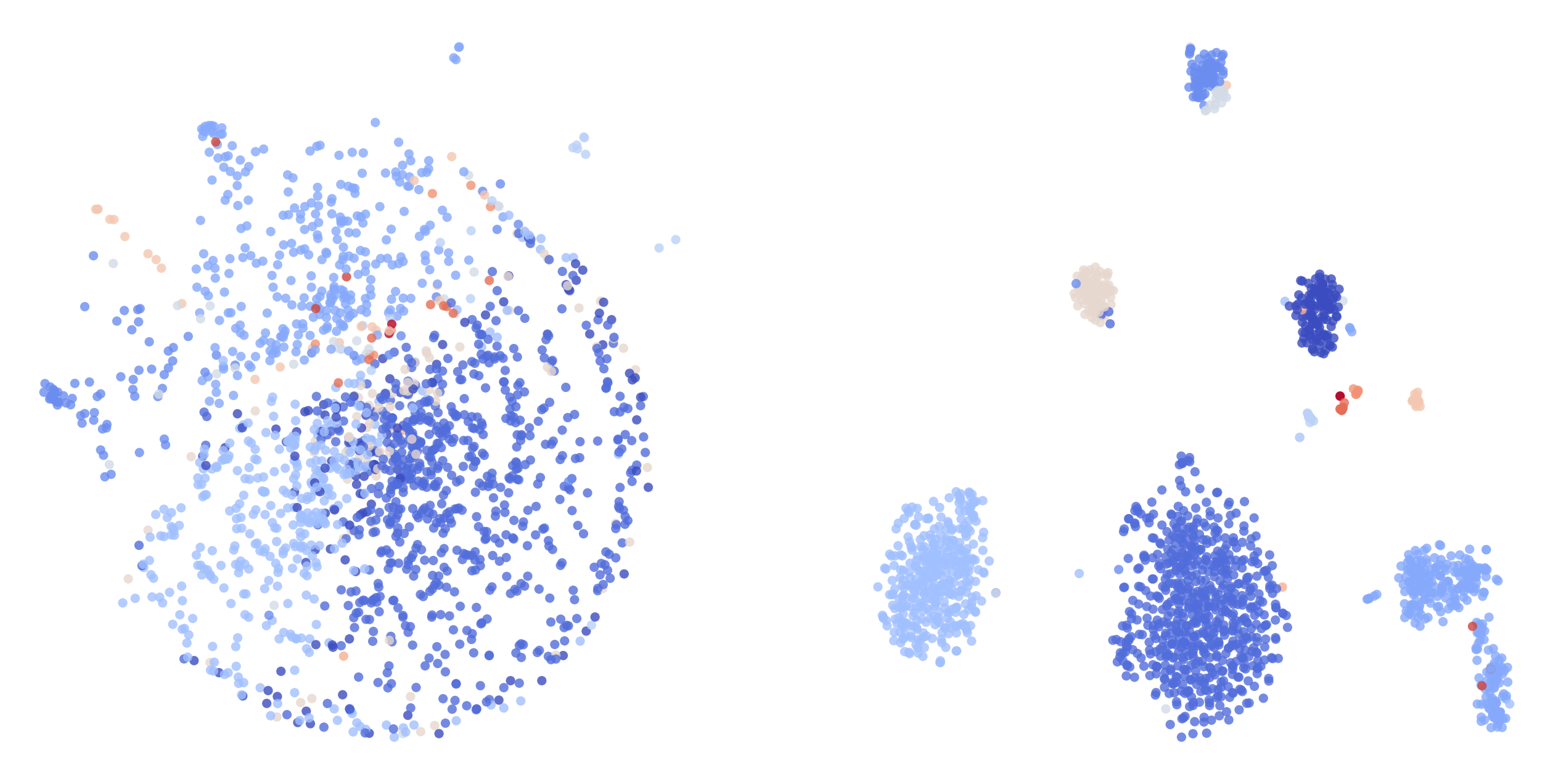}}
\subfloat[Human Pancreas3]{
		\includegraphics[scale=0.10]{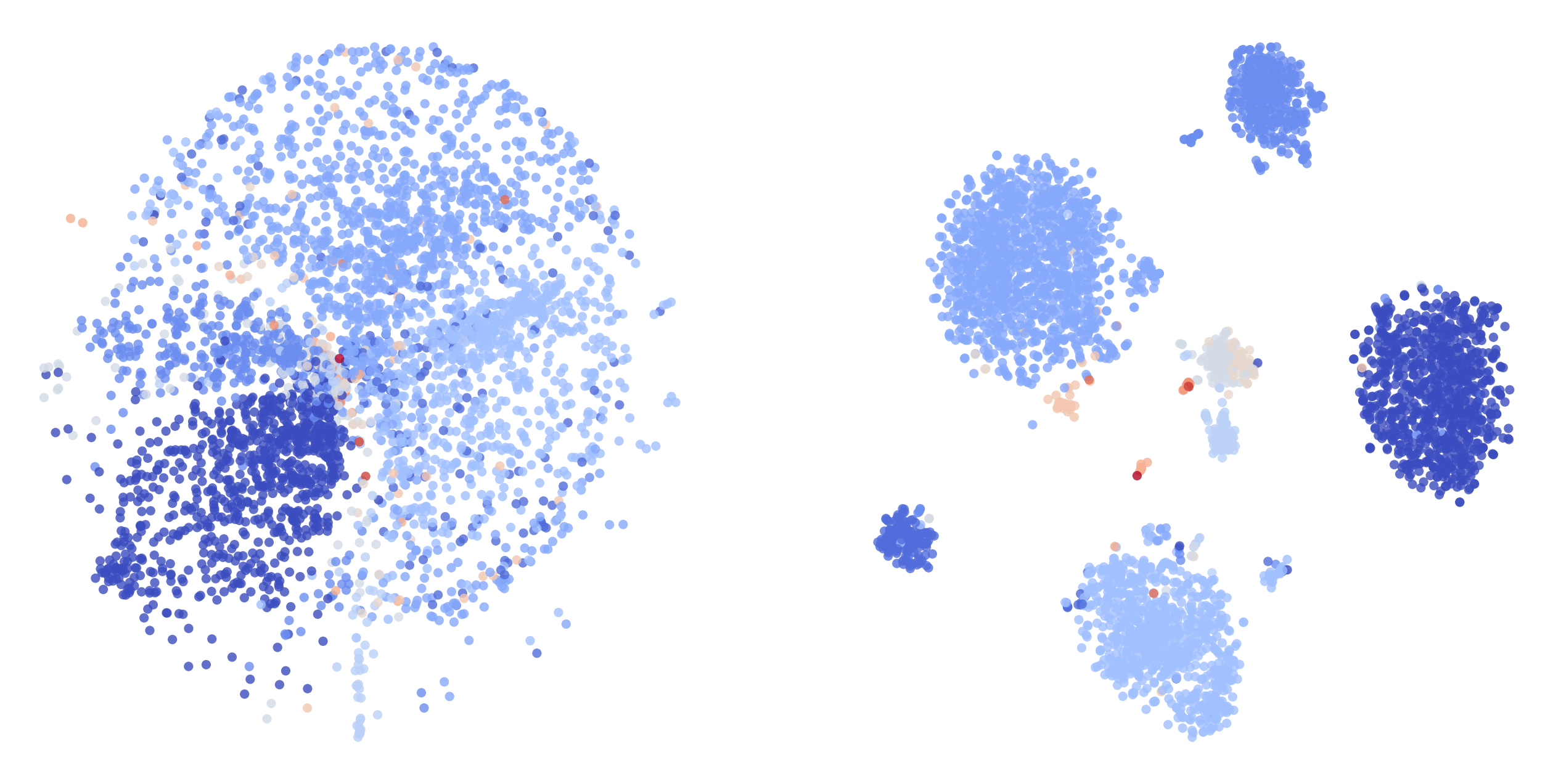}}
\\
\subfloat[Chung]{
		\includegraphics[scale=0.10]{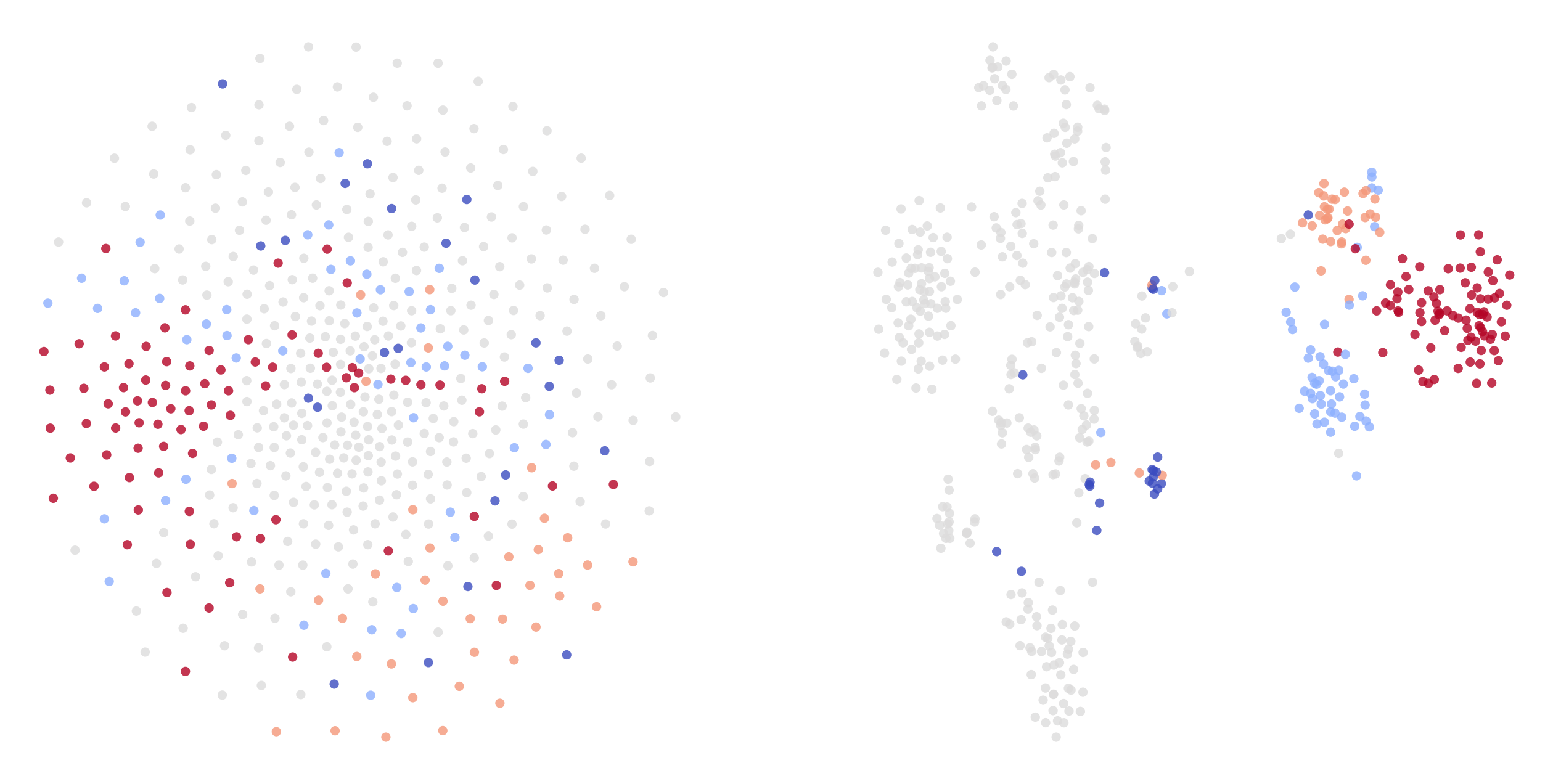}}
\subfloat[Darmanis]{
		\includegraphics[scale=0.10]{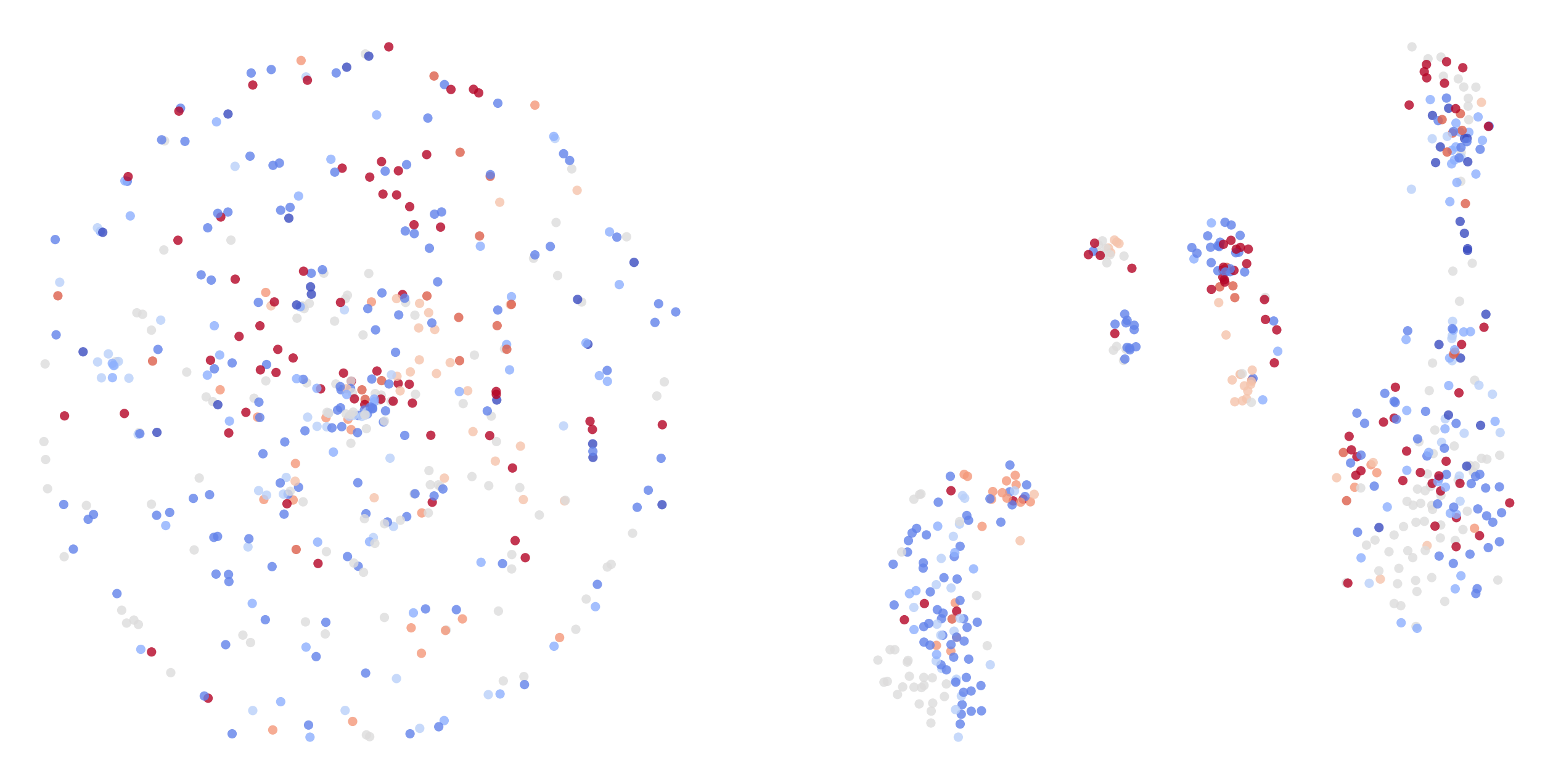}}
\subfloat[Engel]{
		\includegraphics[scale=0.10]{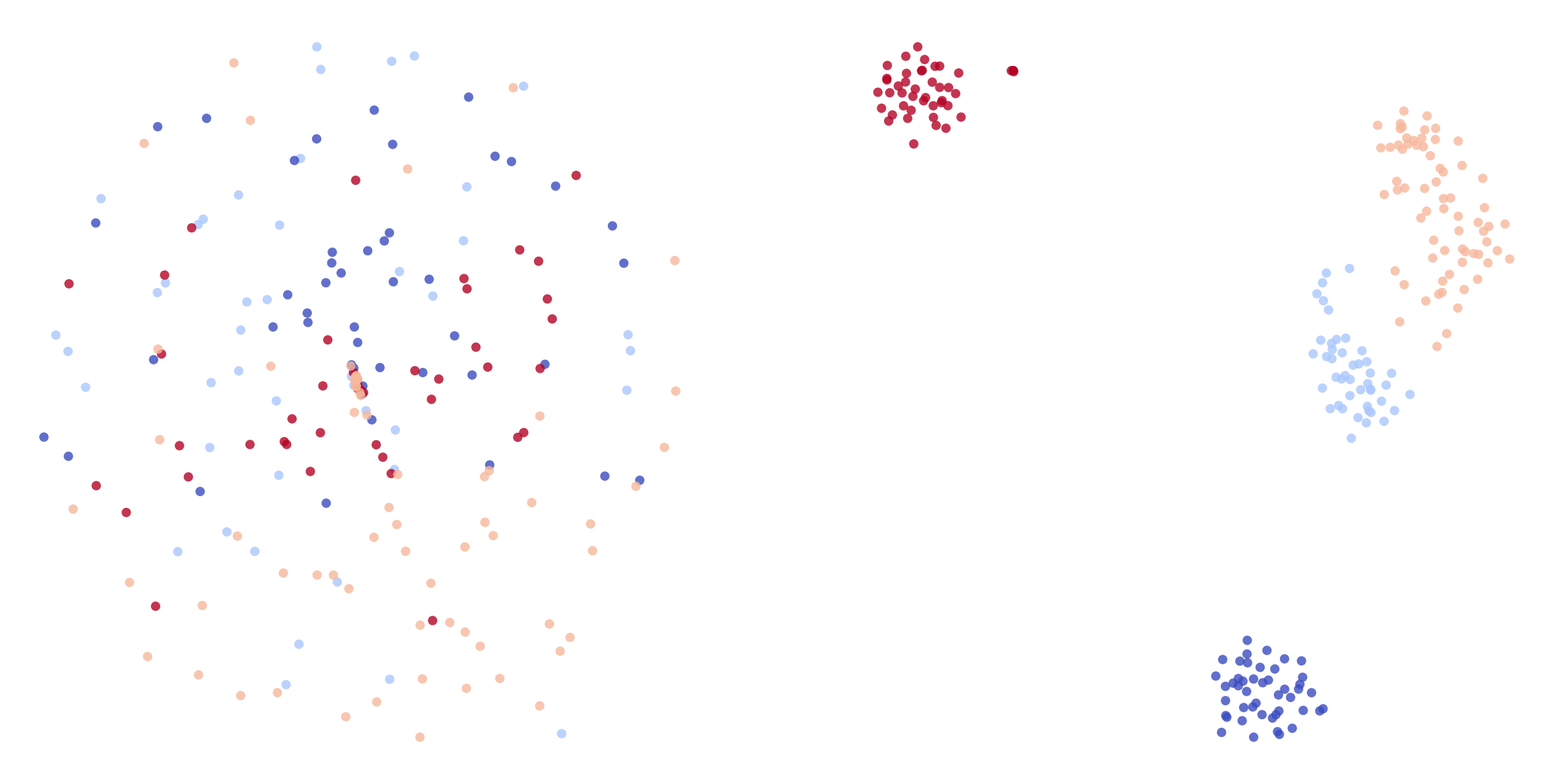}}
\subfloat[Goolam]{
		\includegraphics[scale=0.10]{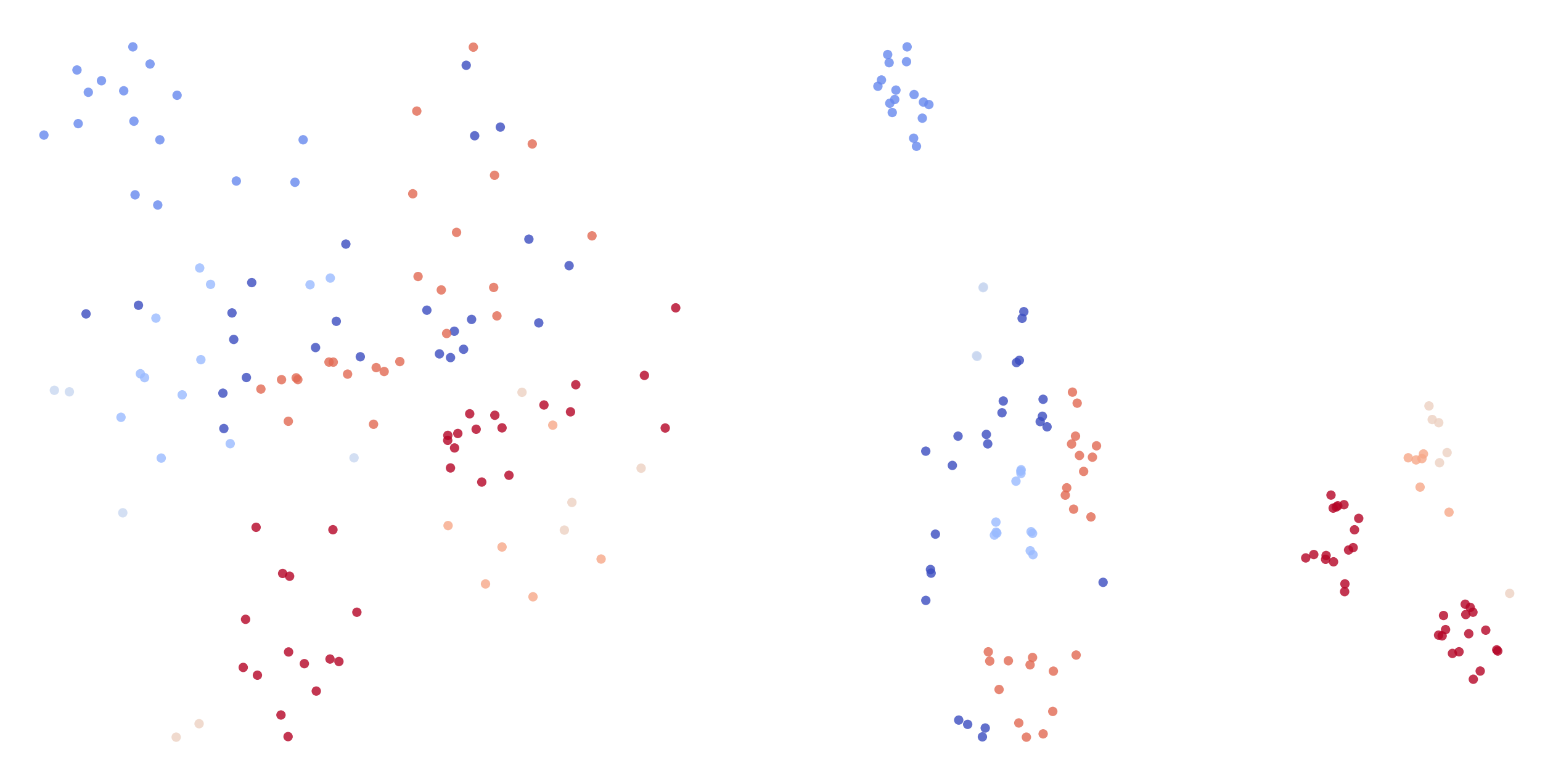}}
\\
\subfloat[Koh]{
		\includegraphics[scale=0.10]{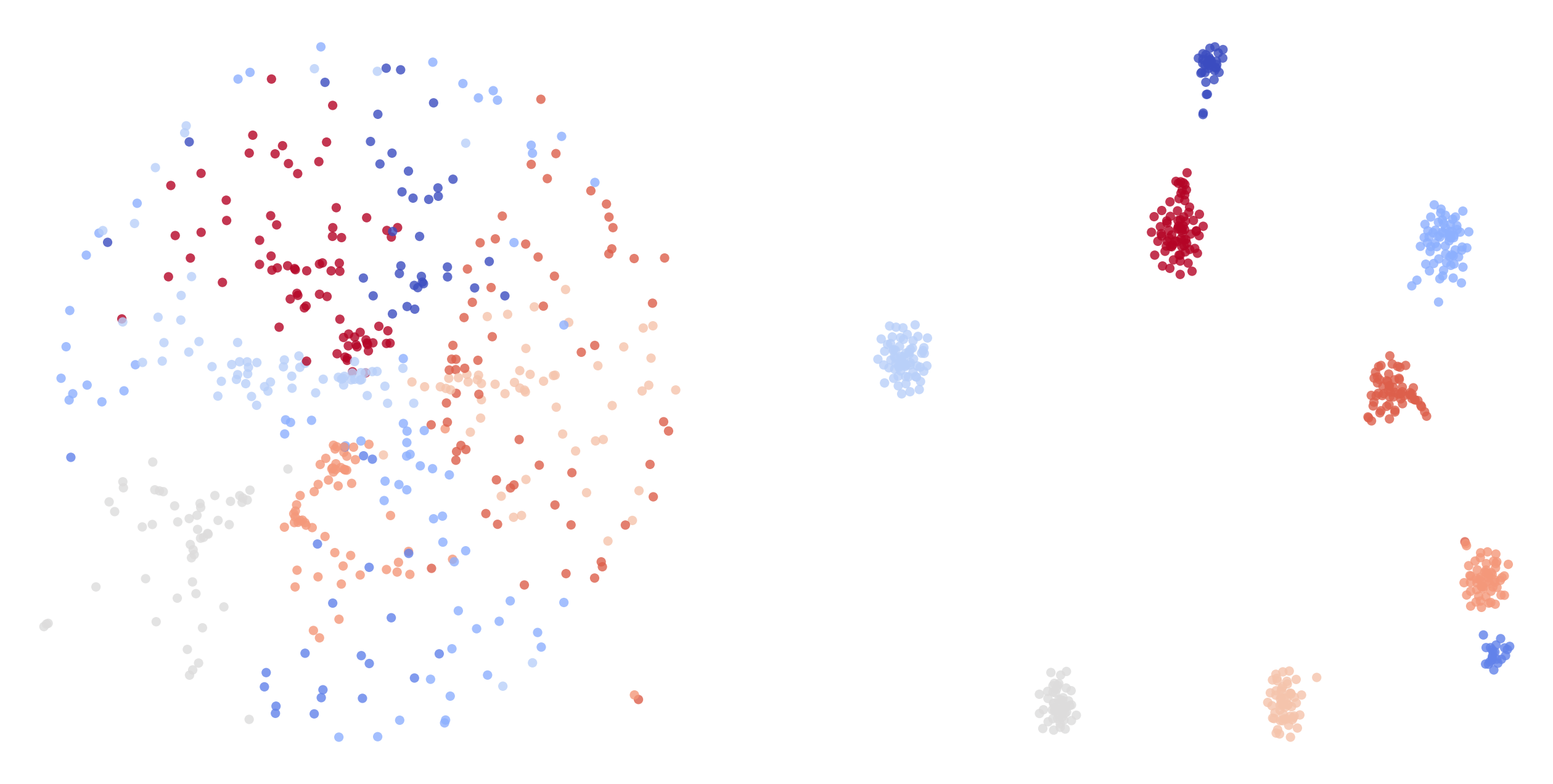}}
\subfloat[Kumar]{
		\includegraphics[scale=0.10]{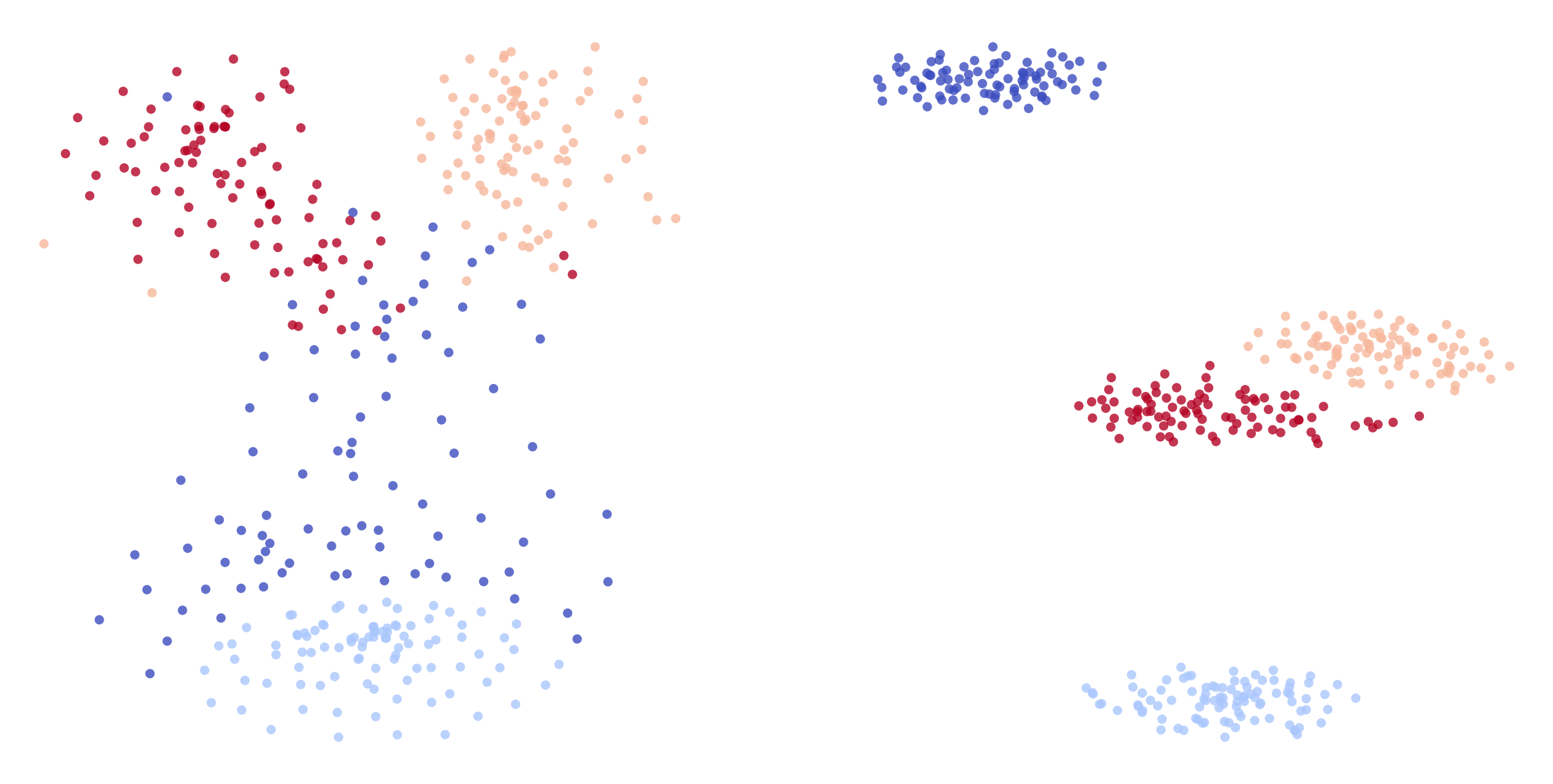}}
\subfloat[Leng]{
		\includegraphics[scale=0.10]{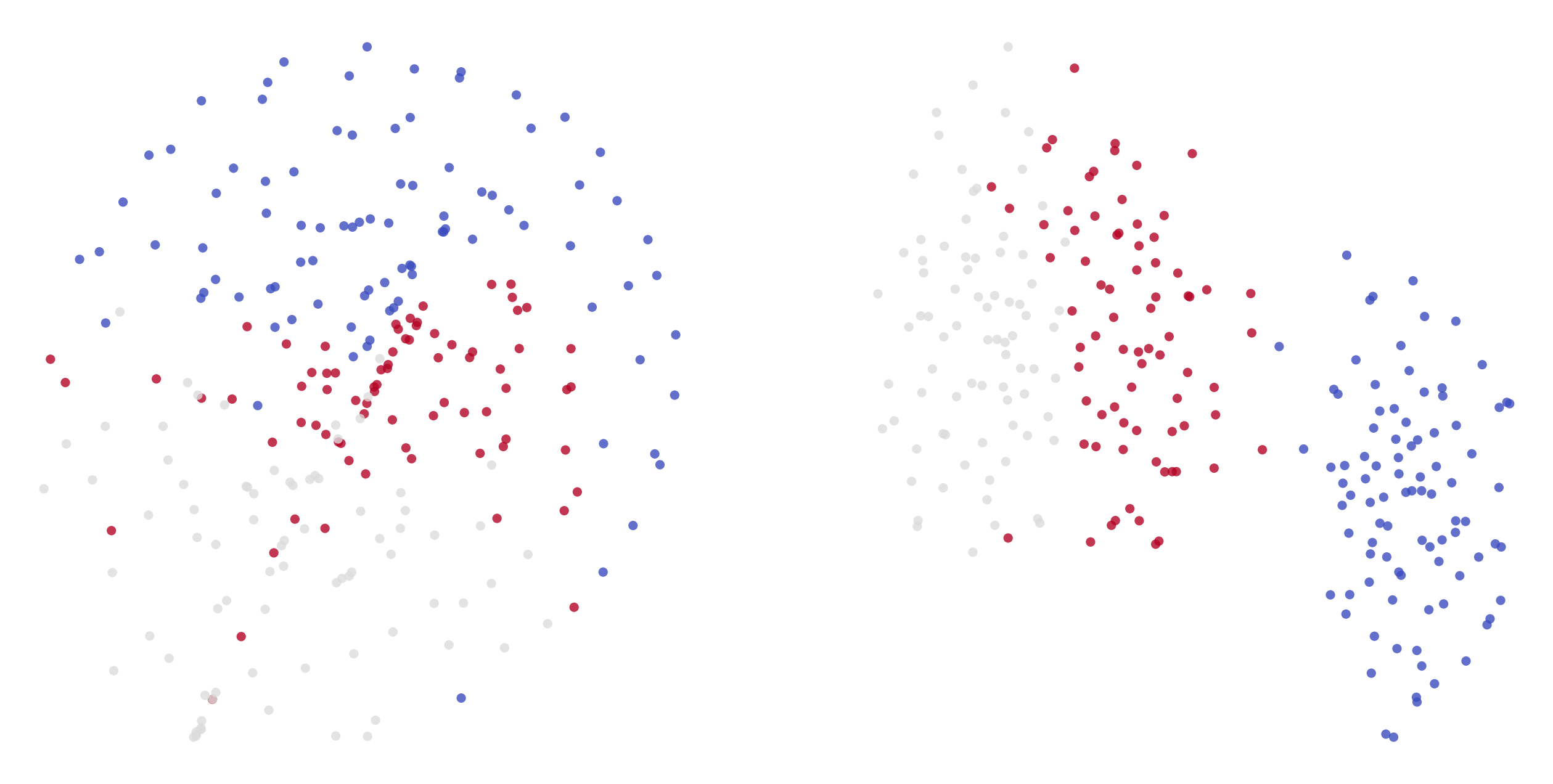}}
\subfloat[Li]{
		\includegraphics[scale=0.10]{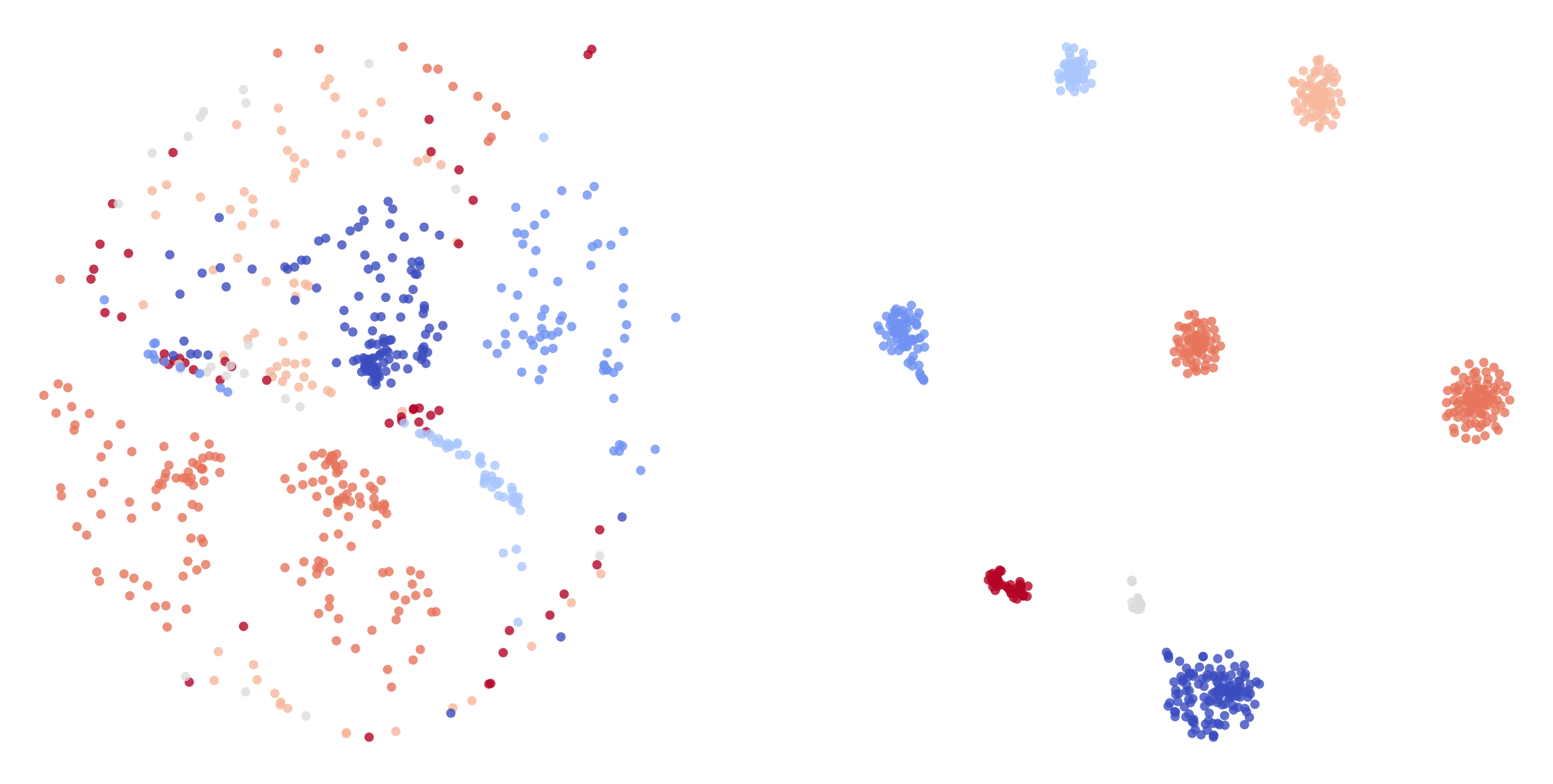}}
\\
\subfloat[Maria1]{
		\includegraphics[scale=0.10]{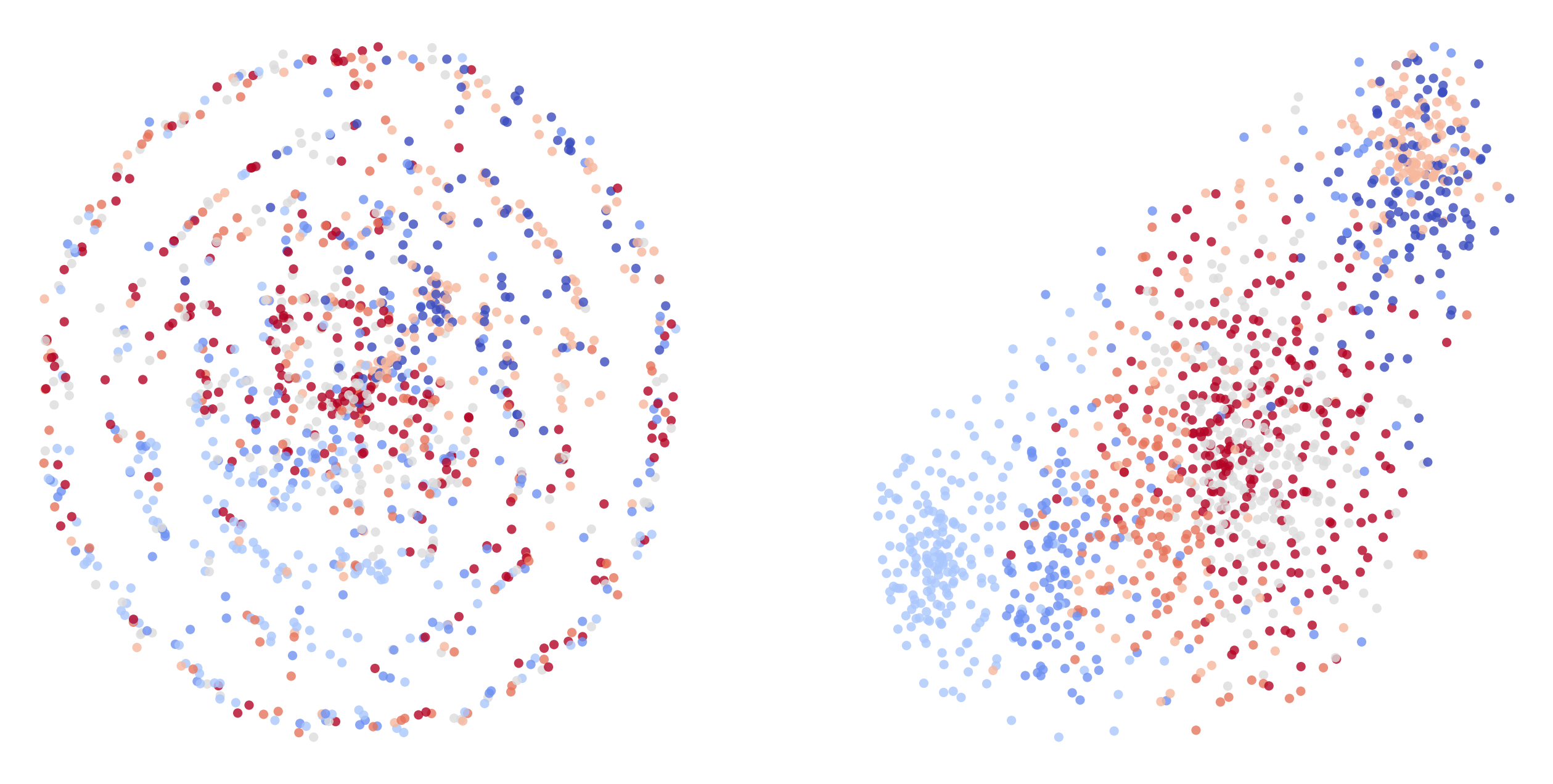}}
\subfloat[Maria2]{
		\includegraphics[scale=0.10]{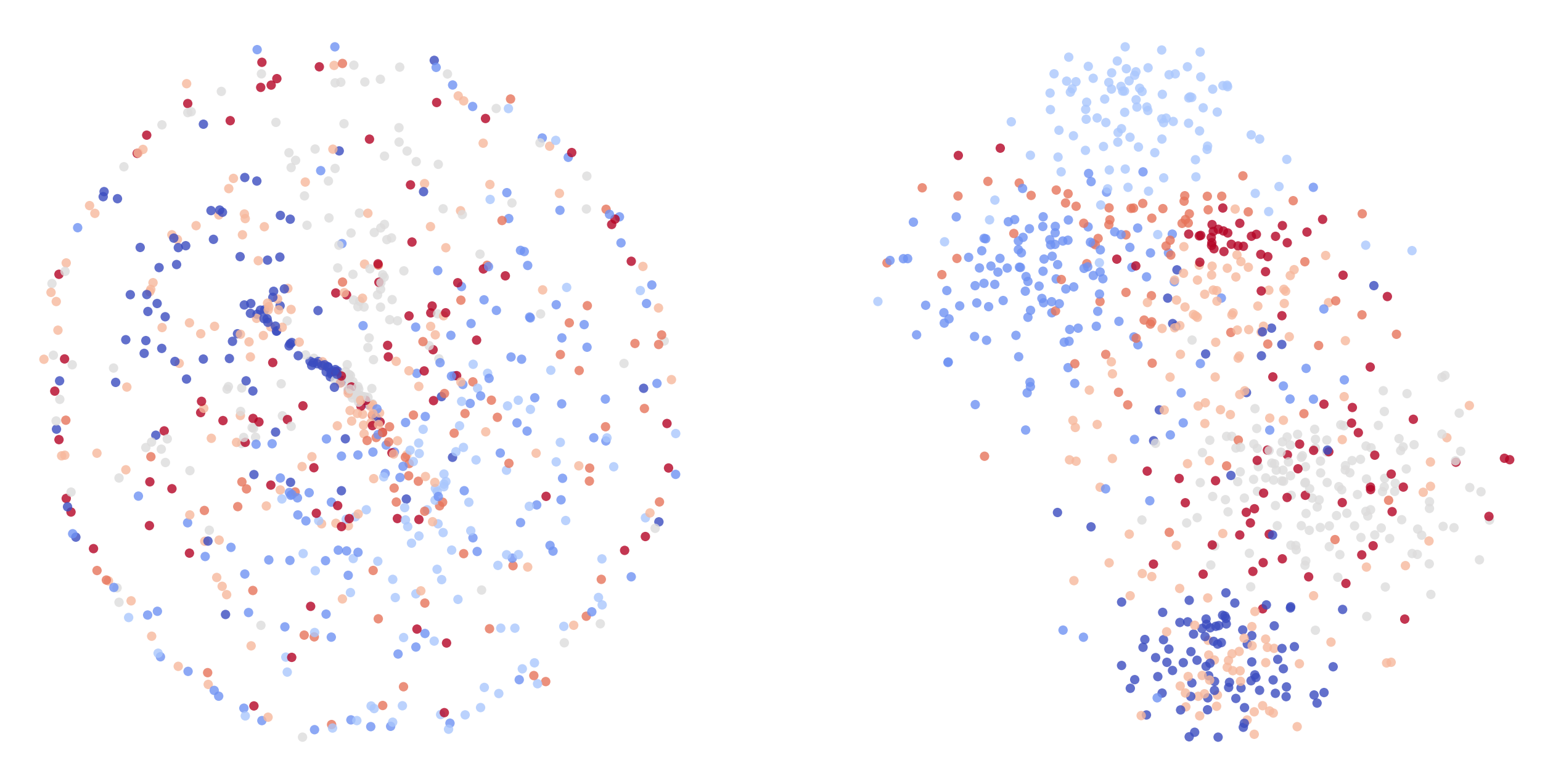}}
\subfloat[Mouse Pancreas1]{
		\includegraphics[scale=0.10]{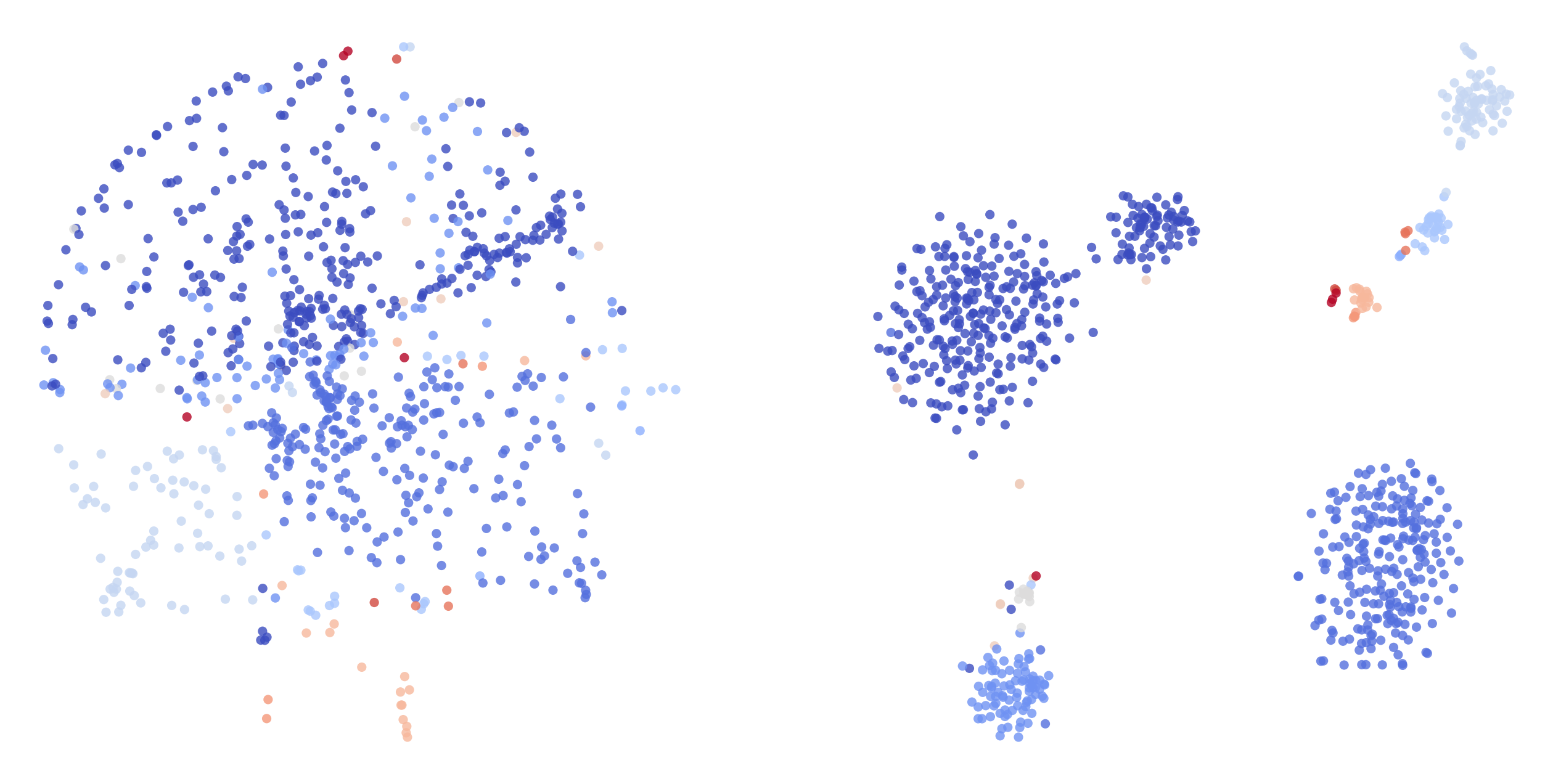}}
\subfloat[MacParland]{
		\includegraphics[scale=0.10]{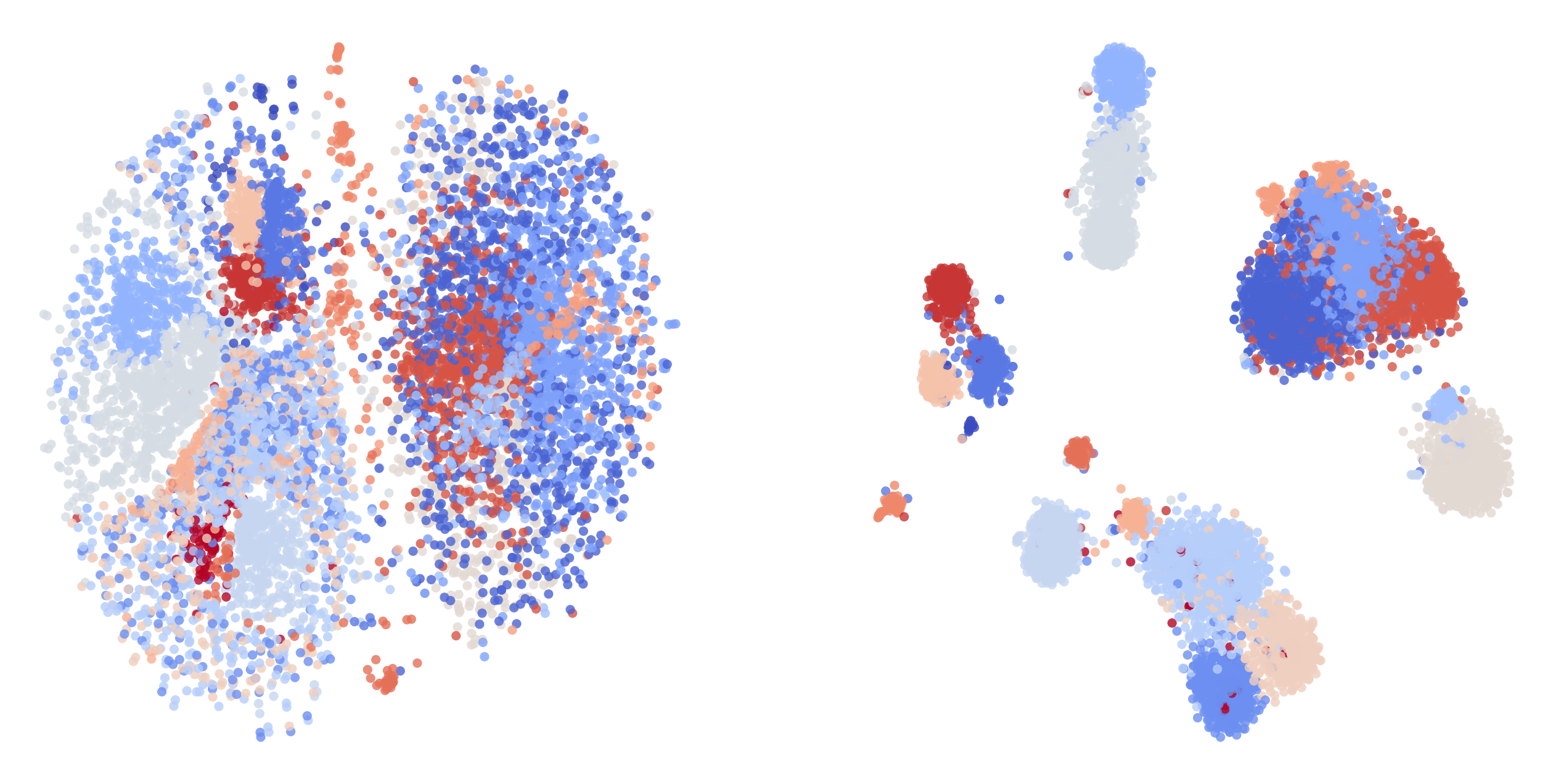}}
\\
\subfloat[Mouse Pancreas2]{
		\includegraphics[scale=0.10]{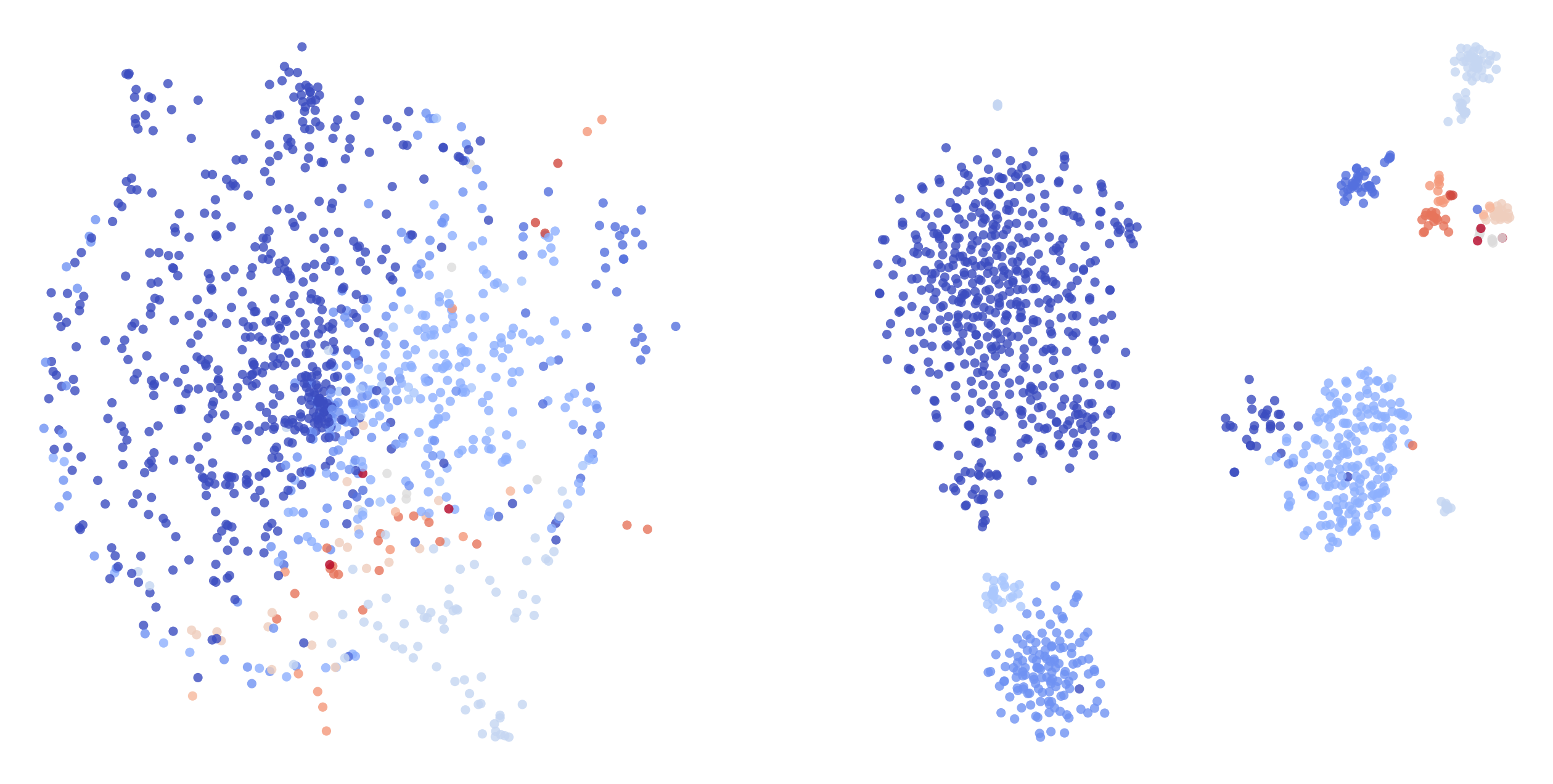}}
\subfloat[Robert]{
		\includegraphics[scale=0.10]{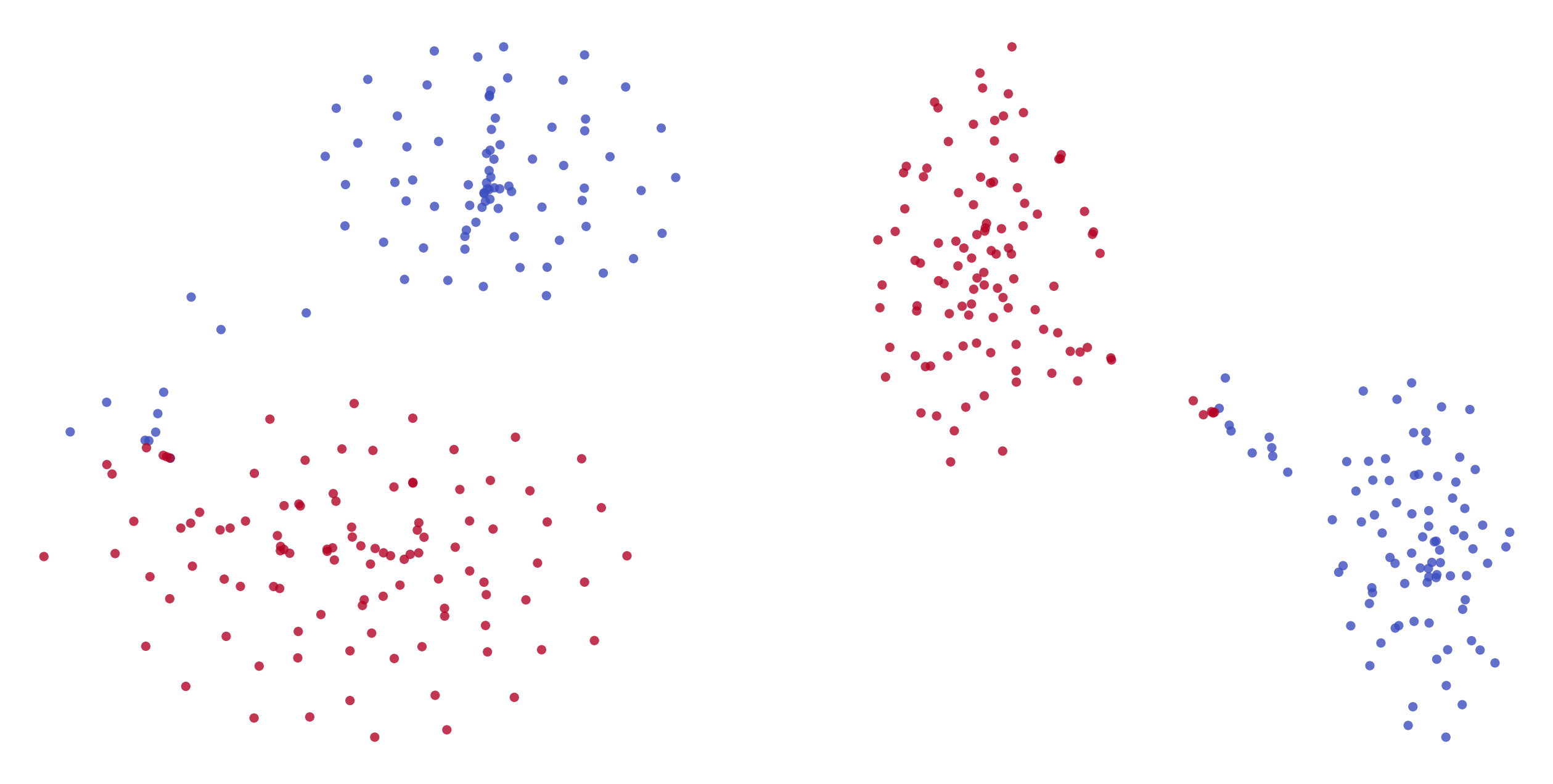}}
\subfloat[Ting]{
		\includegraphics[scale=0.10]{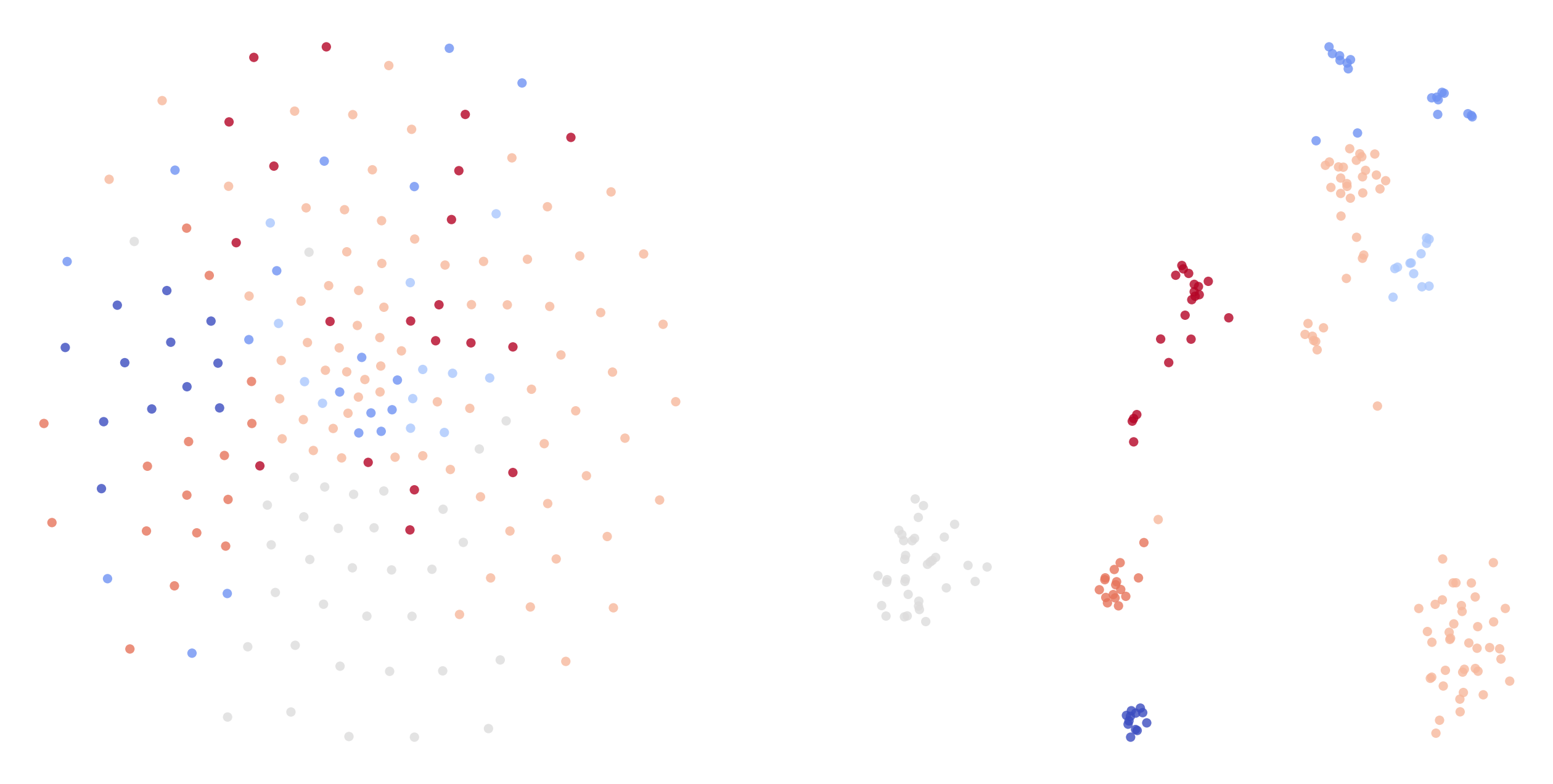}}
\subfloat[Yang]{
		\includegraphics[scale=0.10]{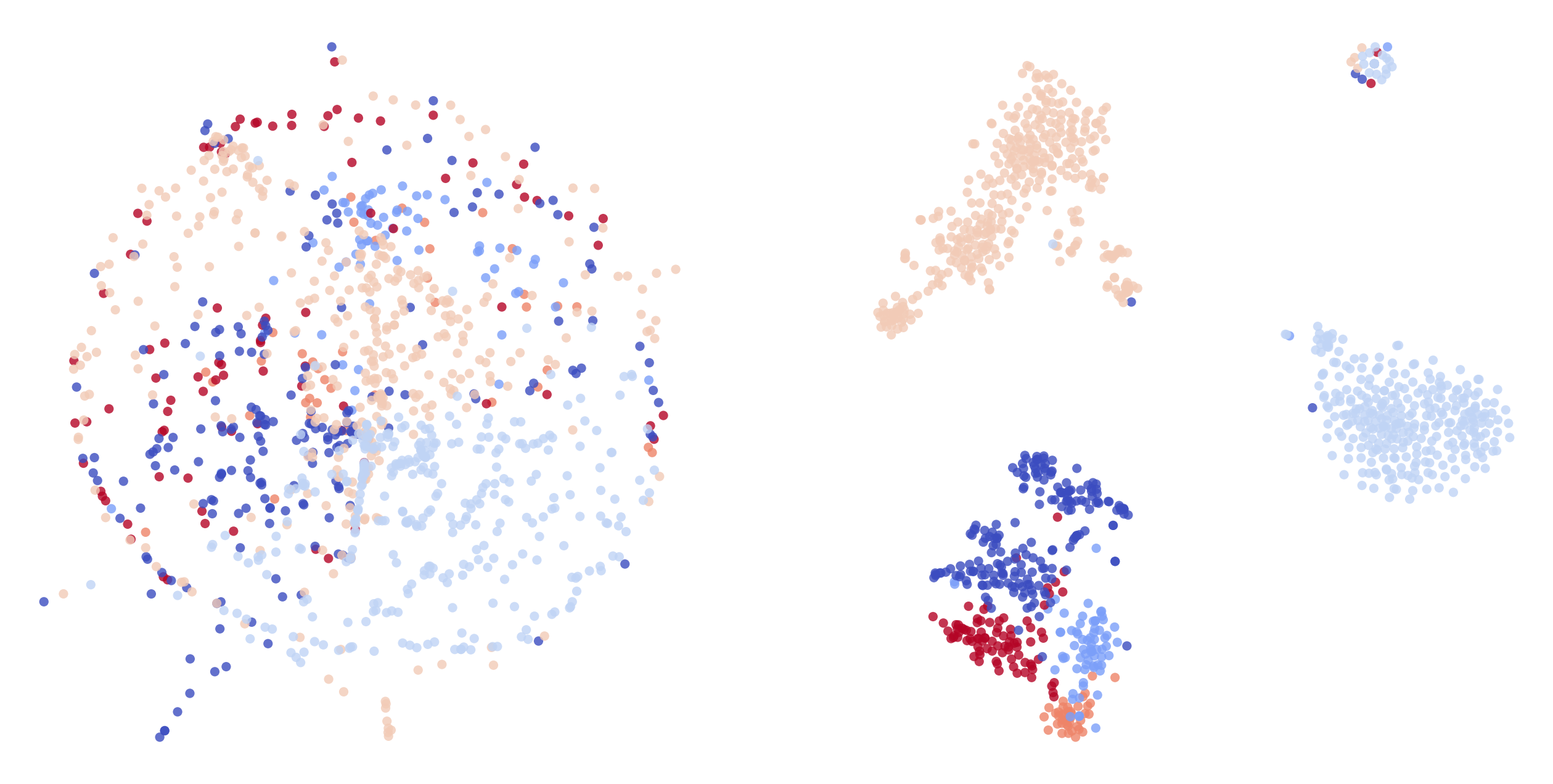}}
\caption{t-SNE visualization of the rest datasets, where the figure in the left panel is visualized from the original dataset.}
\label{tsne_vis}
\end{figure}

The t-SNE visualization and expression heatmap for the rest of the 24 datasets is shown in Figure~\ref{tsne_vis} and Figure~\ref{heat_map}, respectively.

% \clearpage
\subsection{Details of Experiment Settings}\label{exp_settings}
For more information on experiment settings, please refer to Appendix~\ref{dataset_des} (Dataset Description), Appendix~\ref{metrics} (Evaluation Metrics), Appendix~\ref{baselines} (Compared Methods) Appendix~\ref{hyper} (Hyperparameter Settings), and Appendix~\ref{platform} (Platform Setting).

\subsubsection{Dataset Description}\label{dataset_des}
Our research involved 25 single-cell RNA sequencing (scRNA-seq) datasets derived from various sequencing technologies and representing diverse biological conditions. 
These datasets were collected from several public databases
~\cite{edgar2002gene, brazma2003arrayexpress, leinonen2010sequence}
, including the National Center for Biotechnology Information's Gene Expression Omnibus (GEO), ArrayExpress, and the Sequence Read Archive (SRA), etc.
The "Cao" dataset was procured from a study utilizing the sci-RNA-seq method (single-cell combinatorial indexing RNA sequencing), as detailed in the publication by Cao et al.~\cite{Cao}. 
The "Han" dataset originates from the Mouse Cell Atlas, as published by Han et al.~\cite{Han}.
The "10X PBMC" focusing on peripheral blood mononuclear cells was acquired from the 10X Genomics website~\cite{10xpbmc}. 
Detailed specifics, including each dataset's origins, description, and size of cells and genes, are provided in Table~\ref{tab:dataset_detailed_information}. 
% \section{Detailed information of the datasets}
\begin{table}[ht]
\centering
\caption{Detailed information of the datasets used in this study}
\label{tab:dataset_detailed_information}
\resizebox{0.95\textwidth}{!}{%
\begin{tabular}{ccccccc}
\toprule
Dataset & Size & \#Cells & \#Genes & \#Types & Accession & Description  \\ \midrule
\rowcolor[HTML]{EFEFEF} 
{\color[HTML]{333333} Chu1} & S & {\color[HTML]{333333} 758} & {\color[HTML]{333333} 19176} & {\color[HTML]{333333} 6} & {\color[HTML]{333333} GSE75748} & {\color[HTML]{333333} human pluripotent stem cells}  \\
Chung & S & 515 & 20345 & 5 & GSE75688 & human tumor and immune cells \\
\rowcolor[HTML]{EFEFEF} 
Darmanis & S & 466 & 22085 & 9 & GSE67835 & human brain cells \\
Engel & S & 203 & 23337 & 4 & GSE74596 & mouse Natural killer T cells \\
\rowcolor[HTML]{EFEFEF} 
Goolam & S & 124 & 41388 & 8 & E-MTAB-3321 & mouse cells from different stages\\
Koh & S & 498 & 60483 & 9 & GSM2257302 & human embryonic stem cells \\
\rowcolor[HTML]{EFEFEF} 
Kumar & S & 361 & 22394 & 4 & GSE60749 & mouse embryonic stem cells \\
Leng & S & 247 & 19084 & 3 & GSE64016 & human embryonic stem cells \\
\rowcolor[HTML]{EFEFEF} 
Li & S & 561 & 57241 & 7 & GSE81861 & human cell lines \\
Maria2 & S & 759 & 33694 & 7 & GSE124731 & human innate T cells \\
\rowcolor[HTML]{EFEFEF} 
Robert & S & 194 & 23418 & 2 & GSE74923 & \begin{tabular}[c]{@{}c@{}}mouse leukemia cell line\\ and primary CD8+ T-cells\end{tabular} \\
Ting & S & 187 & 21583 & 7 & GSE51372 & mouse circulating tumor cells \\
\rowcolor[HTML]{EFEFEF} 
\begin{tabular}[c]{@{}c@{}}Mouse\\ Pancreas1\end{tabular} & S & 822 & 14878 & 13 & GSE84133 & mouse bladder cells \\ \midrule
Cao &L & 4186 & 13488 & 10 & \begin{tabular}[c]{@{}c@{}}sci-RNA-seq\\ platform\end{tabular} & worm neuron cells \\
\rowcolor[HTML]{EFEFEF} 
Chu2 &L & 1018 & 19097 & 7 & GSE75748 & human pluripotent stem cells \\
Han & L & 2746 & 20670 & 16 & \begin{tabular}[c]{@{}c@{}}Mouse Cell\\ Atlas project\end{tabular} & mouse bladder cells \\
\rowcolor[HTML]{EFEFEF} 
MacParland & L & 14653 & 5000 & 11 & GSE115469 & human liver cells \\
Maria1 & L & 1277 & 33694 & 7 & GSE124731 & human innate T cells \\
\rowcolor[HTML]{EFEFEF} 
Puram & L & 3363 & 23686 & 8 & GSE103322 & \begin{tabular}[c]{@{}c@{}}non-malignant cells\\ in Head and Neck Cancer\end{tabular} \\
Yang & L & 1119 & 46609 & 6 & GSE90848 & \begin{tabular}[c]{@{}c@{}}mouse bulge hair follicle stem cell, \\ hair germ, basal transient amplifying \\ cells (TACs) and dermal papilla\end{tabular} \\
\rowcolor[HTML]{EFEFEF} 
\begin{tabular}[c]{@{}c@{}}CITE\\ CBMC\end{tabular} & L & 8617 & 2000 & 15 & GSE108097 & \begin{tabular}[c]{@{}c@{}}mouse peripheral blood \\ mononuclear cells\end{tabular} \\
\begin{tabular}[c]{@{}c@{}}Human\\ Pancreas1\end{tabular} & L & 1937 & 20125 & 14 & GSE84133 & Human Pancreas cells \\
\rowcolor[HTML]{EFEFEF} 
\begin{tabular}[c]{@{}c@{}}Human\\ Pancreas2\end{tabular} & L & 1724 & 20125 & 14 & GSE84133 & Human Pancreas cells \\
\begin{tabular}[c]{@{}c@{}}Human\\ Pancreas3\end{tabular} & L & 3605 & 20125 & 14 & GSE84133 & Human Pancreas cells \\
\rowcolor[HTML]{EFEFEF} 
\begin{tabular}[c]{@{}c@{}}Mouse\\ Pancreas2\end{tabular} & L & 1064 & 14878 & 13 & GSE84133 & mouse bladder cells \\ \bottomrule
\end{tabular}%
}
\begin{tablenotes}
    \small
    \item * Larger datasets reflect the true distribution of single-cell types more than smaller datasets used for specific research purposes, but they are also more imbalanced. So, we divide all datasets into small and large datasets using 1000 cells as the limit. A size of "S" means that the dataset is small, and "L" means that the dataset is large.
    \end{tablenotes}
    \vspace{-0.4cm}
\end{table}

\subsubsection{Evaluation Metrics.}\label{metrics}
To compare the performance of these methods, we evaluate the cell-type-discriminating performance of genes via cell clustering. 
We follow the same setting as CellBRF~\cite{CellBRF} by adopting a graph-based Louvain community detection algorithm in Seurat~\cite{Seurat} as the downstream cell clustering model, a commonly used software toolkit for scRNA-seq clustering. 
We adopted three widely used metrics to assess the model performance, including normalized mutual information (NMI)~\cite{NMI}, adjusted rand index (ARI)~\cite{ARI}, and silhouette index (SI)~\cite{SI}. 
All metrics range from 0 to 1, where the higher the value, the better the model performance. 

% \subsection{Baseline Methods}\label{baselines}

\subsubsection{Baselines and Ablation Variations.}\label{baselines}
Our comparative analysis evaluated \model\ against seven widely used baselines, categorized into gene panel selection and feature selection methods. 
The detailed descriptions are listed as follows:
(1) \textbf{K-Best}~\cite{KBest} selects the top K features based on their scores, providing a straightforward approach to feature prioritization; 
(2) \textbf{mRMR}~\cite{mRMR} chooses features that maximize relevance to the target variable while minimizing redundancy among the features; 
(3) \textbf{LASSO}~\cite{LASSO} employs regularization to shrink the coefficients of less useful features to zero, effectively performing feature selection during model fitting;
(4) \textbf{RFE}~\cite{rfe}  is a recursive feature elimination method that systematically removes the weakest features based on model performance until a specified number of features remains;  
(5) \textbf{HRG}~\cite{HRG} utilizes a graph-based approach to identify genes that exhibit regional expression patterns within a cell-cell similarity network;
(6) \textbf{geneBasis}~\cite{geneBasis} aims to select a small, targeted panel of genes from scRNA-seq datasets that can effectively capture the transcriptional variability present across different cells and cell types; 
(7) \textbf{CellBRF}~\cite{CellBRF}, selects the most significant gene subset evaluated using Random Forest.
HRG, CellBRF, and geneBasis are commonly used gene panel selection methods. 
We provide a pseudo-pre-clustering label for all feature selection methods rather than the golden label to maintain the consistency of the gene panel selection problem setting.

\subsubsection{Hyperparameter Setting and Reproducibility}\label{hyper}
For all experiments and datasets, we ran 400 epochs for exploration and optimization. The memory size is set to 400. 
The basic methods for the gene pre-filtering module consist of Random Forest, SVM, RFE, geneBasis, and KBest.
We adopt the Louvain community detection algorithm to generate pseudo-labels for feature selection methods and reward estimation, as same as the downstream clustering method. 
The gene state representation component consists of an autoencoder, which includes two structurally mirrored three-layer feed-forward networks. 
The first network serves as the encoder, with the first layer containing 256 hidden units, the second layer containing 128 hidden units, and the third layer containing 64 hidden units, progressively compressing the data to capture its intrinsic features. 
The second network acts as the decoder, mirroring the encoder structure. 
The first layer contains 64 hidden units, the second layer contains 128 hidden units, and the final layer contains 256 hidden units, all of which aim to reconstruct the original input data from the compressed representation. 
This symmetric design of the networks enables the autoencoder to effectively learn a low-dimensional representation of the data while minimizing reconstruction error.
The training epochs in each step for the gene subset state representation component are set to 10. 
For the knowledge injection setting, we adopt the gene subsets selected by CellBRF, geneBasis, HRG, mRMR, and KBest as our prior knowledge.
In the reinforcement iteration, we set each gene agent's actor and critic network as a two-layer neural network with 64 and 8 hidden sizes in the first and second layers, respectively. 
According to the hyperparameter study, we set alpha (the trade-off between spatial coefficient and quantity suppression in  Equation~\ref{reward_overall_func}) to 0.5, so each part of the reward function has a balanced weight.
To train the policy network in each gene agent, we set the minibatch size to 32 and used the Adam optimizer with a learning rate of 0.005. 
The parameter settings of all baseline models follow the corresponding papers.

\subsubsection{Experiment Platform Settings}\label{platform} All experiments were ran on the Ubuntu 18.04.6 LTS operating system, Intel(R) Xeon(R) Gold 6338 CPU, and 4 NVIDIA V100 GPUs, with the framework of Python 3.11.5 and PyTorch 2.1.1~\cite{pytorch}.

\clearpage

% \section{Detailed of the Overall Comparision}
% \usepackage{booktabs}
% \usepackage{multirow}
% \usepackage{graphicx}

\useunder{\uline}{\ul}{}
\begin{table}[htbp]
\centering
\caption{Details of the model performance comparison on each dataset regarding NMI, ARI, and SI. We use light red shade and \textbf{bold font} to highlight the best performance. We use light blue shade and {\ul underline} to highlight the second-best performance.}
\label{tab:main_table}
\resizebox{\columnwidth}{!}{%
\begin{tabular}{ccccccccccccc}
\toprule
 & \multicolumn{3}{c}{KBest} & \multicolumn{3}{c}{mRMR} & \multicolumn{3}{c}{LASSO} & \multicolumn{3}{c}{RFE} \\
\multirow{-2}{*}{Dataset} & NMI & \cellcolor[HTML]{EFEFEF}ARI & SI & \cellcolor[HTML]{EFEFEF}NMI & ARI & \cellcolor[HTML]{EFEFEF}SI & NMI & \cellcolor[HTML]{EFEFEF}ARI & SI & \cellcolor[HTML]{EFEFEF}NMI & ARI & \cellcolor[HTML]{EFEFEF}SI \\ \midrule
Chu1 & 80.01 & 66.99 & \cellcolor[HTML]{CBCEFB}{\ul 17.04} & 81.80 & 71.30 & 11.95 & 69.07 & 4.77 & 5.97 & 80.22 & 68.65 & 12.56 \\
Chung & \cellcolor[HTML]{FFCCC9}\textbf{49.01} & 18.42 & 15.38 & 46.63 & 17.90 & 15.76 & 45.23 & 7.29 & 7.15 & \cellcolor[HTML]{CBCEFB}{\ul 48.76} & 18.30 & 12.48 \\
Darmanis & 17.24 & 4.75 & 9.20 & \cellcolor[HTML]{CBCEFB}{\ul 18.61} & \cellcolor[HTML]{CBCEFB}{\ul 6.12} & 13.54 & 15.01 & 4.85 & 8.44 & 15.96 & 4.83 & 7.80 \\
Engel & 71.68 & 63.78 & 6.76 & \cellcolor[HTML]{FFCCC9}\textbf{85.57} & \cellcolor[HTML]{FFCCC9}\textbf{77.96} & 14.49 & 79.14 & 4.92 & 5.14 & 77.97 & 70.72 & 6.11 \\
Goolam & \cellcolor[HTML]{CBCEFB}{\ul 73.55} & \cellcolor[HTML]{FFCCC9}\textbf{56.30} & 11.27 & 68.28 & 50.27 & 15.01 & 67.75 & 21.69 & 22.07 & 72.23 & \cellcolor[HTML]{CBCEFB}{\ul 56.23} & 13.63 \\
Koh & 93.61 & 90.09 & 8.36 & 93.69 & 90.89 & 18.53 & 88.82 & 4.35 & 2.31 & 85.52 & 81.10 & 4.55 \\
Kumar & 86.53 & 87.23 & 16.28 & 76.55 & 65.90 & 22.54 & 91.14 & 21.81 & \cellcolor[HTML]{FFCCC9}\textbf{24.43} & 95.16 & 96.32 & 7.30 \\
Leng & 63.77 & 62.94 & 3.81 & 50.23 & 36.55 & 3.42 & 64.13 & 1.69 & 1.88 & 55.70 & 57.79 & 1.24 \\
Li & 87.19 & 77.08 & 32.01 & 84.12 & 65.44 & 28.50 & 84.69 & 13.17 & 13.82 & 84.44 & 68.90 & 13.31 \\
Maria2 & 31.83 & 20.72 & 1.18 & 41.31 & \cellcolor[HTML]{FFCCC9}\textbf{44.18} & \cellcolor[HTML]{CBCEFB}{\ul 4.21} & 42.58 & 7.91 & 2.26 & 28.12 & 19.68 & 0.13 \\
Robert & \cellcolor[HTML]{CBCEFB}{\ul 64.26} & \cellcolor[HTML]{CBCEFB}{\ul 60.69} & \cellcolor[HTML]{CBCEFB}{\ul 30.83} & 52.27 & 35.44 & 2.10 & 57.65 & 6.22 & 10.38 & 60.72 & 56.63 & 19.72 \\
Ting & \cellcolor[HTML]{CBCEFB}{\ul 83.07} & \cellcolor[HTML]{CBCEFB}{\ul 61.87} & 16.51 & 76.39 & 54.93 & 22.69 & 78.18 & 12.37 & 12.85 & 80.37 & 59.14 & 13.73 \\
\begin{tabular}[c]{@{}c@{}}Mouse\\ Pancreas1\end{tabular} & \cellcolor[HTML]{CBCEFB}{\ul 80.62} & \cellcolor[HTML]{CBCEFB}{\ul 74.45} & 10.35 & 77.34 & 62.17 & \cellcolor[HTML]{CBCEFB}{\ul 15.99} & 78.77 & 8.84 & 13.33 & 75.26 & 64.12 & 7.61 \\
Cao & \cellcolor[HTML]{FFCCC9}\textbf{68.29} & \cellcolor[HTML]{CBCEFB}{\ul 48.06} & 6.44 & 56.15 & 27.48 & 7.24 & 60.37 & 6.98 & \cellcolor[HTML]{FFCCC9}\textbf{13.14} & 58.63 & 42.05 & 2.83 \\
Chu2 & 96.55 & 96.22 & 19.22 & 92.24 & 77.99 & 22.96 & 96.69 & 6.77 & 7.37 & 96.41 & 93.87 & 17.71 \\
Han & 76.18 & 64.96 & 4.50 & 68.44 & 53.26 & 9.23 & 74.69 & 10.24 & 9.78 & 76.00 & \cellcolor[HTML]{CBCEFB}{\ul 67.17} & 4.21 \\
MacParland & 80.61 & 66.26 & 4.21 & 66.36 & 48.41 & 0.76 & 78.08 & 4.51 & \cellcolor[HTML]{CBCEFB}{\ul 6.89} & 78.86 & 64.13 & 2.78 \\
Maria1 & 38.66 & 25.33 & 1.21 & 40.59 & 27.26 & 1.13 & 13.59 & 4.82 & \cellcolor[HTML]{CBCEFB}{\ul 3.88} & 27.00 & 21.82 & 0.15 \\
Puram & 70.62 & 40.22 & 7.93 & 57.28 & 17.58 & 14.95 & 75.41 & 2.85 & 6.69 & 69.56 & 41.04 & 5.49 \\
Yang & 61.22 & 41.80 & 8.66 & 58.99 & 39.03 & 14.08 & 62.34 & 7.88 & 8.17 & 63.62 & 53.14 & 10.79 \\
CITE & 16.34 & 7.76 & 2.91 & 57.23 & 40.48 & 2.16 & 63.14 & 10.65 & 9.15 & 62.74 & \cellcolor[HTML]{CBCEFB}{\ul 50.15} & 9.58 \\
\begin{tabular}[c]{@{}c@{}}Human\\ Pancreas1\end{tabular} & 82.45 & 59.83 & 11.01 & 79.56 & 50.10 & 18.93 & 80.50 & 10.79 & 15.62 & 81.73 & 61.15 & 8.35 \\
\begin{tabular}[c]{@{}c@{}}Human\\ Pancreas2\end{tabular} & 85.72 & 72.94 & 12.13 & 79.53 & 59.20 & \cellcolor[HTML]{CBCEFB}{\ul 22.65} & 84.68 & 12.81 & 18.33 & \cellcolor[HTML]{CBCEFB}{\ul 87.83} & 88.18 & 10.39 \\
\begin{tabular}[c]{@{}c@{}}Human\\ Pancreas3\end{tabular} & 86.67 & 87.58 & 17.14 & 77.81 & 63.75 & 19.36 & 85.42 & 15.37 & \cellcolor[HTML]{CBCEFB}{\ul 22.47} & \cellcolor[HTML]{CBCEFB}{\ul 88.98} & \cellcolor[HTML]{CBCEFB}{\ul 92.33} & 16.55 \\
\begin{tabular}[c]{@{}c@{}}Mouse\\ Pancreas2\end{tabular} & 70.95 & 39.41 & 4.76 & 69.52 & 37.62 & 12.84 & 72.11 & 5.81 & \cellcolor[HTML]{CBCEFB}{\ul 13.23} & 67.46 & 34.47 & 2.69 \\ \midrule
 & \multicolumn{3}{c}{HRG} & \multicolumn{3}{c}{GeneBasis} & \multicolumn{3}{c}{CellBRF} & \multicolumn{3}{c}{\model} \\
\multirow{-2}{*}{Dataset} & NMI & \cellcolor[HTML]{EFEFEF}ARI & SI & \cellcolor[HTML]{EFEFEF}NMI & ARI & \cellcolor[HTML]{EFEFEF}SI & NMI & \cellcolor[HTML]{EFEFEF}ARI & SI & \cellcolor[HTML]{EFEFEF}NMI & ARI & \cellcolor[HTML]{EFEFEF}SI \\ \midrule
Chu1 & 80.71 & 67.68 & 13.23 & 74.53 & 62.29 & 15.32 & \cellcolor[HTML]{CBCEFB}{\ul 84.87} & \cellcolor[HTML]{CBCEFB}{\ul 78.25} & 16.39 & \cellcolor[HTML]{FFCCC9}\textbf{88.83} & \cellcolor[HTML]{FFCCC9}\textbf{82.71} & \cellcolor[HTML]{FFCCC9}\textbf{17.90} \\
Chung & 47.36 & 17.89 & 16.86 & 47.43 & \cellcolor[HTML]{CBCEFB}{\ul 19.10} & \cellcolor[HTML]{FFCCC9}\textbf{18.97} & 46.73 & \cellcolor[HTML]{FFCCC9}\textbf{19.24} & 16.70 & 48.29 & 18.31 & \cellcolor[HTML]{CBCEFB}{\ul 18.00} \\
Darmanis & 17.47 & 4.80 & 10.32 & 13.04 & 4.79 & \cellcolor[HTML]{CBCEFB}{\ul 13.64} & 17.79 & 5.24 & \cellcolor[HTML]{FFCCC9}\textbf{13.86} & \cellcolor[HTML]{FFCCC9}\textbf{19.65} & \cellcolor[HTML]{FFCCC9}\textbf{6.75} & 11.54 \\
Engel & 75.72 & 67.16 & 8.89 & 68.20 & 64.01 & 7.30 & \cellcolor[HTML]{CBCEFB}{\ul 80.85} & \cellcolor[HTML]{CBCEFB}{\ul 73.64} & \cellcolor[HTML]{FFCCC9}\textbf{20.18} & 80.49 & 72.55 & \cellcolor[HTML]{CBCEFB}{\ul 14.80} \\
Goolam & 67.13 & 38.11 & \cellcolor[HTML]{FFCCC9}\textbf{27.97} & 60.59 & 36.92 & 10.70 & 71.27 & 46.31 & \cellcolor[HTML]{CBCEFB}{\ul 22.82} & \cellcolor[HTML]{FFCCC9}\textbf{74.45} & 51.60 & 18.36 \\
Koh & 93.85 & 90.52 & 9.08 & 89.38 & 85.76 & 10.05 & \cellcolor[HTML]{CBCEFB}{\ul 98.44} & \cellcolor[HTML]{CBCEFB}{\ul 98.28} & \cellcolor[HTML]{FFCCC9}\textbf{23.59} & \cellcolor[HTML]{FFCCC9}\textbf{99.09} & \cellcolor[HTML]{FFCCC9}\textbf{99.21} & \cellcolor[HTML]{CBCEFB}{\ul 20.58} \\
Kumar & \cellcolor[HTML]{CBCEFB}{\ul 96.70} & \cellcolor[HTML]{CBCEFB}{\ul 97.05} & 12.93 & 88.29 & 87.66 & 13.62 & 90.30 & 86.85 & 22.65 & \cellcolor[HTML]{FFCCC9}\textbf{98.07} & \cellcolor[HTML]{FFCCC9}\textbf{98.51} & \cellcolor[HTML]{CBCEFB}{\ul 24.30} \\
Leng & 56.80 & 59.11 & 1.01 & \cellcolor[HTML]{CBCEFB}{\ul 6.97} & 2.79 & 0.93 & \cellcolor[HTML]{CBCEFB}{\ul 70.37} & \cellcolor[HTML]{CBCEFB}{\ul 71.18} & \cellcolor[HTML]{FFCCC9}\textbf{8.77} & \cellcolor[HTML]{FFCCC9}\textbf{82.82} & \cellcolor[HTML]{FFCCC9}\textbf{85.72} & \cellcolor[HTML]{CBCEFB}{\ul 7.97} \\
Li & 88.07 & 78.57 & 34.26 & \cellcolor[HTML]{CBCEFB}{\ul 89.06} & \cellcolor[HTML]{CBCEFB}{\ul 78.98} & 25.74 & 89.03 & 78.92 & \cellcolor[HTML]{CBCEFB}{\ul 40.63} & \cellcolor[HTML]{FFCCC9}\textbf{93.41} & \cellcolor[HTML]{FFCCC9}\textbf{83.33} & \cellcolor[HTML]{FFCCC9}\textbf{41.63} \\
Maria2 & 33.01 & 21.28 & 1.45 & 35.37 & 29.79 & 1.64 & \cellcolor[HTML]{FFCCC9}\textbf{53.73} & \cellcolor[HTML]{CBCEFB}{\ul 43.87} & \cellcolor[HTML]{FFCCC9}\textbf{7.00} & \cellcolor[HTML]{CBCEFB}{\ul 43.20} & 33.14 & 2.90 \\
Robert & 59.84 & 51.06 & 27.83 & 52.57 & 36.83 & 9.58 & 55.38 & 38.49 & 17.21 & \cellcolor[HTML]{FFCCC9}\textbf{71.13} & \cellcolor[HTML]{FFCCC9}\textbf{73.10} & \cellcolor[HTML]{FFCCC9}\textbf{52.86} \\
Ting & 79.28 & 58.92 & 16.90 & 81.48 & 60.58 & 18.37 & 77.63 & 58.34 & \cellcolor[HTML]{FFCCC9}\textbf{28.36} & \cellcolor[HTML]{FFCCC9}\textbf{83.17} & \cellcolor[HTML]{FFCCC9}\textbf{62.30} & \cellcolor[HTML]{CBCEFB}{\ul 25.78} \\
\begin{tabular}[c]{@{}c@{}}Mouse\\ Pancreas1\end{tabular} & 74.04 & 62.67 & 7.23 & 75.93 & 54.28 & \cellcolor[HTML]{FFCCC9}\textbf{16.57} & 77.19 & 60.87 & 15.54 & \cellcolor[HTML]{FFCCC9}\textbf{84.57} & \cellcolor[HTML]{FFCCC9}\textbf{78.14} & 14.17 \\
Cao & 56.93 & 41.67 & 1.48 & 49.75 & 32.23 & 7.20 & 47.12 & 26.53 & \cellcolor[HTML]{CBCEFB}{\ul 10.54} & \cellcolor[HTML]{CBCEFB}{\ul 63.66} & \cellcolor[HTML]{FFCCC9}\textbf{50.53} & 8.13 \\
Chu2 & 96.30 & 95.95 & 12.85 & 99.05 & 99.23 & 20.08 & \cellcolor[HTML]{CBCEFB}{\ul 99.40} & \cellcolor[HTML]{CBCEFB}{\ul 99.64} & \cellcolor[HTML]{FFCCC9}\textbf{33.92} & \cellcolor[HTML]{FFCCC9}\textbf{100.00} & \cellcolor[HTML]{FFCCC9}\textbf{100.00} & \cellcolor[HTML]{CBCEFB}{\ul 29.81} \\
Han & \cellcolor[HTML]{CBCEFB}{\ul 76.54} & \cellcolor[HTML]{FFCCC9}\textbf{68.41} & 3.15 & 68.94 & 51.47 & \cellcolor[HTML]{FFCCC9}\textbf{11.26} & 76.05 & 66.38 & 10.08 & \cellcolor[HTML]{FFCCC9}\textbf{78.87} & 66.57 & \cellcolor[HTML]{CBCEFB}{\ul 10.33} \\
MacParland & 73.22 & 48.05 & 2.32 & \cellcolor[HTML]{CBCEFB}{\ul 70.89} & 51.93 & \cellcolor[HTML]{FFCCC9}\textbf{8.91} & \cellcolor[HTML]{CBCEFB}{\ul 82.74} & \cellcolor[HTML]{CBCEFB}{\ul 70.48} & 5.71 & \cellcolor[HTML]{FFCCC9}\textbf{83.94} & \cellcolor[HTML]{FFCCC9}\textbf{75.51} & 5.73 \\
Maria1 & 37.93 & 22.95 & 0.97 & \cellcolor[HTML]{CBCEFB}{\ul 45.52} & \cellcolor[HTML]{CBCEFB}{\ul 37.00} & 2.66 & \cellcolor[HTML]{FFCCC9}\textbf{51.22} & \cellcolor[HTML]{FFCCC9}\textbf{40.16} & \cellcolor[HTML]{FFCCC9}\textbf{5.34} & 43.77 & 30.86 & 3.25 \\
Puram & 71.28 & 43.69 & 3.65 & 65.75 & 37.69 & 11.06 & \cellcolor[HTML]{CBCEFB}{\ul 79.51} & \cellcolor[HTML]{CBCEFB}{\ul 65.57} & \cellcolor[HTML]{FFCCC9}\textbf{17.04} & \cellcolor[HTML]{FFCCC9}\textbf{80.16} & \cellcolor[HTML]{FFCCC9}\textbf{67.58} & \cellcolor[HTML]{CBCEFB}{\ul 15.26} \\
Yang & 62.06 & 42.25 & 12.15 & 62.72 & 42.43 & \cellcolor[HTML]{CBCEFB}{\ul 15.99} & \cellcolor[HTML]{FFCCC9}\textbf{66.58} & \cellcolor[HTML]{CBCEFB}{\ul 53.60} & 13.96 & \cellcolor[HTML]{CBCEFB}{\ul 66.36} & \cellcolor[HTML]{FFCCC9}\textbf{54.50} & \cellcolor[HTML]{FFCCC9}\textbf{18.68} \\
CITE & \cellcolor[HTML]{CBCEFB}{\ul 68.69} & 48.50 & 5.81 & \cellcolor[HTML]{FFCCC9}\textbf{69.79} & 49.03 & 11.30 & 63.73 & 44.57 & \cellcolor[HTML]{CBCEFB}{\ul 14.49} & 66.98 & \cellcolor[HTML]{FFCCC9}\textbf{53.86} & \cellcolor[HTML]{FFCCC9}\textbf{16.60} \\
\begin{tabular}[c]{@{}c@{}}Human\\ Pancreas1\end{tabular} & 67.87 & 60.55 & 2.52 & 80.74 & 55.38 & \cellcolor[HTML]{CBCEFB}{\ul 19.65} & \cellcolor[HTML]{CBCEFB}{\ul 83.29} & \cellcolor[HTML]{CBCEFB}{\ul 66.37} & \cellcolor[HTML]{FFCCC9}\textbf{19.81} & \cellcolor[HTML]{FFCCC9}\textbf{86.60} & \cellcolor[HTML]{FFCCC9}\textbf{73.78} & 18.50 \\
\begin{tabular}[c]{@{}c@{}}Human\\ Pancreas2\end{tabular} & 85.30 & \cellcolor[HTML]{CBCEFB}{\ul 89.35} & 5.26 & 82.70 & 68.12 & \cellcolor[HTML]{FFCCC9}\textbf{22.84} & 79.14 & 59.97 & 15.71 & \cellcolor[HTML]{FFCCC9}\textbf{89.58} & \cellcolor[HTML]{FFCCC9}\textbf{91.55} & 21.37 \\
\begin{tabular}[c]{@{}c@{}}Human\\ Pancreas3\end{tabular} & 84.17 & 87.76 & 11.47 & 84.88 & 88.05 & 20.65 & 80.47 & 63.87 & 10.56 & \cellcolor[HTML]{FFCCC9}\textbf{89.98} & \cellcolor[HTML]{FFCCC9}\textbf{93.14} & \cellcolor[HTML]{FFCCC9}\textbf{28.97} \\
\begin{tabular}[c]{@{}c@{}}Mouse\\ Pancreas2\end{tabular} & \cellcolor[HTML]{CBCEFB}{\ul 74.56} & \cellcolor[HTML]{CBCEFB}{\ul 48.43} & 8.56 & 67.80 & 37.35 & \cellcolor[HTML]{FFCCC9}\textbf{14.45} & 69.51 & 38.33 & 12.16 & \cellcolor[HTML]{FFCCC9}\textbf{77.92} & \cellcolor[HTML]{FFCCC9}\textbf{60.07} & 13.04 \\ \bottomrule
\end{tabular}%
}
\end{table}

% \section{Detailed of the All Dataset Visualization Analysis}

\begin{figure}[htbp]
\centering
\renewcommand{\thesubfigure}{\arabic{subfigure}}
\subfloat[Cao]{
		\includegraphics[scale=0.07]{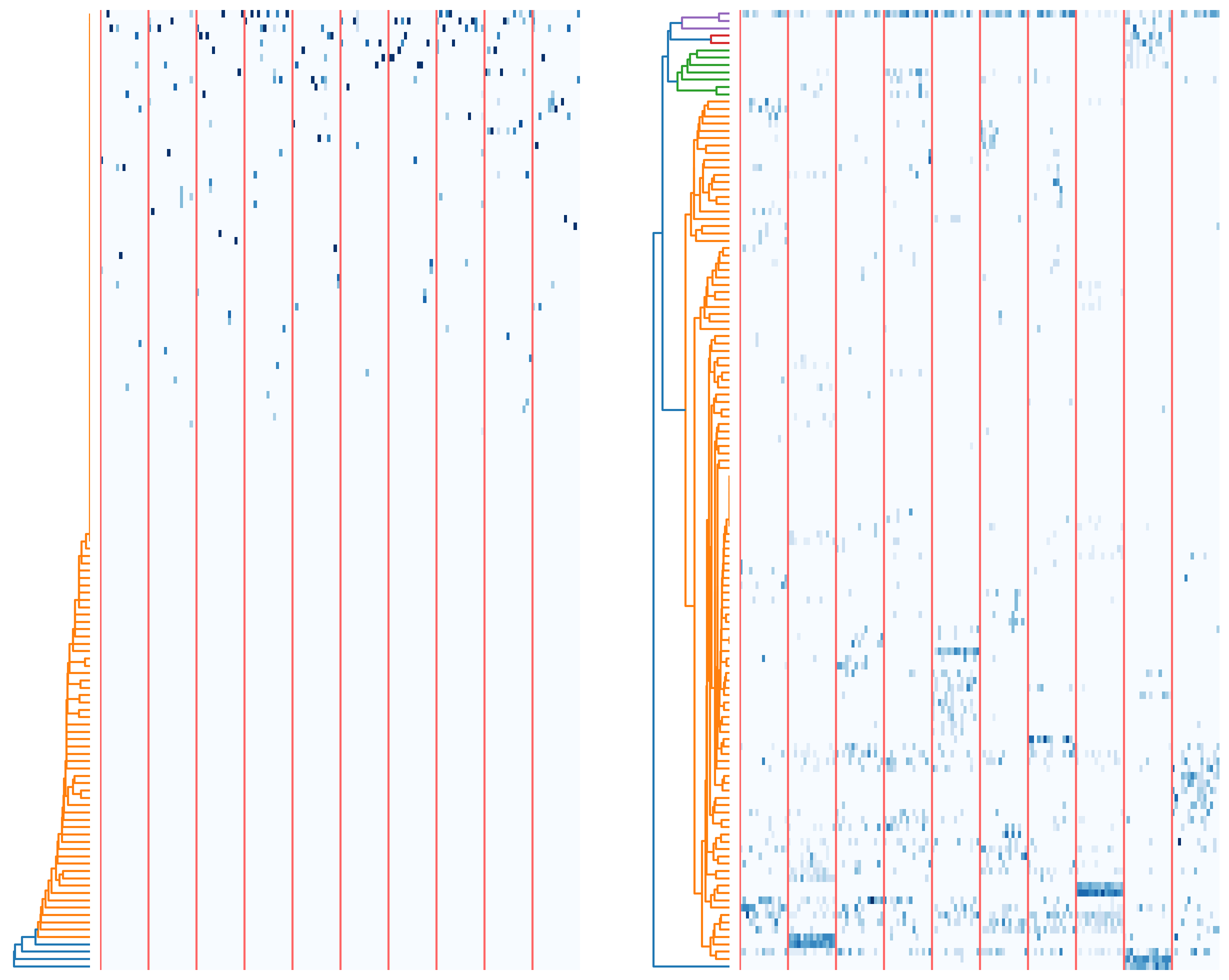}}
\subfloat[Chu1]{
		\includegraphics[scale=0.07]{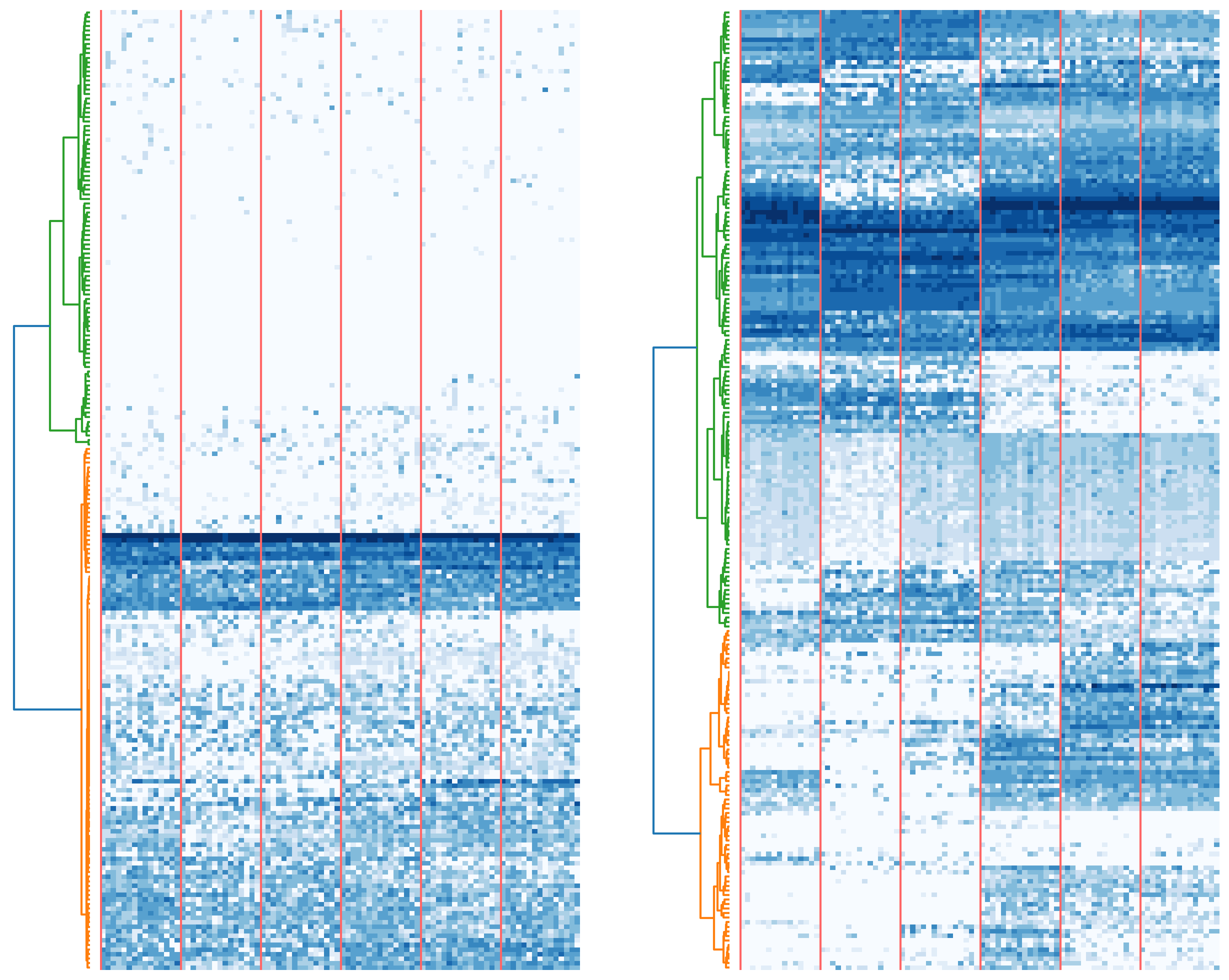}}
\\
\subfloat[Chu2]{
		\includegraphics[scale=0.07]{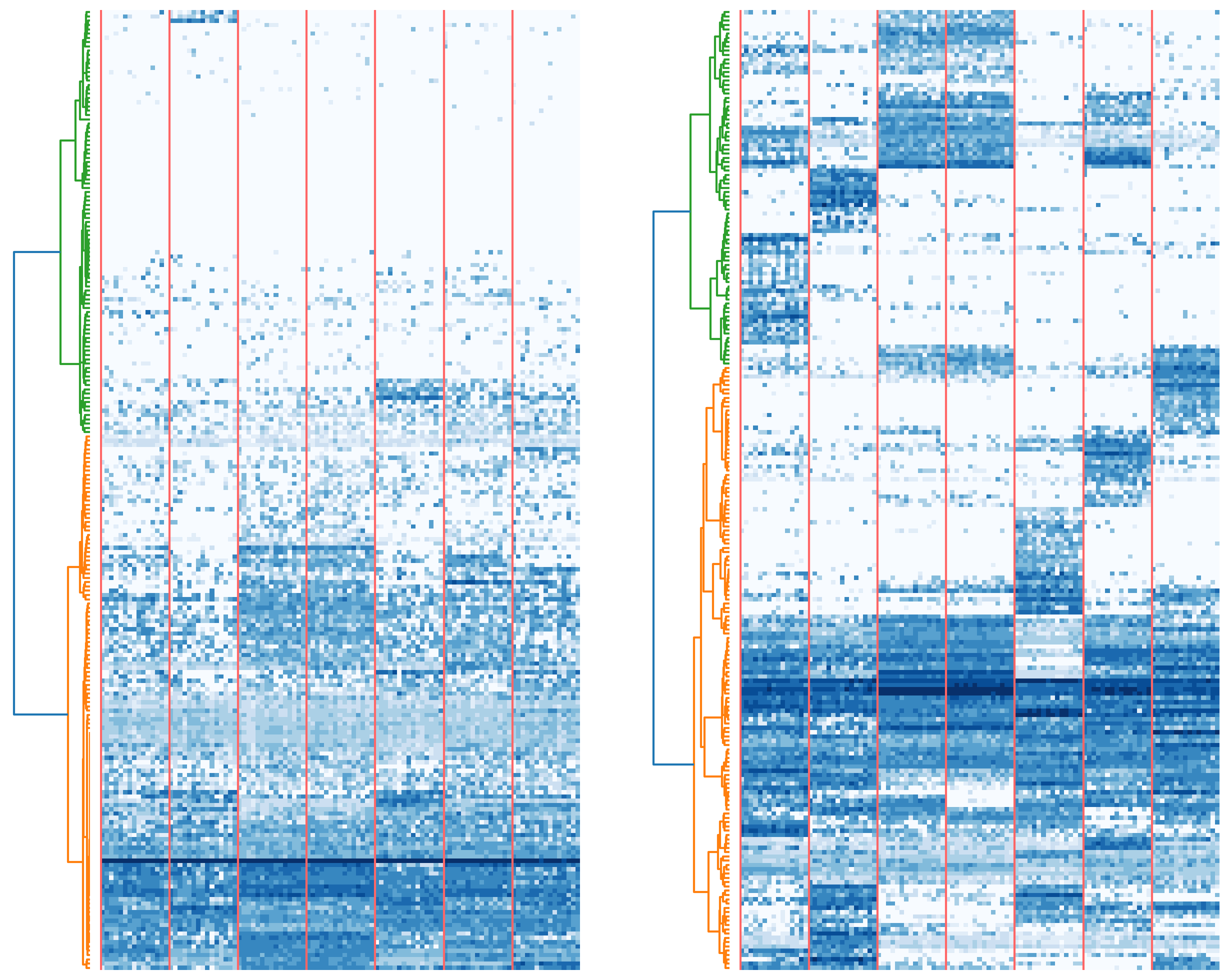}}
\subfloat[Han]{
		\includegraphics[scale=0.07]{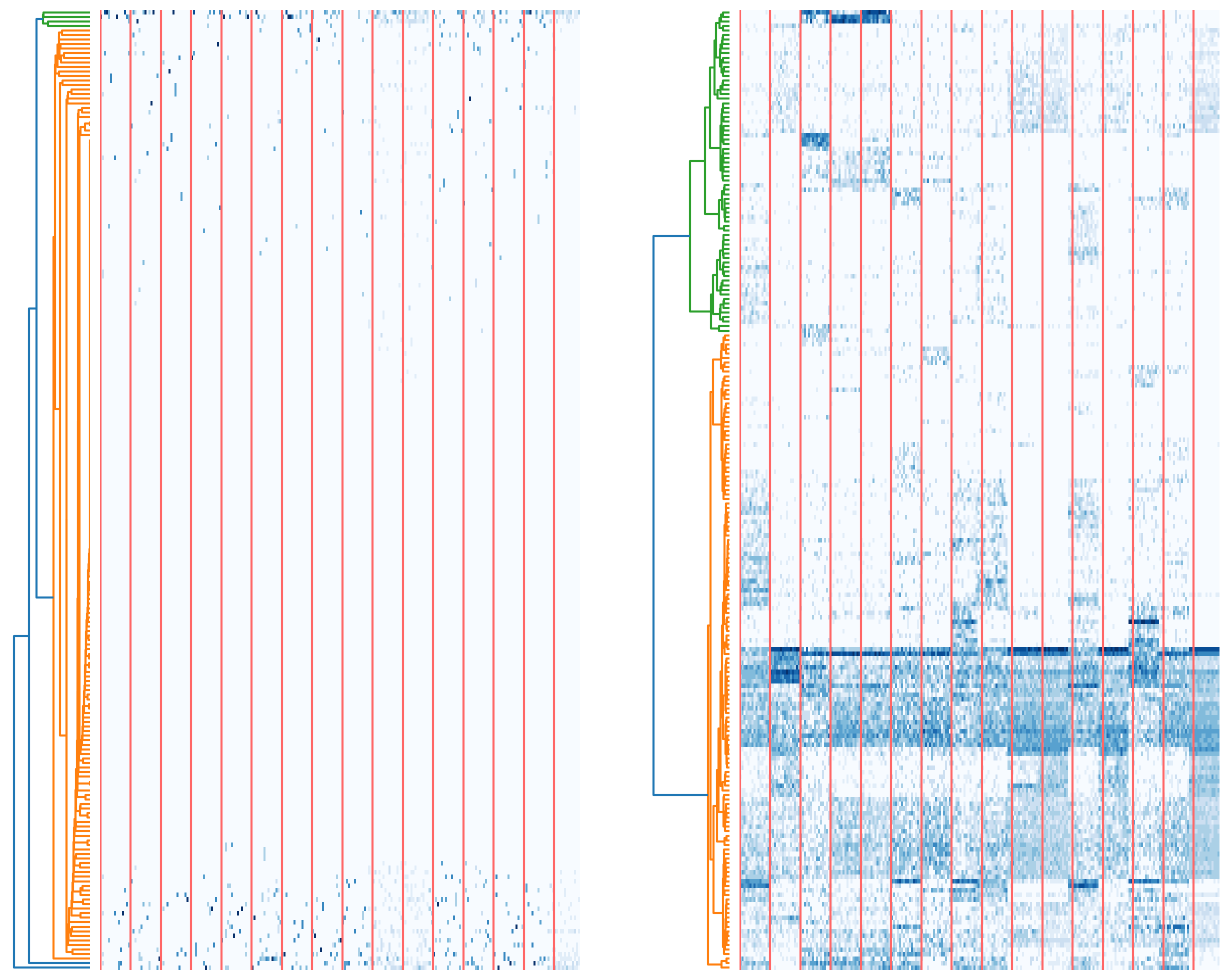}}
\\
\subfloat[CITE CBMC]{
		\includegraphics[scale=0.07]{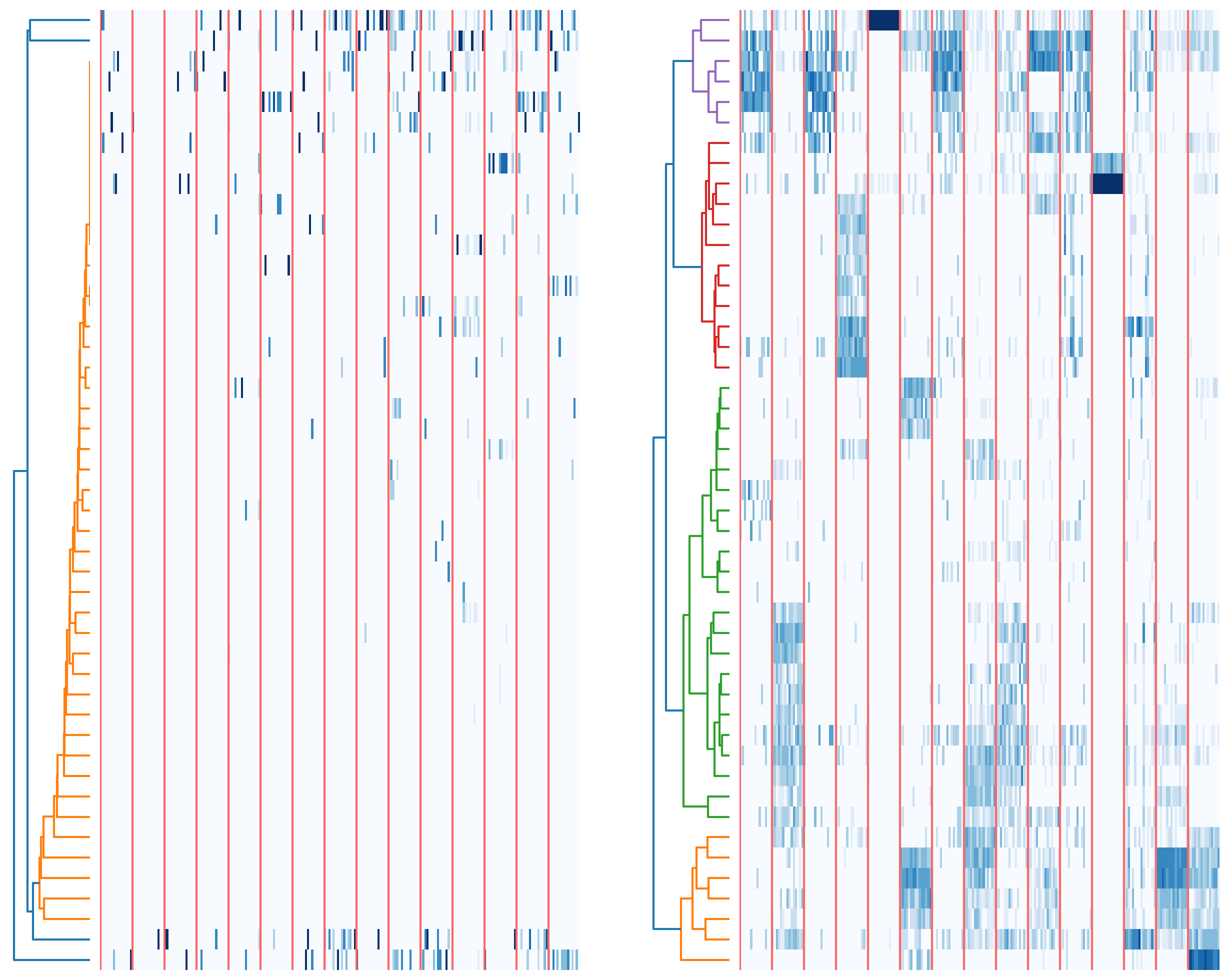}}
\subfloat[Human Pancreas1]{
		\includegraphics[scale=0.07]{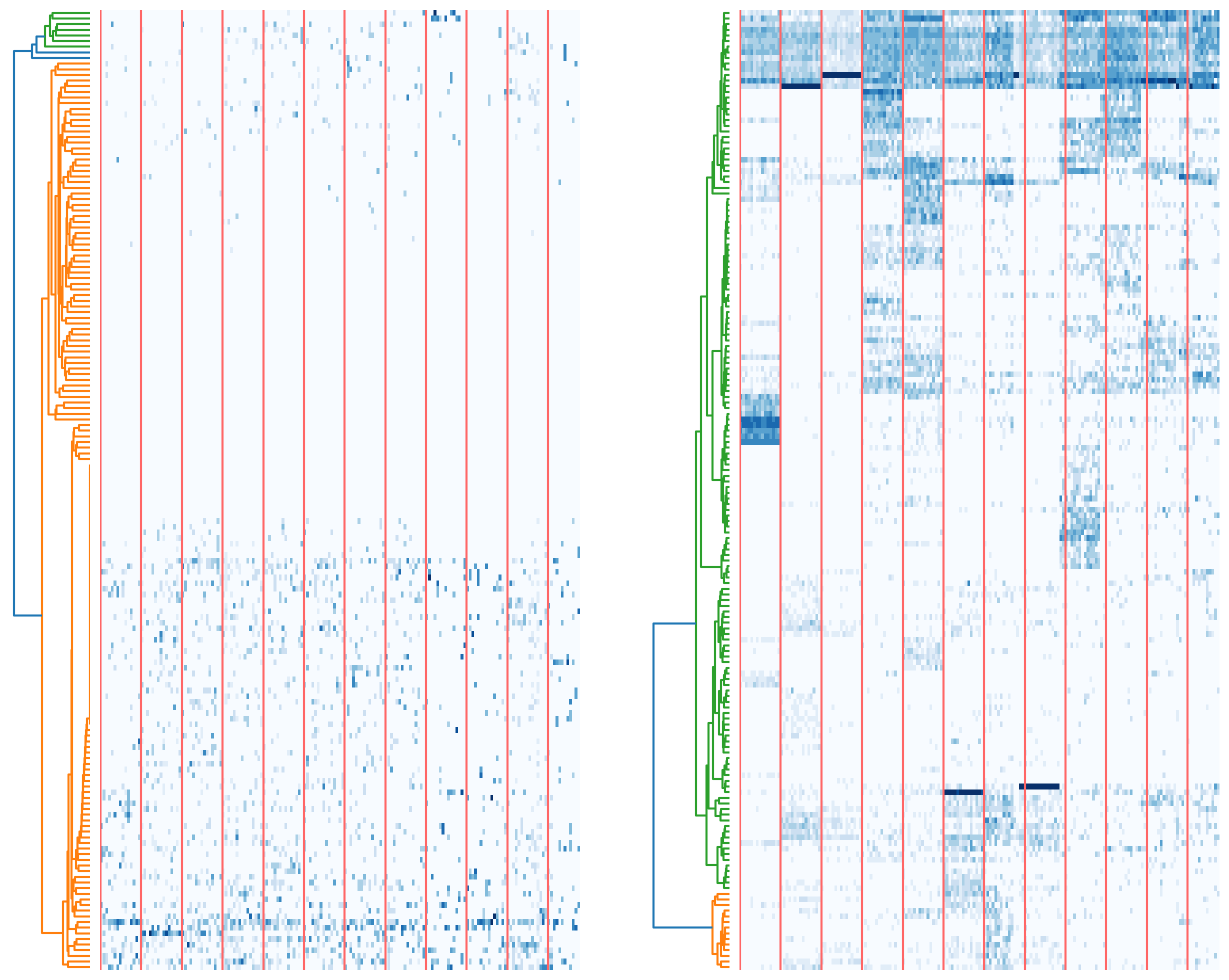}}
\\
\subfloat[Human Pancreas2]{
		\includegraphics[scale=0.07]{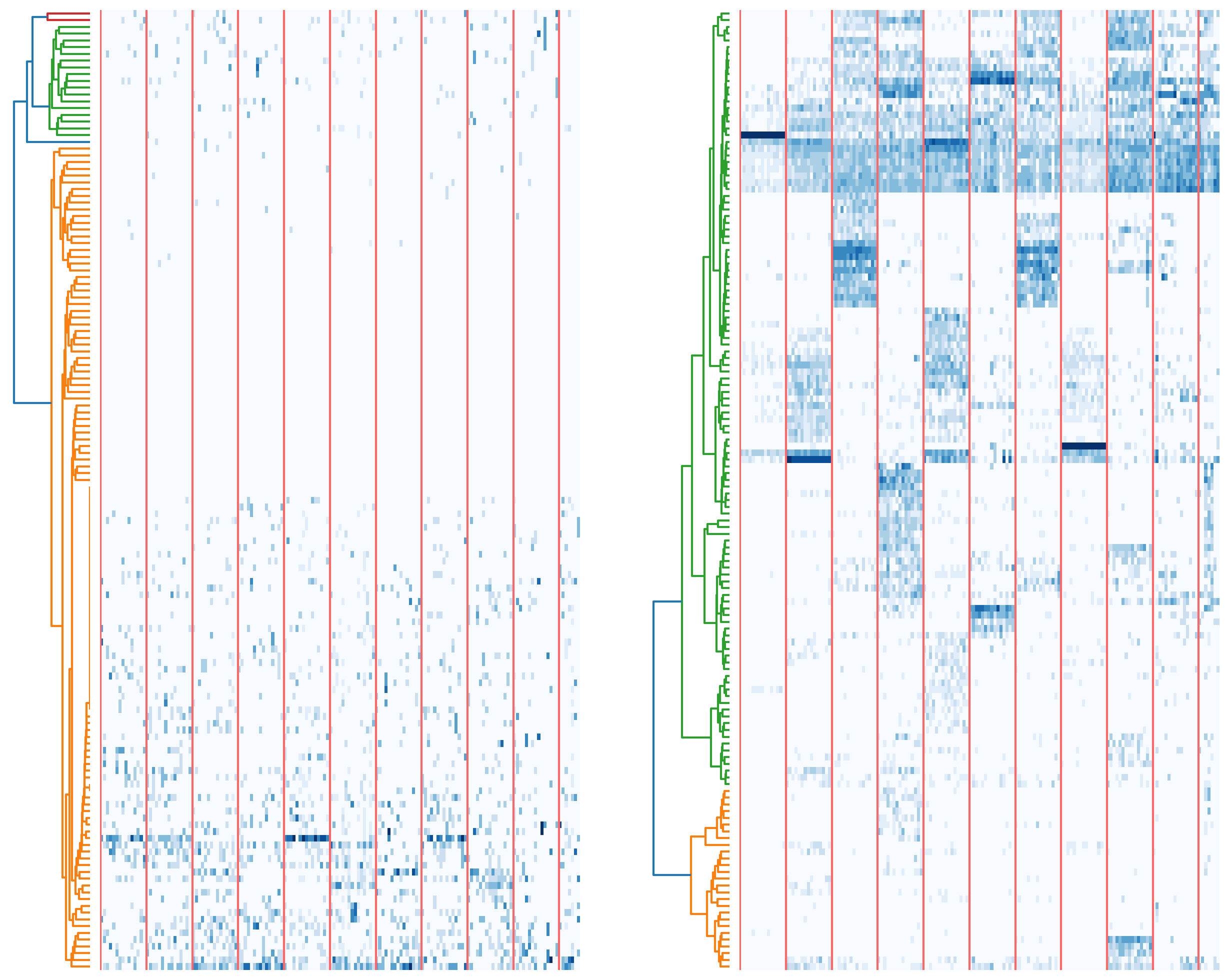}}
\subfloat[Human Pancreas3]{
		\includegraphics[scale=0.07]{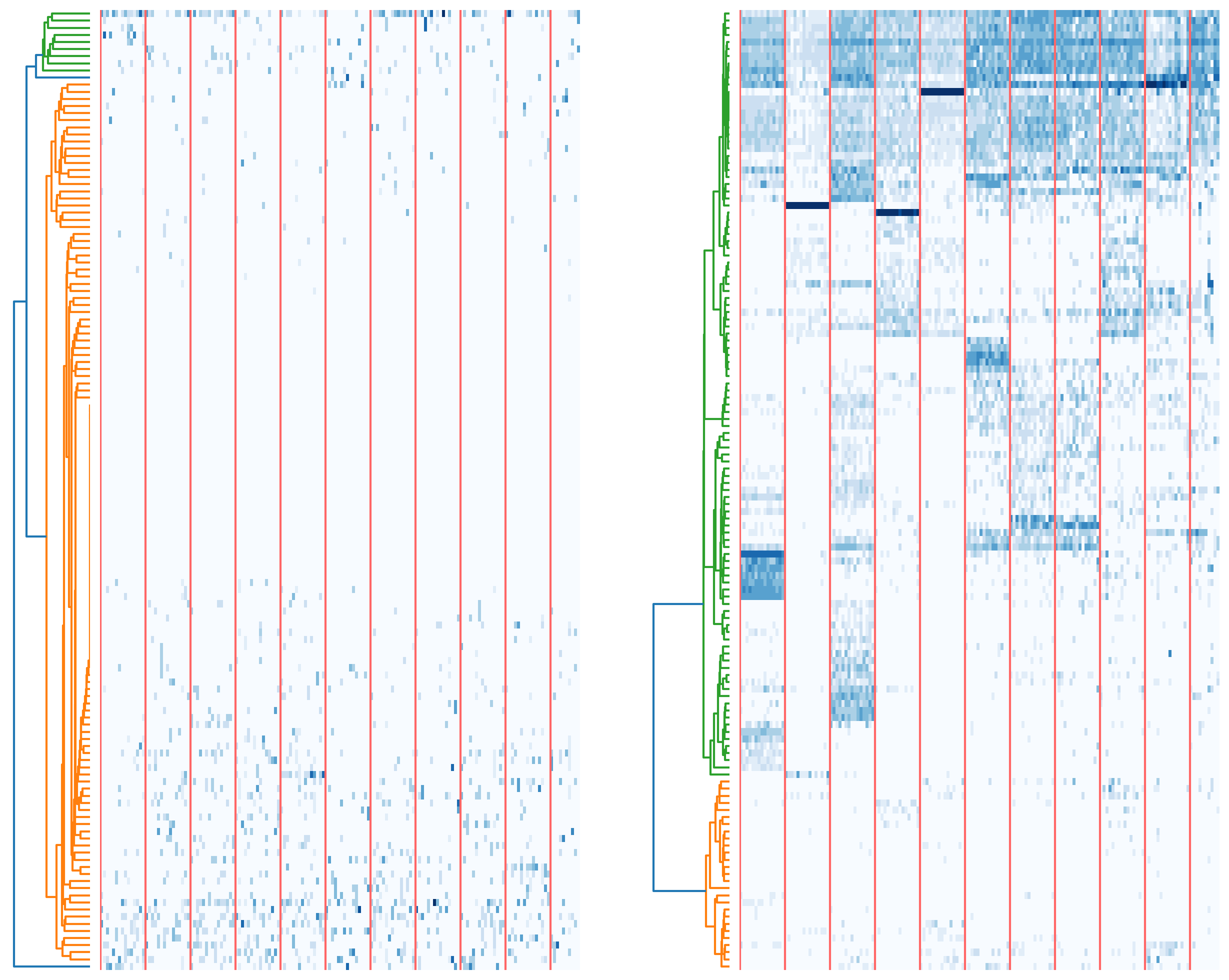}}
\\
\end{figure}

\begin{figure}[htbp]
\renewcommand{\thesubfigure}{\arabic{subfigure}}
\centering
\ContinuedFloat
\subfloat[Chung]{
		\includegraphics[scale=0.07]{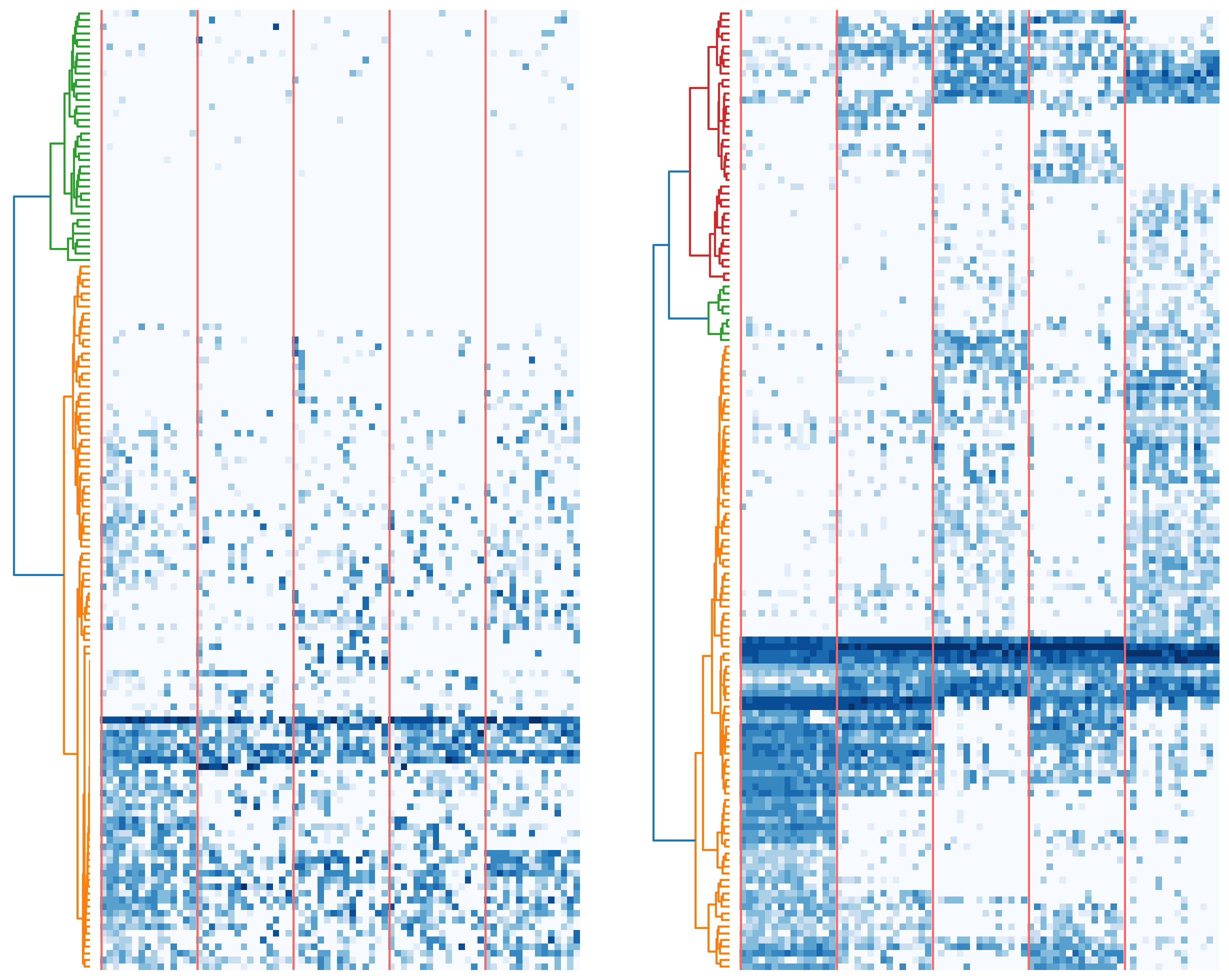}}
\subfloat[Darmanis]{
		\includegraphics[scale=0.07]{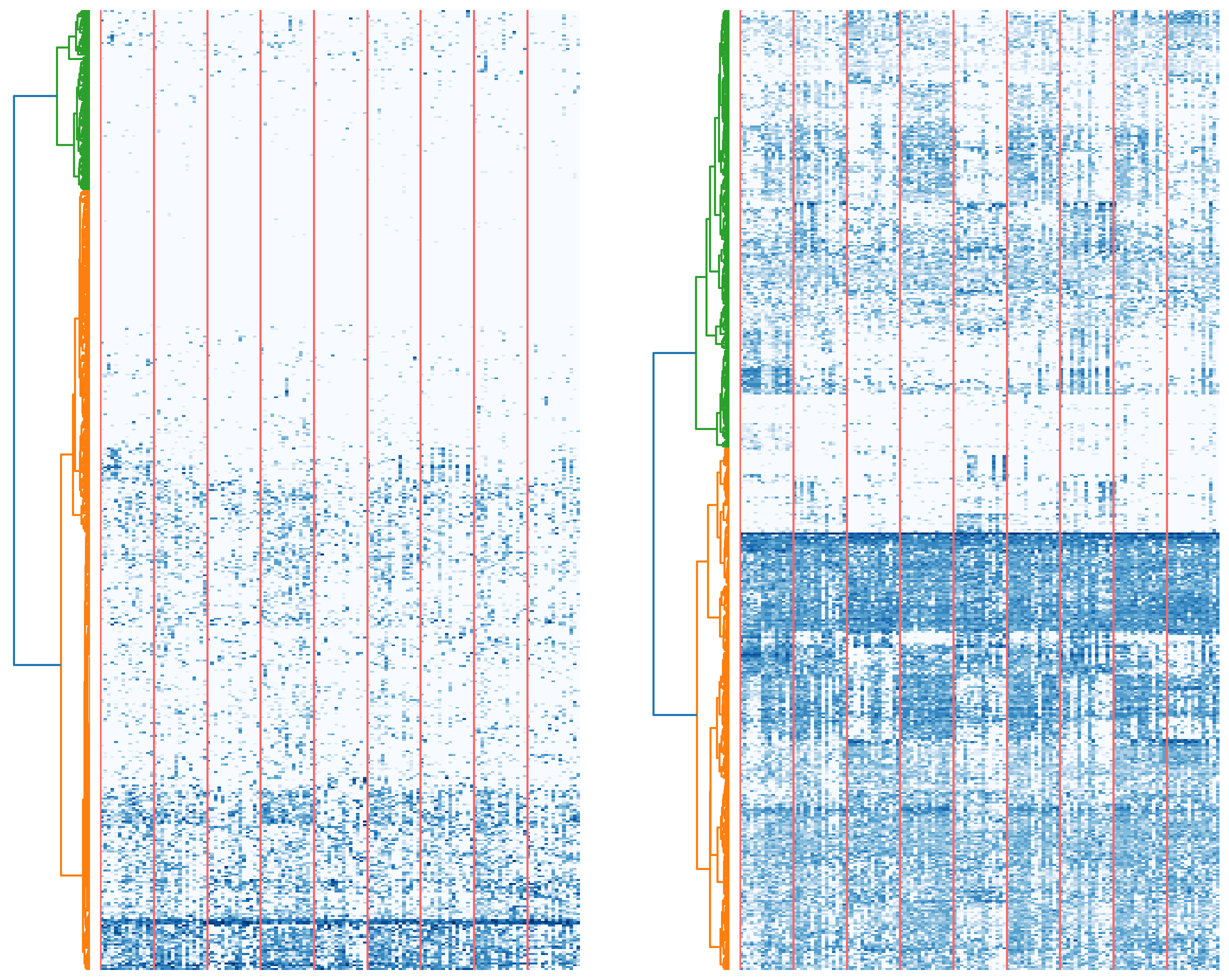}}
\\
\subfloat[Engel]{
		\includegraphics[scale=0.07]{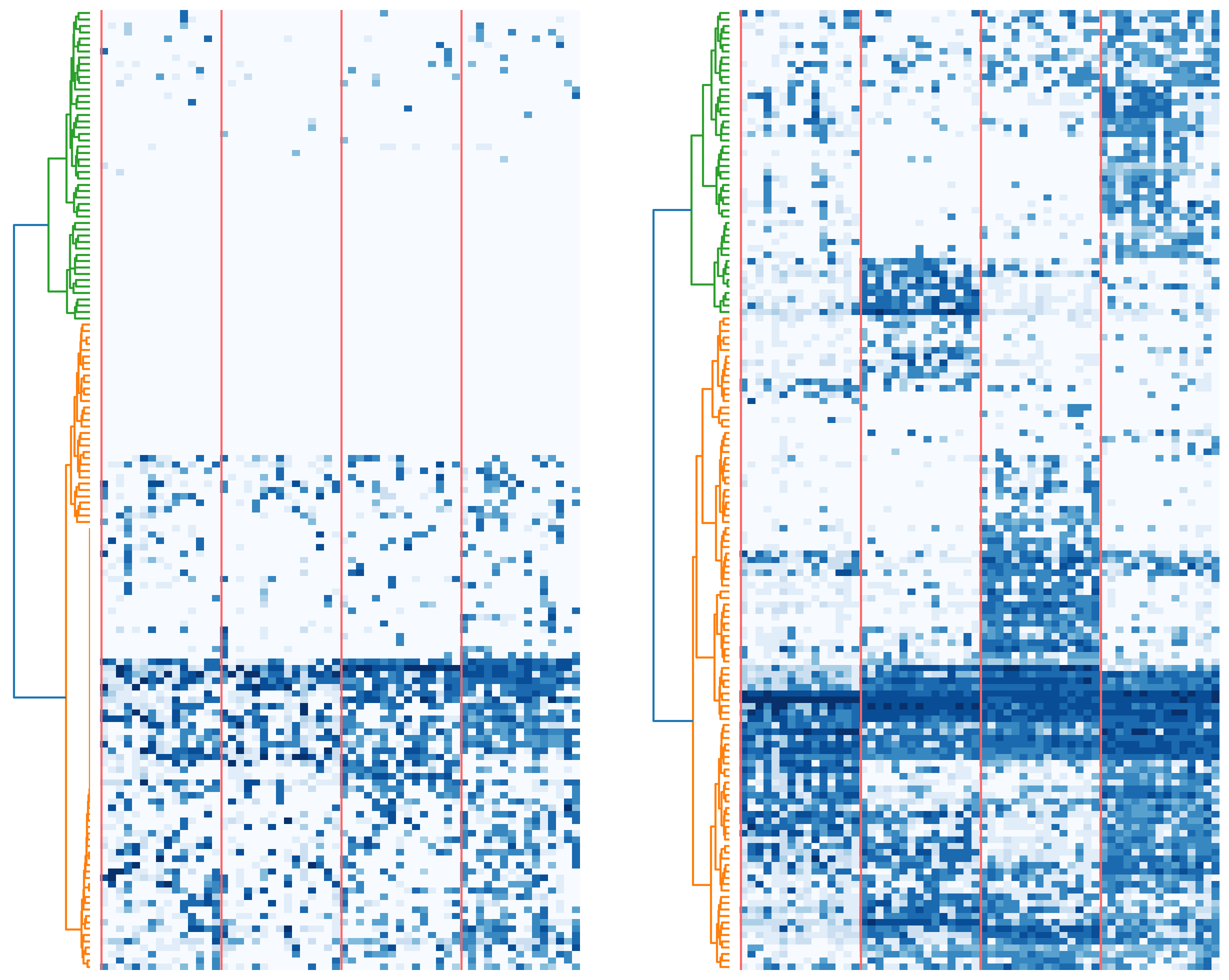}}
\subfloat[Goolam]{
		\includegraphics[scale=0.07]{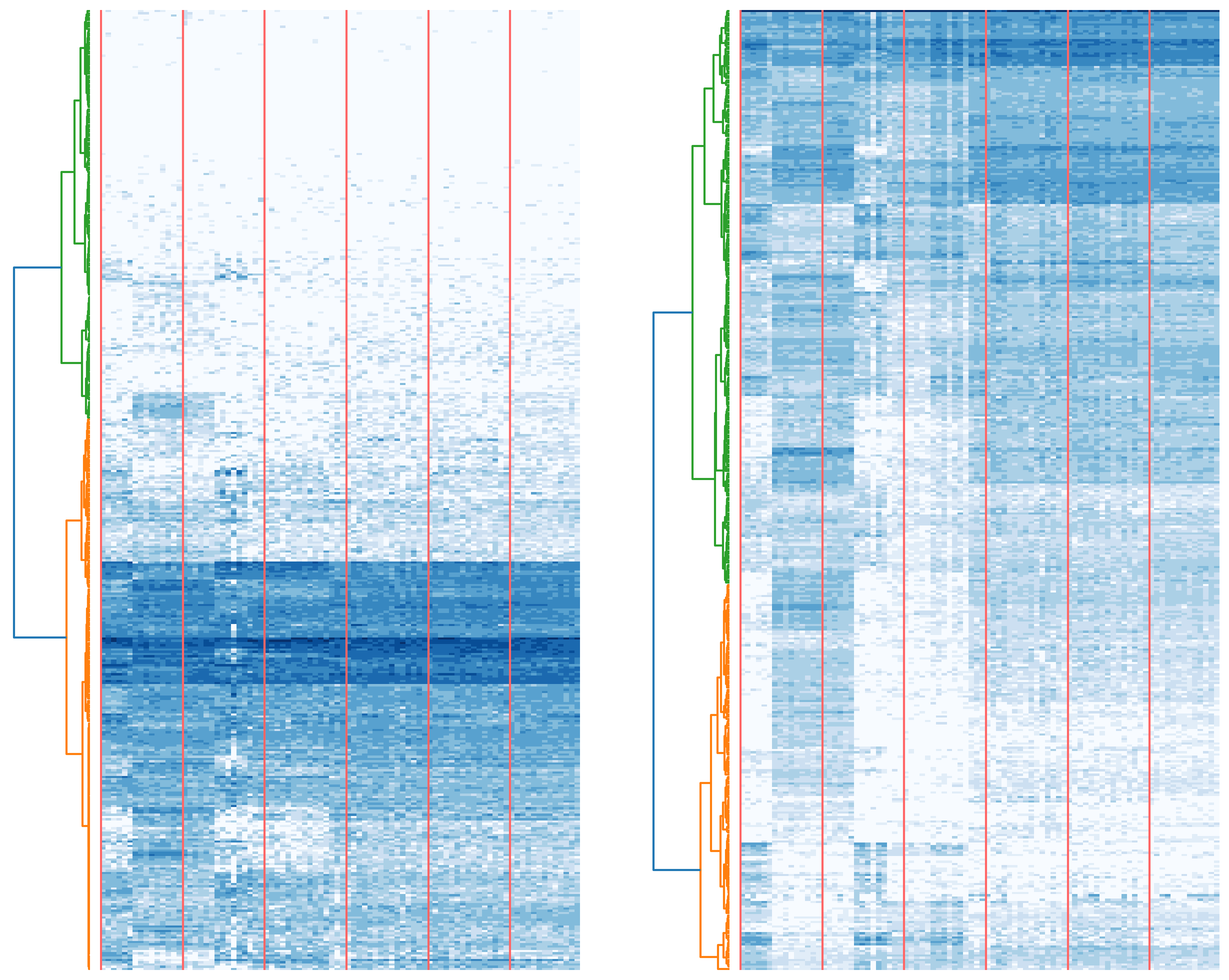}}
\\
\subfloat[Koh]{
		\includegraphics[scale=0.07]{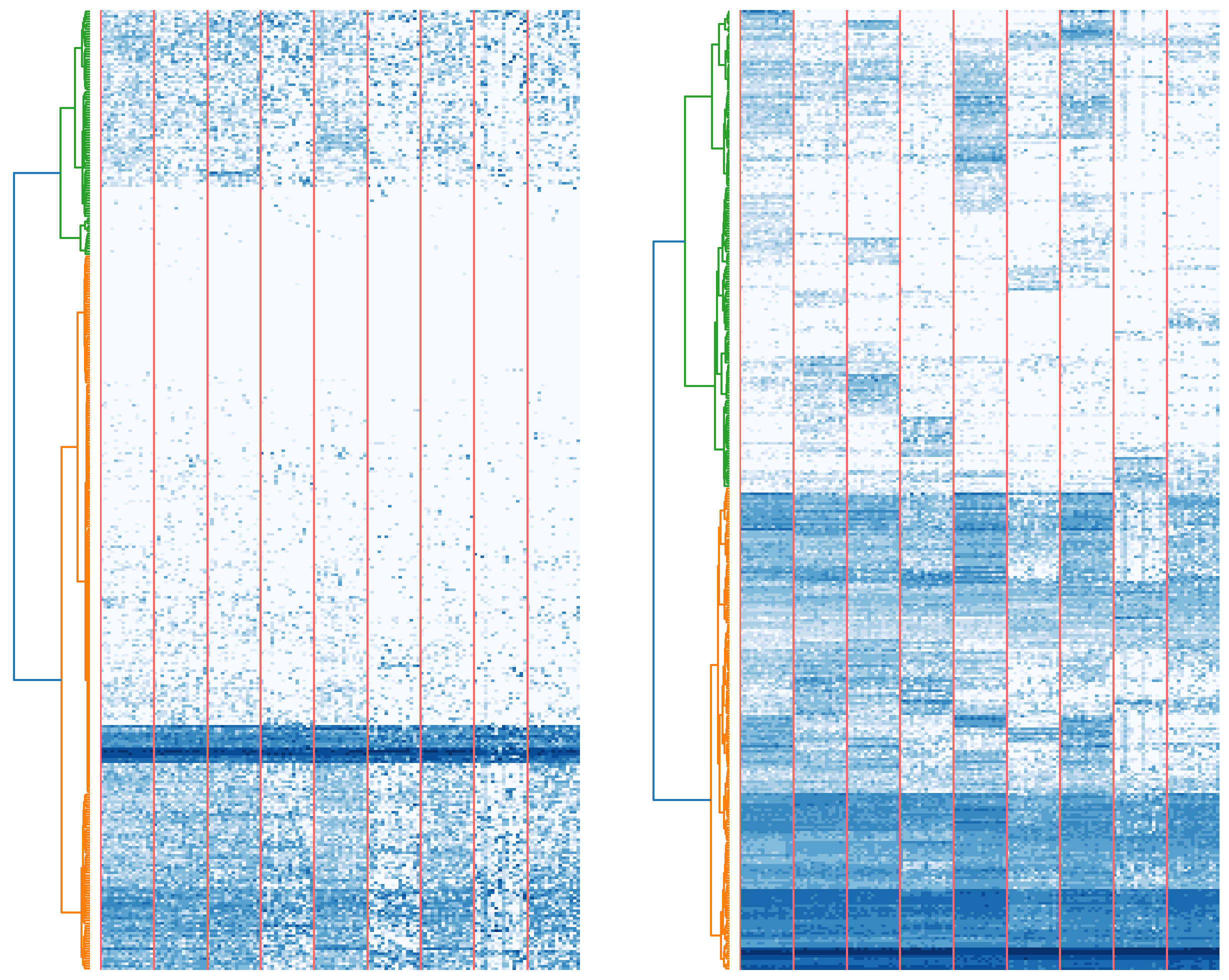}}
\subfloat[Kumar]{
		\includegraphics[scale=0.07]{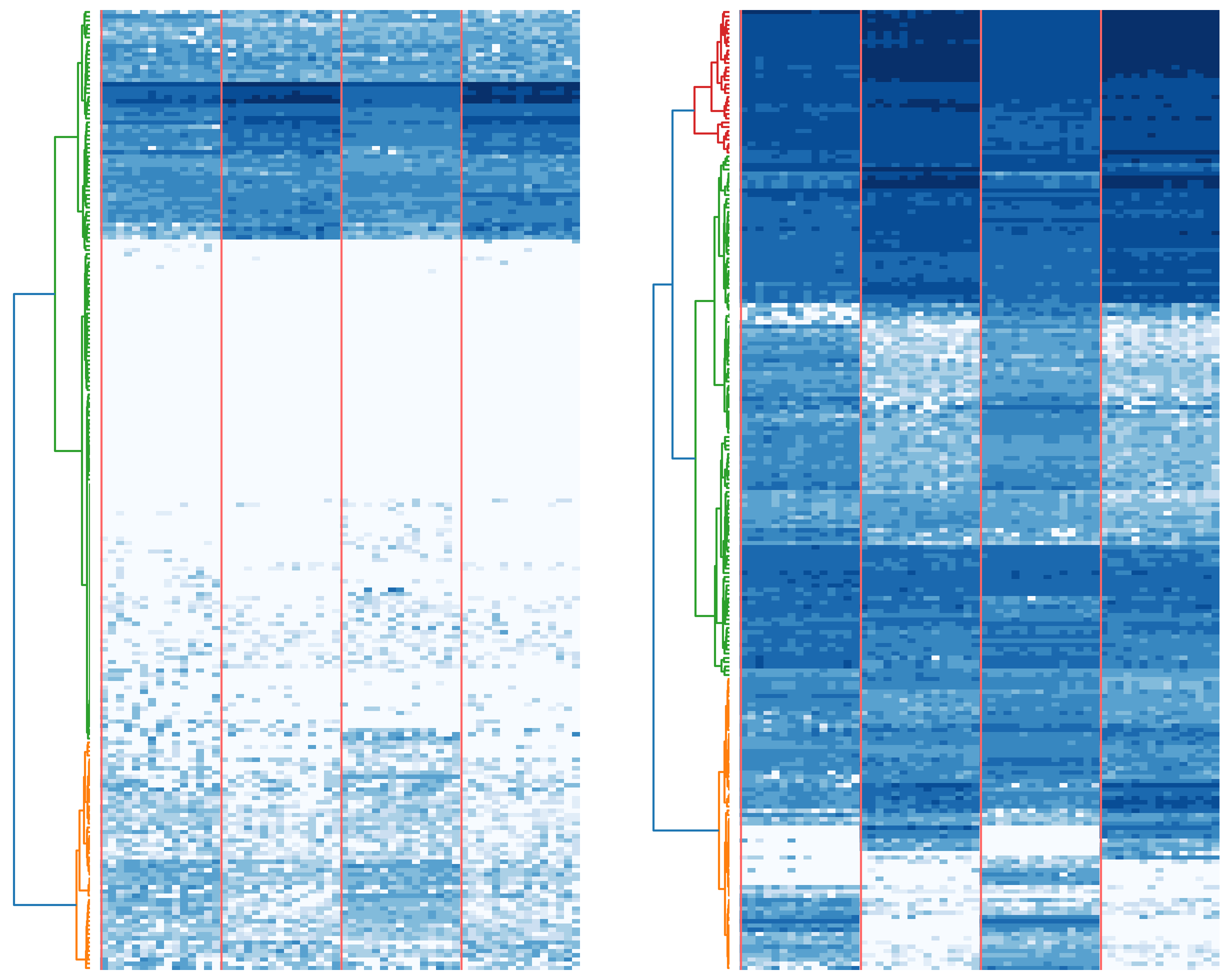}}
\\
\subfloat[Leng]{
		\includegraphics[scale=0.07]{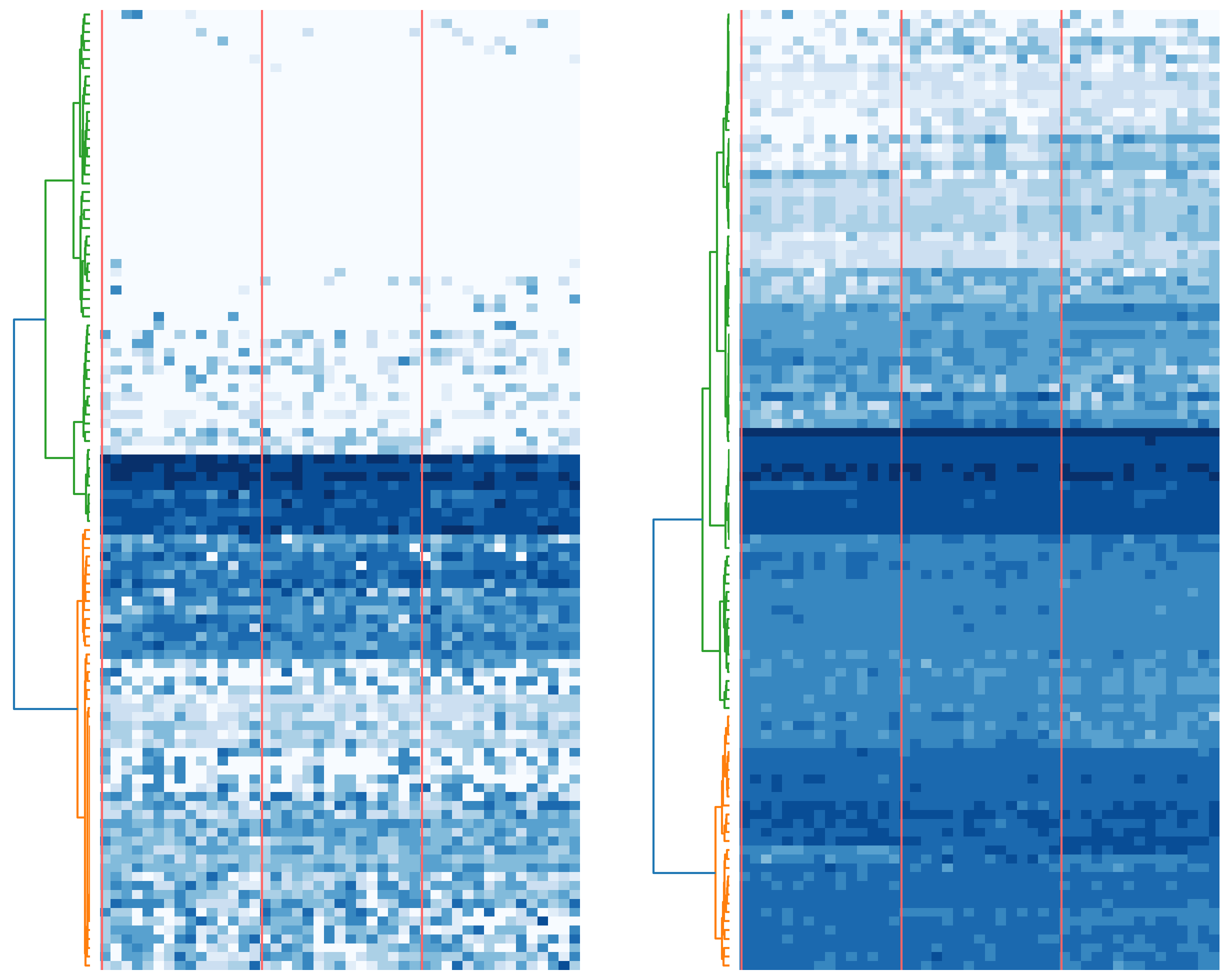}}
\subfloat[Li]{
		\includegraphics[scale=0.07]{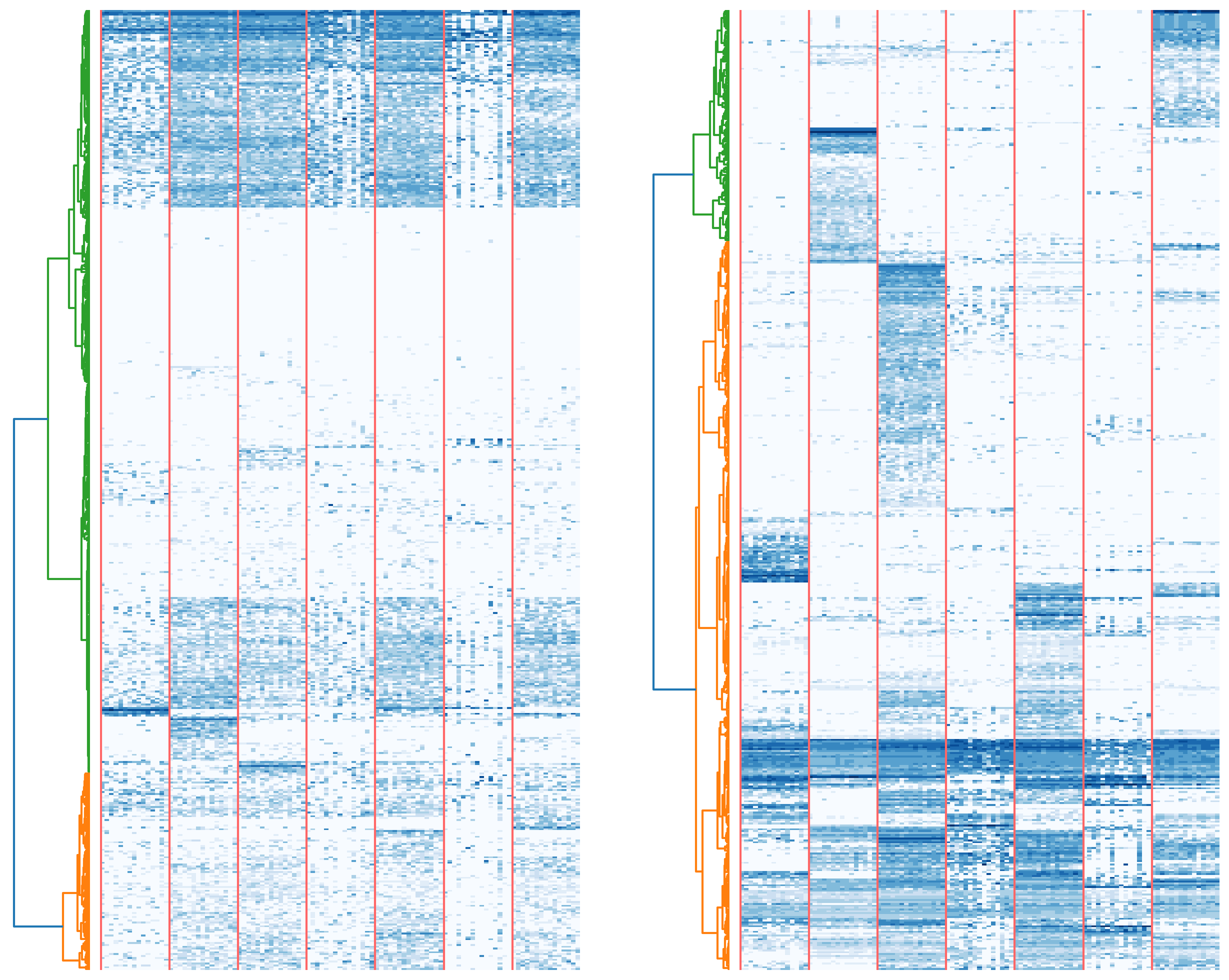}}
\\
\end{figure}

\begin{figure}[htbp]
\renewcommand{\thesubfigure}{\arabic{subfigure}}
\centering
\ContinuedFloat
\subfloat[Maria1]{
		\includegraphics[scale=0.07]{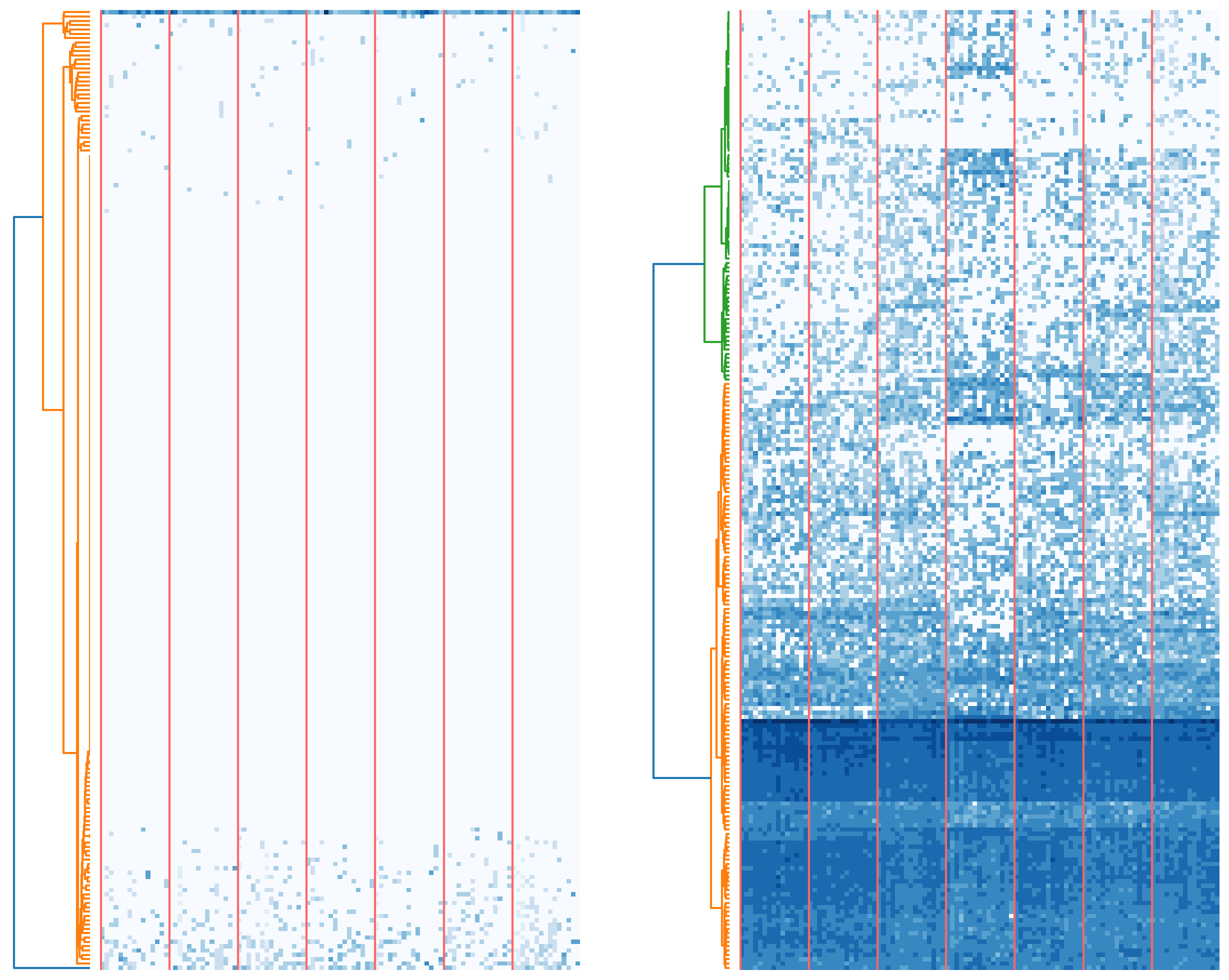}}
\subfloat[Maria2]{
		\includegraphics[scale=0.07]{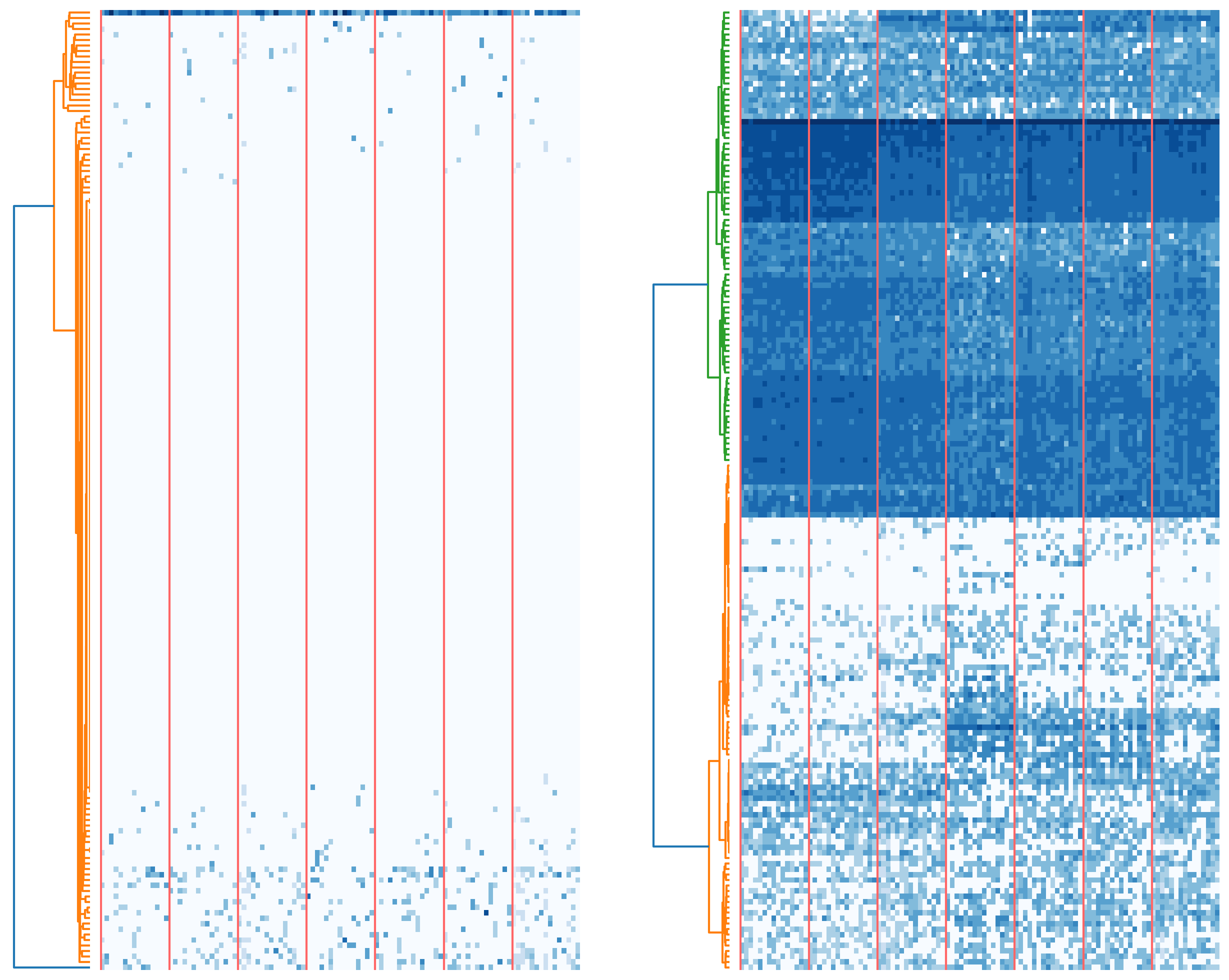}}
\\
\subfloat[Mouse Pancreas1]{
		\includegraphics[scale=0.07]{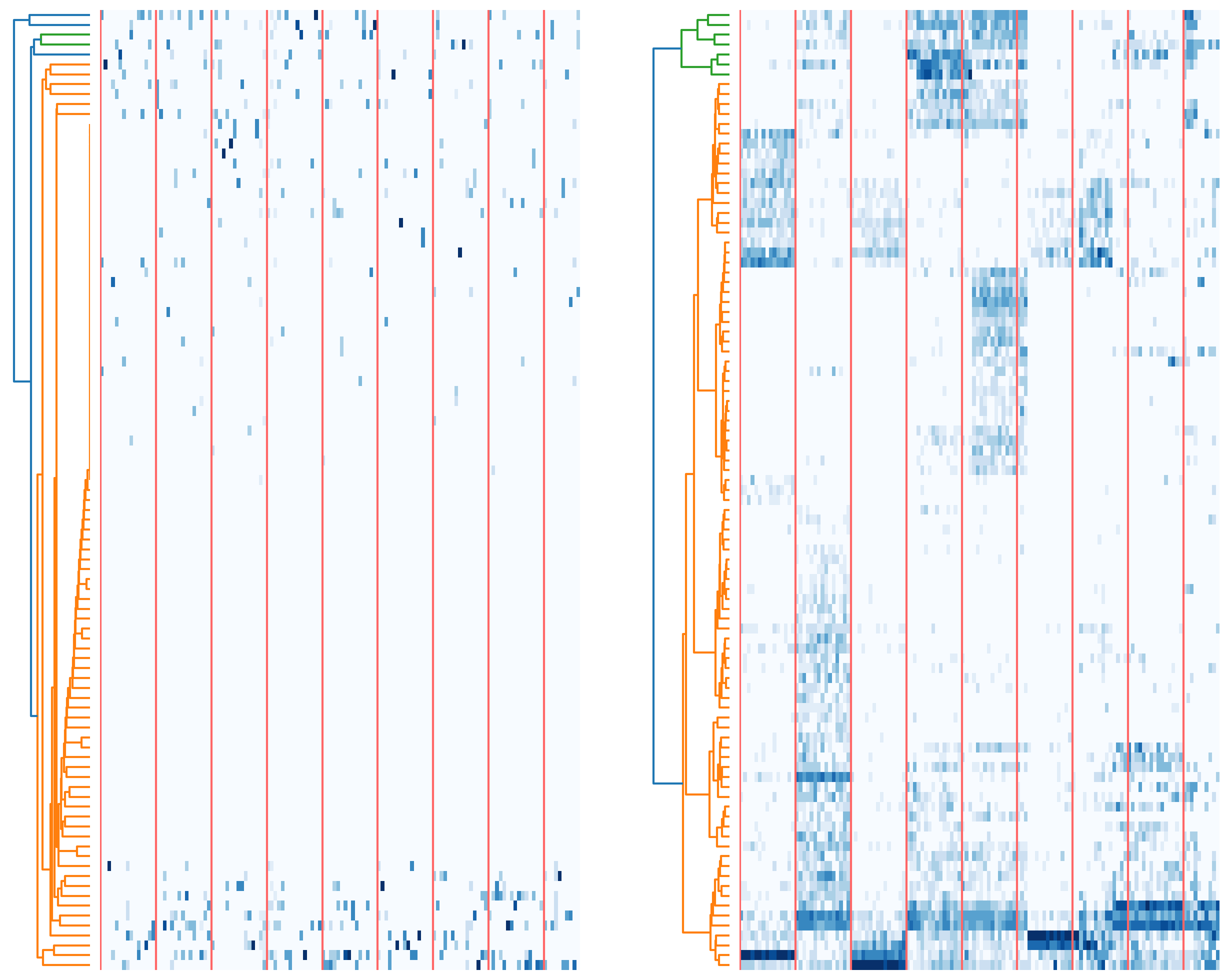}}
\subfloat[MacParland]{
		\includegraphics[scale=0.07]{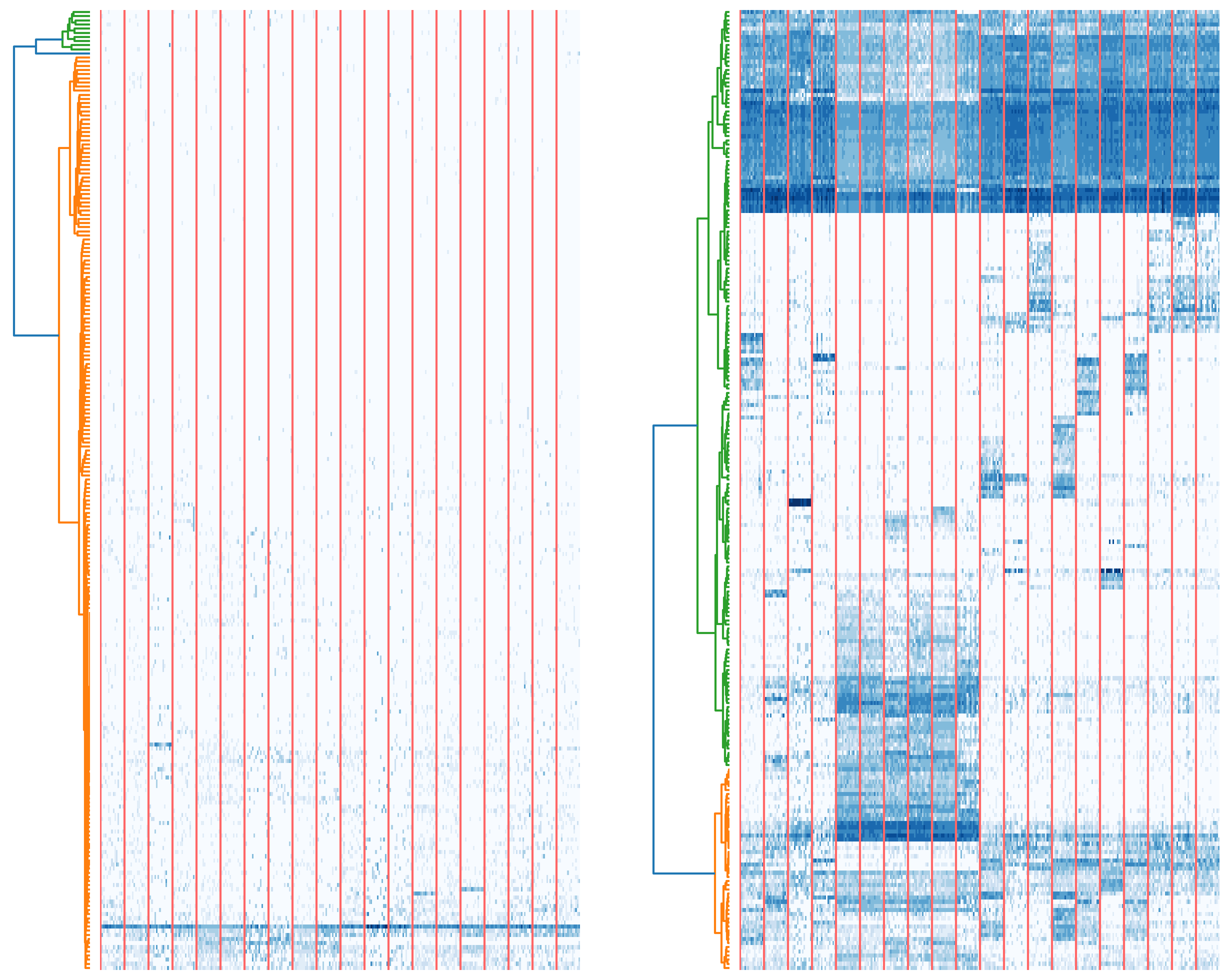}}
\\
\subfloat[Mouse Pancreas2]{
		\includegraphics[scale=0.07]{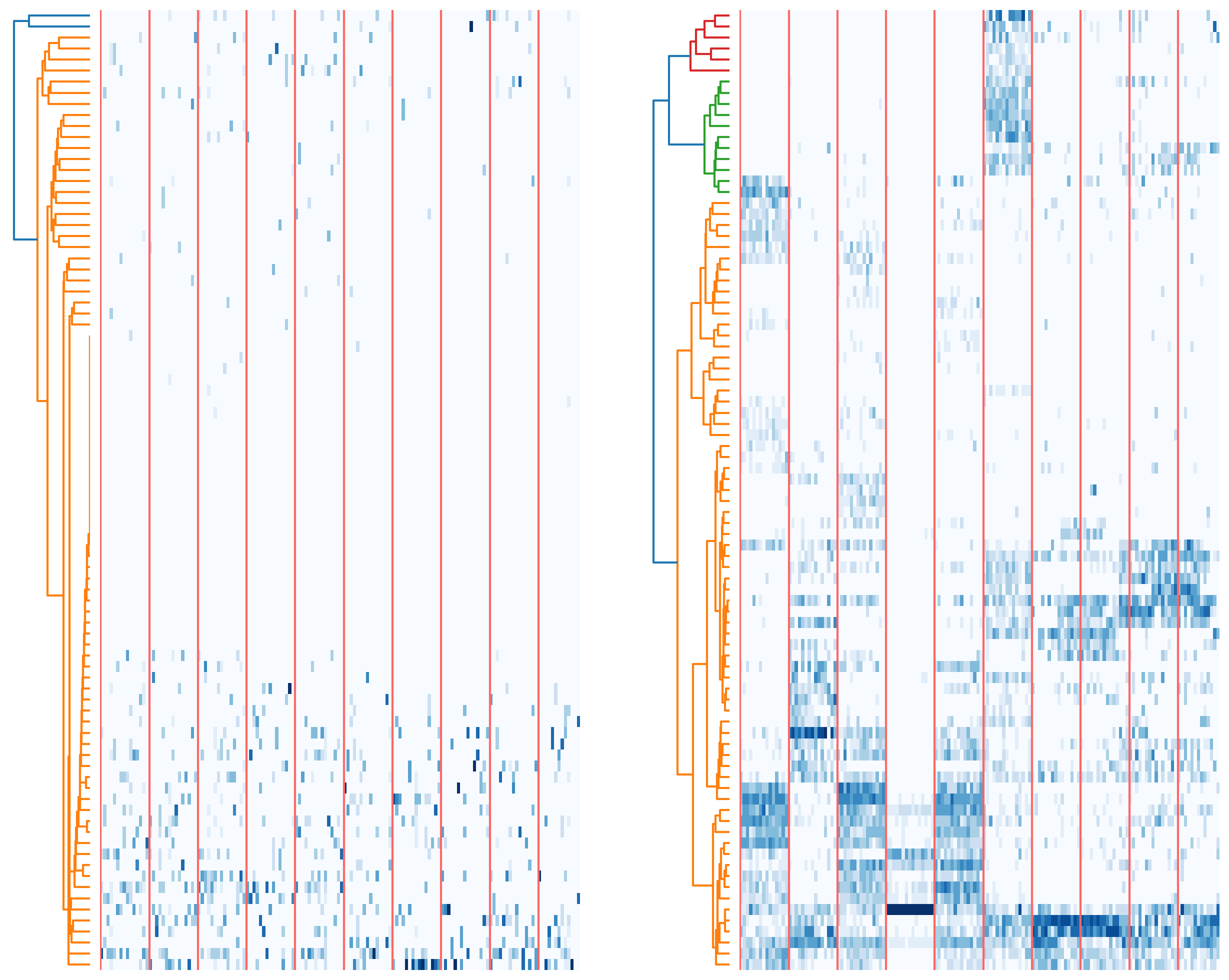}}
% \subfloat[Puram]{
% 		\includegraphics[scale=0.07]{figure/heatmap/Puram.png}}
\subfloat[Robert]{
		\includegraphics[scale=0.07]{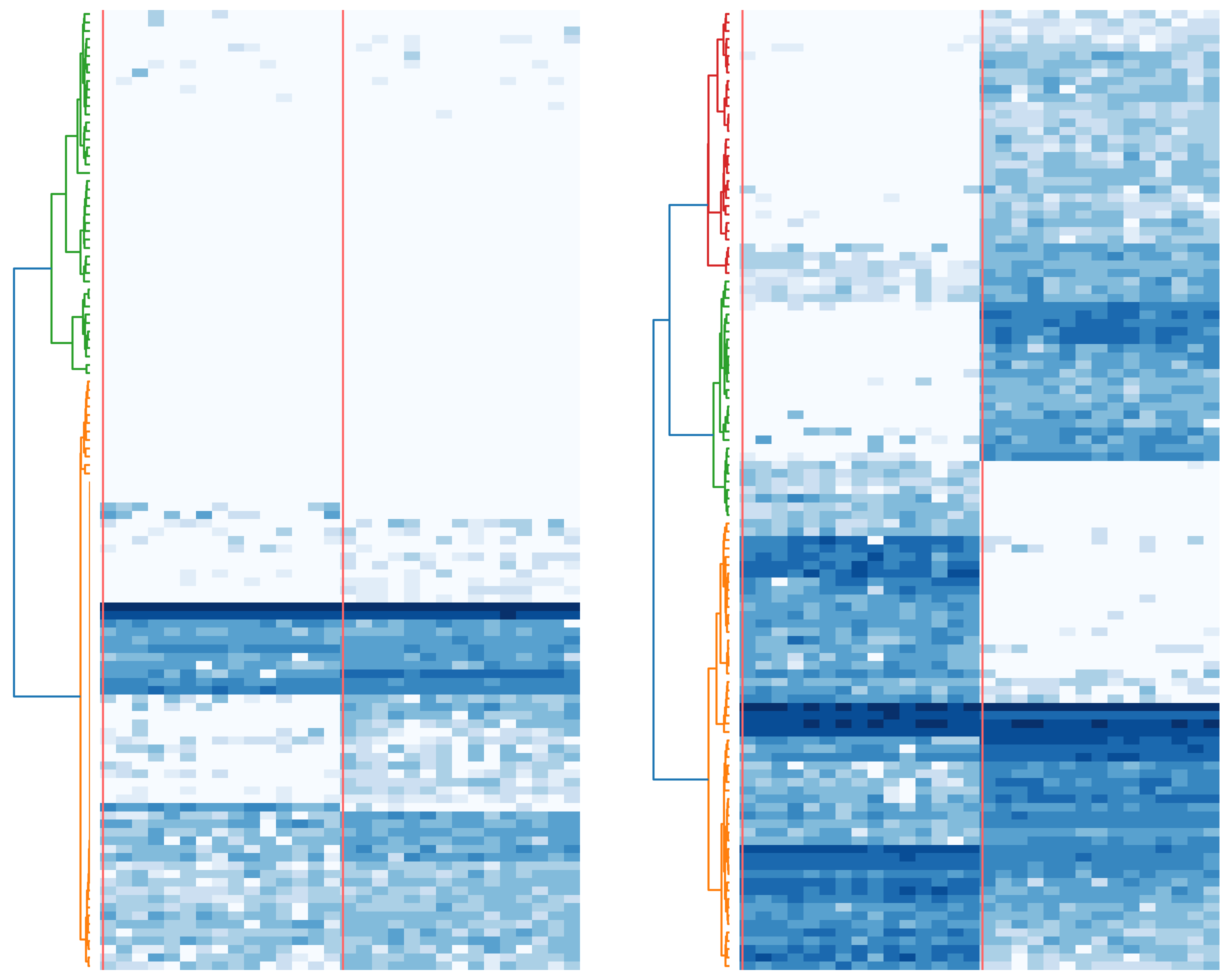}}
\\
\subfloat[Ting]{
		\includegraphics[scale=0.07]{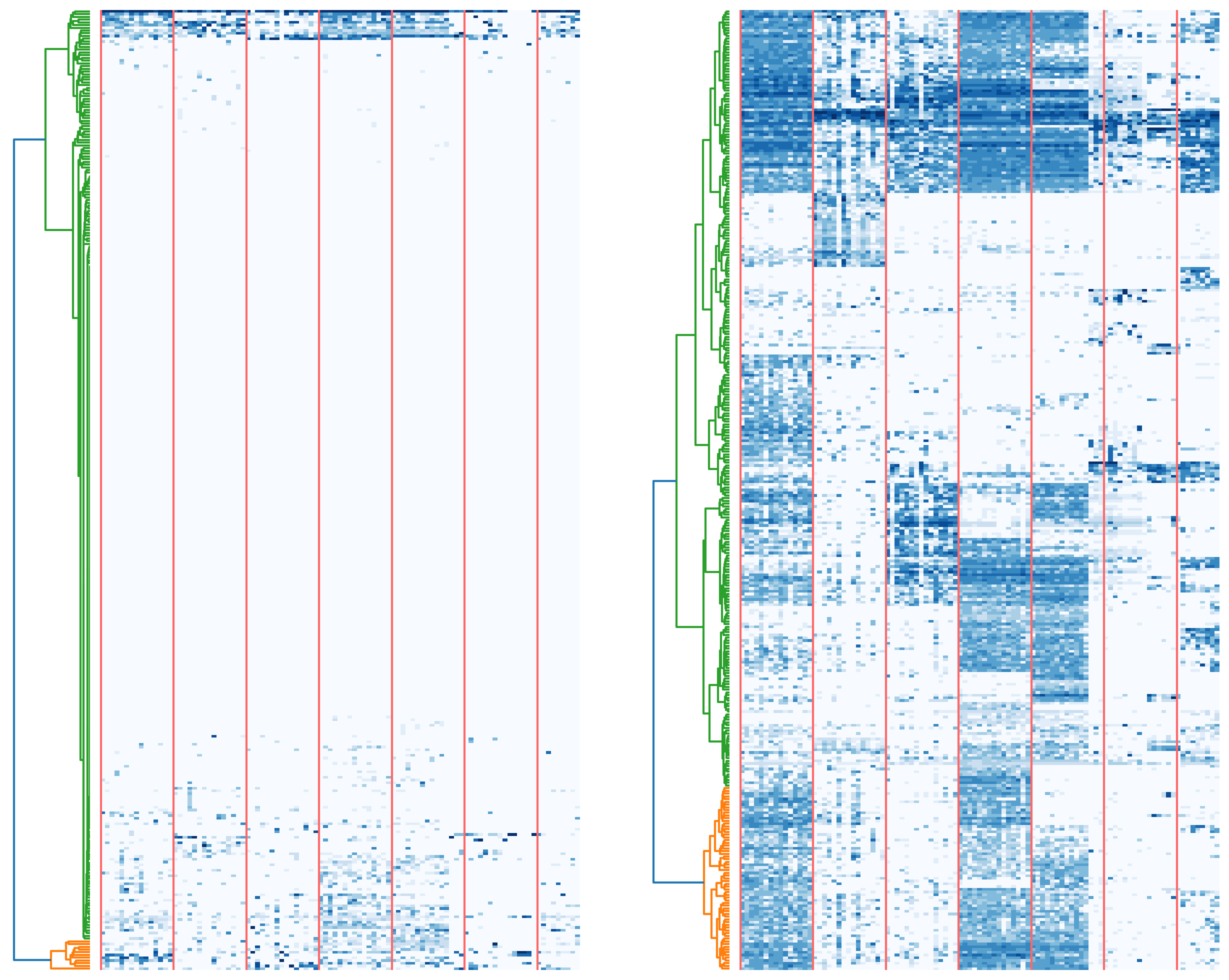}}
\subfloat[Yang]{
		\includegraphics[scale=0.07]{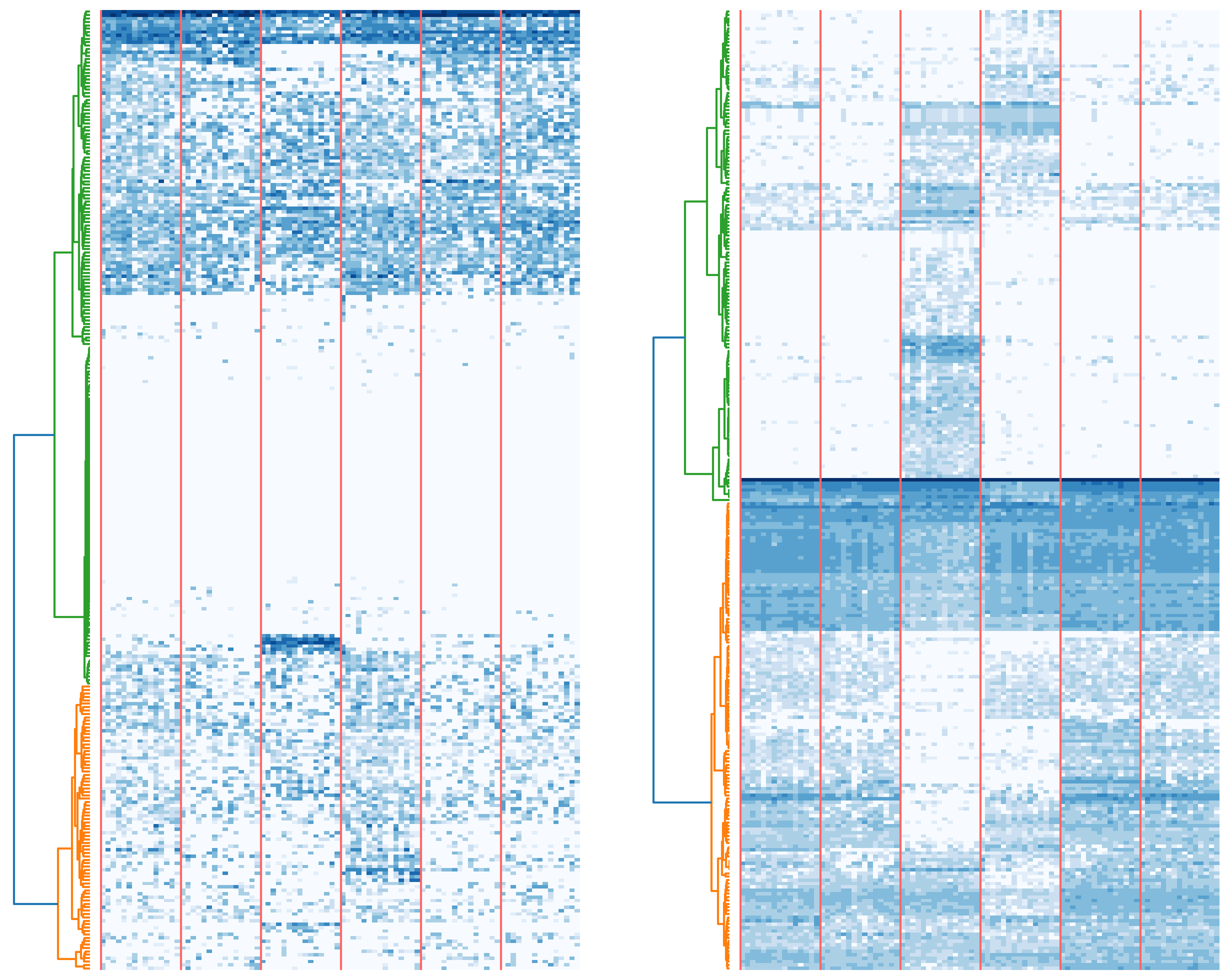}}
\\
\caption{Expression heatmap of the rest datasets, where the figure in the left panel is visualized from the original dataset.}
\label{heat_map}
\end{figure}

\clearpage

\end{document}